\documentclass[preprint,12pt]{elsarticle}
\usepackage[utf8]{inputenc}
\usepackage{times}
\usepackage{fancyhdr,graphicx,amsmath,amssymb}
\usepackage[ruled,vlined]{algorithm2e}
\include{pythonlisting}
\usepackage{booktabs}
\usepackage{graphicx}
\usepackage{hyperref} 
\usepackage{multirow}
\usepackage{subcaption}
\usepackage{amsmath}
\usepackage{tikz}
\usepackage{xcolor}
\usepackage{soul}
\usetikzlibrary{matrix,chains,positioning,decorations.pathreplacing,arrows}
\usepackage{bm}
\usepackage[justification=centering]{caption}

\usepackage{caption}
\usepackage{xcolor}
\usepackage{tabularx}
\usepackage{adjustbox}
\usepackage{setspace}
\usepackage{lineno}
\usepackage{ragged2e}

\DeclareCaptionLabelFormat{mylabel}{#1 #2}
\usepackage{xcolor}

\usepackage{natbib}

\begin{document}

\begin{frontmatter}

\title{Decision Support Framework for Home Health Caregiver Allocation Using Optimally Tuned Spectral Clustering and Genetic Algorithm}

% use optional labels to link authors explicitly to addresses:
\author[1]{S. M. Ebrahim Sharifnia \corref{cor1}}

\cortext[cor1]{Corresponding author}
\ead{ssharifn@vols.utk.edu}

\author[1]{Faezeh Bagheri}
% \ead{fbagher1@vols.utk.edu}
\author[1,5]{Rupy Sawhney}
% \ead{sawhney@utk.edu}
\author[1]{John E. Kobza}
% \ead{jkobza@utk.edu}
\author[1]{Enrique Macias De Anda}
% \ead{emaciasd@utk.edu}
\author[2]{Mostafa Hajiaghaei-Keshteli}
% \ead{mostafahaji@tec.mx}
\author[4]{Michael Mirrielees}
% \ead{m.mirrielees@gmail.com}

\affiliation[1]{Department of Industrial & Systems Engineering University of Tennessee, Knoxville}
\affiliation[2]{Campus Puebla, School of Engineering and Sciences , Tecnológico de Monterrey}
\affiliation[4]{Independent Researcher}
\affiliation[5]{Deceased Co-author}

\begin{abstract}
%% Text of abstract

\par
Population aging is a global challenge, leading to increased demand for health care and social services for the elderly. Home Health Care (HHC) is a vital solution to serve this segment of the population. Given the increasing demand for HHC, it is essential to coordinate and regulate caregiver allocation efficiently. This is crucial for both budget-optimized planning and ensuring the delivery of high-quality care. This research addresses a fundamental question in home health agencies (HHAs): “How can caregiver allocation be optimized, especially when caregivers prefer flexibility in their visit sequences?”. While earlier studies proposed rigid visiting sequences, our study introduces a decision support framework that allocates caregivers through a hybrid method that considers the flexibility in visiting sequences and aims to reduce travel mileage, increase the number of visits per planning period, and maintain the continuity of care – a critical metric for patient satisfaction. Utilizing data from an HHA in Tennessee, United States, our approach led to an impressive reduction in average travel mileage (up to 42\%, depending on discipline) without imposing restrictions on caregivers. Furthermore, the proposed framework is used for caregivers’ supply analysis to provide valuable insights into caregiver resource management.

\end{abstract}


\end{frontmatter}
%%%%%%%%%%%%%%%%%%%%%%%%%%%%%%%%%%%%%%%%%%%%%%%%%%%%%%%%%%%%%%%%%%%%%%
\section{Introduction}
\par
The increasing phenomenon of population aging has become a noteworthy issue in contemporary society. Recent demographic findings derived from WorldMeters\footnote{https://www.worldometers.info/demographics/life-expectancy}, highlight that the mean life expectancy in the year 2020 reached 72.91 years, which signifies an advancement of approximately 1\% within the last five years and 3\% over the preceding decade. Population aging has profound and wide-ranging consequences that significantly impact society \citep{grinin2023global}.

One prominent consequence is the escalating demand for healthcare and social services \citep{jakovljevic2023global}. As individuals age, they become more susceptible to chronic illnesses and age-related conditions, necessitating specialized care and support. The increasing prevalence of these healthcare needs places considerable strain on healthcare systems and social service programs, posing challenges in resource allocation, prolonging wait times for treatment and services, and potentially compromising overall quality of care \citep{fathollahi2022bi}. To address these challenges, Home Health Care (HHC) is emerging as a promising solution to bridge the gap between imbalanced demand and available resources \citep{alves2023hybrid}. 

\par

HHC is a type of healthcare service that provides medical and non-medical support to individuals in their own homes. HHC services can include medical care, nursing assistance, therapy, assistance with daily activities, and other forms of personalized care delivered by healthcare professionals. The goal of HHC is to provide individuals the necessary care and support while remaining in the comfort and familiarity of their own homes, promoting independence and improving overall well-being \citep{gupta2023home}.

Home health agencies (HHA) provide care that is not only cost-effective and accessible but also comparable to, or even surpassing, the quality of care offered by hospitals and nursing homes \citep{johnn2023stochastic}. Implementing HHC programs has the potential to alleviate the strains experienced by hospitals during pandemics such as COVID-19, while simultaneously mitigating the risk of disease transmission \citep{alaszewski2023managing}. To leverage the full potential of HHC in the most efficient manner, the optimization of processes is crucial \citep{belhor2023learning}. By optimizing various levels of decision-making in an HHA, the healthcare system can effectively take the benefits of home care to meet the changing needs of the population, ensure equitable access to high-quality care, and mitigate the risks posed by infectious disease outbreaks.

\par
In designing HHC, a multilayered decision-making schema is employed, generally delineated into three core classifications: Strategic, Tactical, and Operational planning \citep{chabouh2023systematic}. The delivery of healthcare is an intricate network where strategic planning plays a pivotal role in addressing the structural intricacies tied to the system's configuration and magnitude. This level of planning frequently relies on comprehensive data and projections spanning extensive periods \citep{alves2023systematic}. Tactical planning, in contrast, pivots towards actualizing strategic plans, thus it is inherently linked to decision-making within short-to-medium-term scopes. Operational planning, at its most immediate level, involves making critical decisions that directly influence the execution of healthcare delivery \citep{restrepo2020home}.

The allocation of caregivers is a complex process that encapsulates multiple dimensions and is not confined solely to the caregivers' satisfaction. The longevity of any commercial enterprise hinges critically on the assurance of satisfaction among its customers and workforce. Specifically within the purview of HHC, the satisfaction of patients could be engendered through the provision of high-quality care \citep{agustina2023hhc}. 

\par

Improving the quality of care by appropriately assigning caregivers to patients within the HHC context necessitates addressing numerous complex components. Initially, the alignment of caregivers' competencies and qualifications with the service needs of patients is a critical factor. Each patient might require a specific skill or a limited set of skills \citep{di2021routing}.
Subsequently, the principle of Continuity of Care is paramount. This principle ensures that a consistent team of healthcare professionals is responsible for each patient. Complete continuity of care is achieved when a single caregiver performs all visits to a specific patient within a predetermined time frame \citep{lahrichi2022first}. Such timeframes could be either a standard 30-day period or a 60-day certification episode. Hence, maintaining this consistency outside of weekly planning periods is of vital importance \citep{shi2022analysis}.
Moreover, the complexity of workforce allocation can be exacerbated by temporary or permanent interdependencies between care services, which can manifest itself as precedence or disjunction when certain services for a patient must follow or be separated from another set of services \citep{shahnejat2021robust}.

\par
This study is inspired by the rising dissatisfaction rates among caregivers and patients, as reported by an HHA in Tennessee, United States. Through surveys, interviews and discussions with HHA managers and staff, we have identified several concerns and challenges in workforce management.
Our primary observation was the increasing caregiver turnover rate, attributed to substantial dissatisfaction. The underlying causes were identified as perceived disparities between remuneration and responsibilities, inflexibility in visitation sequences, and excessive workload.
We determined that current workforce planning and sequencing approaches are not viable for daily operations, given various inherent uncertainties. Furthermore, the agency's ineffectiveness in properly allocating patients to caregivers has led to critical issues concerning the assurance of care quality.

\par
The objective of this research is to develop a framework that facilitates caregiver allocation, addressing the aforementioned concerns. Most previous studies on HHC planning have adopted traditional industrial methodologies, treating human resources as machines, and viewing work in processes similar to a production line \citep{sawhney2020conceptual}. To date, there is a dearth of research proposing solutions that adequately support carryovers' personal lives by affording greater autonomy.
Our proposed framework helps HHAs in two critical decision-making areas. Firstly, workforce allocation is determined considering the geographical location of the workforce and patients, patient preferences, and an emphasis on the continuity of care regarding patient satisfaction with the dual aims of cost minimization and quality of care maximization. Secondly, the recruiting system is subjected to a sensitivity analysis to ascertain which scenario would yield the most effective workforce allocation. To our knowledge, no previous studies in the field of HHC have developed a solution platform that allows caregivers to be flexible in their visit sequence while addressing uncertainty. This platform would also allow HHA managers to contribute to the planning process by controlling service parameters, such as caregiver travel patterns.

\par
The remainder of the paper is laid out as follows. In \textbf{Section \ref{Literature review}}, the literature addressed the Home Healthcare Routing and Scheduling Problem (HHCRSP) is discussed. Problem specification is discussed in \textbf{Section \ref{Problem statement}}, and the proposed framework is designed accordingly. \textbf{Section \ref{Decision_support_framework}} presents the computational result of the case study and reveals some remarks. In \textbf{Section \ref{Computational_Analysis}}, the framework results are given for the case study, and the sensitivity of its output with respect to various scenarios is analyzed. Lastly, this research wraps up in \textbf{Section \ref{Conclusion}} with some potential future directions.

\section{Background}
\label{Literature review}

HHC is a collection of people and material resources used to provide essential services to patients in their homes. Due to various problematic challenges created by this new alternative to conventional hospitalization, this burgeoning service industry has received a lot of attention from scholars and practitioners during the previous several decades \citep{euchi2022home}. In tandem with the rise in demand, the management of resources is problematic for HHAs, especially for caregivers who, unlike robots, due to fatigue, cannot maintain consistent efficiency \citep{behrens2023fatigue}. As a consequence, the task of caregiver management is growing increasingly complex, thus necessitating additional investigation and analysis. Within the literature of HHC, this planning level has been examined under the purview of the HHCRSP. The principle of HHCRSP revolves around determining the most efficient method for allocating a group of caregivers, who deliver in-home medical assistance to a multitude of patients across a range of circumstances \citep{marcon2017caregivers, erdem2017two,fikar2017home, di2021routing}. 

\par
 Efficient caregiver allocation planning has different influential dimensions that require to be considered. These dimensions could be attributed to caregivers’ specifications, patients’ needs, and preferences of both. The study by \citet {naseem2011impact} confirms that the survival of any business is contingent on ensuring the satisfaction of everyone engaged in the system through fulfilling their needs. 
 
 \par
In connection with HHC, satisfaction has a distinct interpretation for caregivers and patients. Mainly, for caregivers, satisfaction in HHCRSP could be achieved by minimizing the travel time, and transportation cost to the patient’s location \citep{chaieb2020decomposition,nikzad2021matheuristic,alkaabneh2023multi},  minimizing working time of patients visiting \citep{frifita2020vns, malagodi2021home}, minimizing wait times by synchronizing schedules on both ends \citep{euchi2020optimising,liu2021hybrid, du2022multi}, and minimizing the overtime \citep{kandakoglu2020decision,naderi2023novel}.

The level of contentment experienced by patients could be elevated principally if continuity of care is maximized \citep{grenouilleau2019set, krityakierne2022nurse}, if the violation of the proposed availability time slot by patients is minimized \citep{nikzad2021matheuristic,yadav2022integrated}, if caregivers visiting within the preferred time spot by patients is maximized \citep{braekers2016bi,erdem2017two}\cite{}, and if the number of uncovered visits are minimized \citep{demirbilek2021home}.

In addition to the two above-mentioned conditions, there are some known preferences that play the role of a stimulus for strengthening the fulfillment level of both sides.  For example, in the past, wages and pensions were always at the forefront of employees' worries. Nowadays, workload, understaffing, and overtime are three of the most often raised concerns for them. Following this, today’s caregivers’ concerns could be construed in terms of a “Balanced workload” \citep{shi2018modeling, hassani2021scenario}. Besides, patients’ desires in terms of caregivers’ characteristics such as age, gender, and skills are considered as a patient’s preference and could enhance their satisfaction level \citep{di2021routing}.

\par
\begin{figure}[!h]
    \centering
    \includegraphics[width=1
\textwidth]{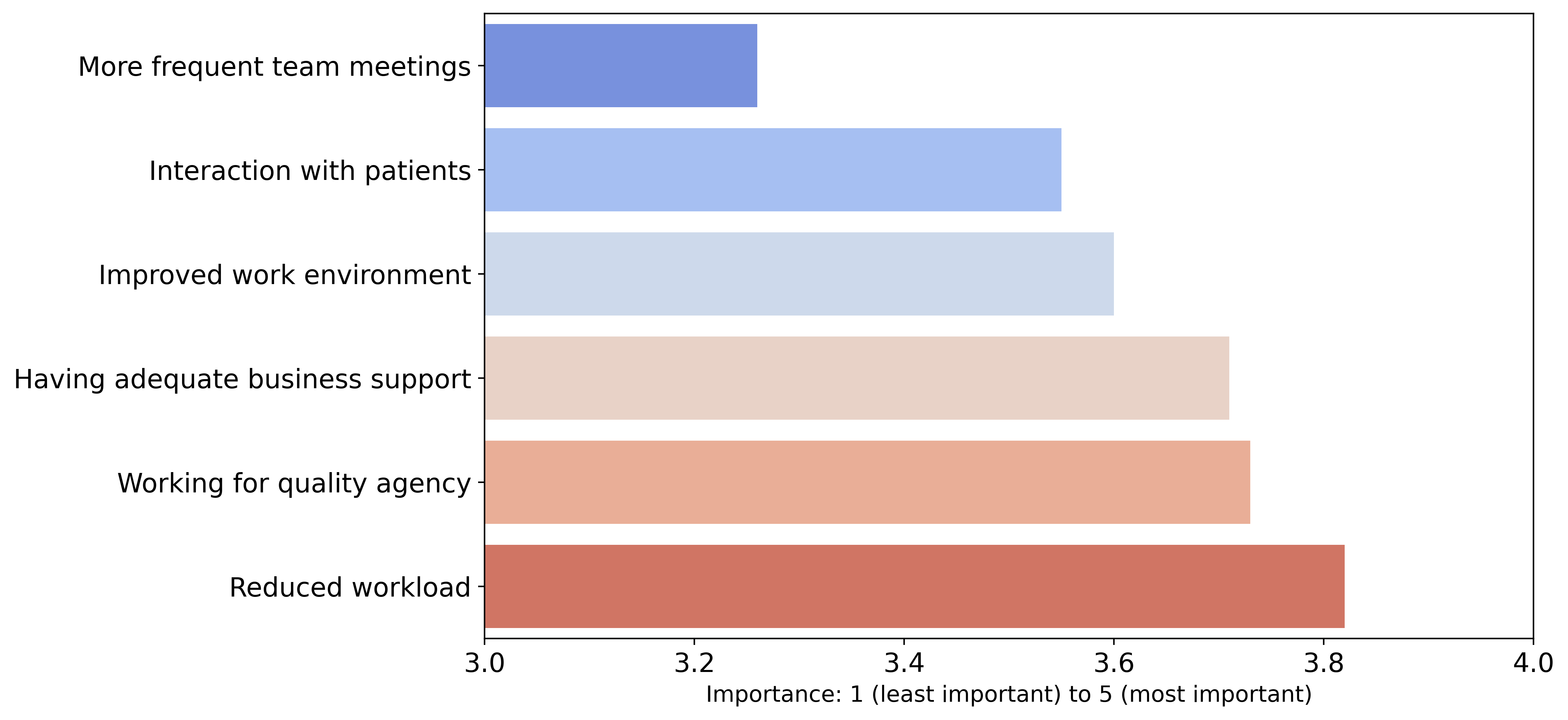}
    \caption{The influential factors in increasing job satisfaction and the quality of care in HHC \citep{navaie2004increasing}.}
    \label{fig_1}
\end{figure}

One of the challenging decisions that HHAs face because of demand growth is workforce allocation, which requires tactical planning \citep{de2023home}. This kind of planning is related to the level of employee satisfaction, which is an inherent characteristic to achieve operational excellence \citep{sawhney2020conceptual, kelly2021impact}. 
The study on the key elements influencing job satisfaction and quality of care in HHC, as shown by \cite{navaie2004increasing}, indicates that decreasing the workload is the primary factor directly tied to HHC workers' contentment. \textbf{Figure \ref{fig_1}} provides a visualization of the importance of each factor in determining caregiver satisfaction. From this study, it was underscored that to prevent staff burnout, it's crucial to enable staff to balance their professional and personal duties in a flexible manner.

Caregivers’ management and allocation by considering above mentioned dimensions are a great deal for HHC agencies. To face them, they continually need a practical tool to assist them with this matter. With reference to HHCRSP, there is a rich literature that contains a substantial number of studies examining diverse types of HHCRSP and offering various efficient solutions \citep{euchi2020optimising}; However, the limited number of studies that provide feasible future-proof decision frameworks motivates us to proceed with further investigation. 

In relation to this unavoidable necessity for HHC agencies, by taking into account the dynamic nature of systems that deal with people and the approach that has been employed in the proposed framework for managing the workforce, the literature of the present research is best categorized under three headings: the decision support framework in HHCRSP, the dynamics of HHCRSP, and clustering algorithm for HHCRSP.

\subsection{Decision support framework in HHCRSP}
The starting point of doing research on the topic of HHCRSP was initiated by \cite{begur1997integrated}, who carried out the first investigation on this topic and proposed a spatial decision support system, incorporating GIS technology to allocate the nurse to patients according to their requirement and availability and determine which route will result in the minimum travel time for nurses. 
A decade later, \citet{eveborn2006laps} detailed a Laps Care using a repeated matching approach based on a hybrid of heuristic and optimization techniques. Their formulated Decision Support System (DSS) is basic and tries to define the best assignment, which has an elevated level of matching with the patient’s needs. 

In 2015, \citeauthor{duque2015home} devised a bi-objective DSS to enhance the process of caregiver allocation to patients, employing a two-tiered strategy. The first stage involves assigning caregivers, taking into consideration the compatibility of patient preferences and time slot availability. The second stage calculates the total travel distance for each caregiver to determine the most efficient route. The foundation of this DSS lies in heuristic algorithms, starting with the formation of potential allocation schemes based on combinations of caregivers and time slots. Through an iterative process within established constraints, the optimal allocation scheme is identified.

\begin{figure}[!h]
    \centering
    \includegraphics[width=1.0 \textwidth]{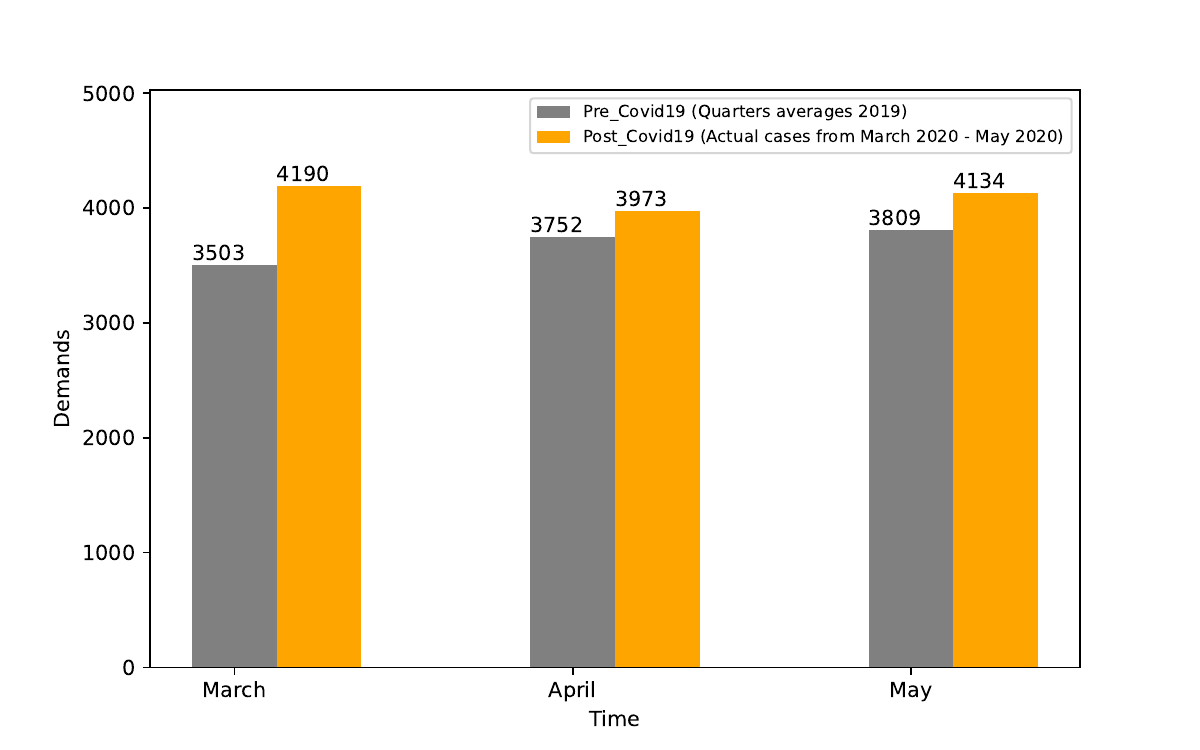}
    \caption{Change in demand for “Registered Nurses”, Pre and Post COVID-19.}
    \label{fig_2}
\end{figure}

\citet{kandakoglu2020decision} developed a DSS for routing and scheduling caregivers skilled in dialysis. The DSS uses a mixed integer linear programming model (MILP) to optimize nurse schedules based on multiple criteria such as travel distance, expenses, overtime, and workload. Their formulated system allows users to prioritize objectives by adjusting their weightings. Besides initial planning, the DSS can also accommodate last-minute schedule changes by assigning available nurses in proximity to the patient or those based in a clinic.
A detailed examination of the data revealed that the evolving demand brought about by the COVID-19 pandemic has aggravated the existing situation. \textbf{Figure \ref{fig_2}} shows the change in demand for "Registered Nurses", the most in-demand skill, in the HHA.

The decision support framework devised by \citet{nasir2020decision} effectively addressed the genuine requirements of HHC services. It comprehensively grapples with the intricacies of formulating concurrent schedules and route plans for a cohort of HHC staff and Home Delivery Vehicles (HDVs). Essential factors incorporated within this framework encompass the synchronization needs between the visits of HHC staff and HDVs, the necessity of multiple visits to patients, and the creation of numerous routes for the HDVs. Their framework  suggests HHC planning decisions, through running their developed hybrid genetic algorithm (GA). 

In addition to providing care to patients, some HHC agencies provide social services for elderly people with diminished autonomy. \cite{vieira2022optimizing} considered social services and investigated home health and social routing and scheduling problems by considering synchronizing among their tasks and designed DSS, accordingly. To run the proposed DSS, they developed two greedy heuristics algorithms that seek to schedule caregivers and social care providers with the aim of maximizing the continuity of care and minimizing ineffective working time. The obtained solution through their DSS has been adjusted by imposing a set of hard constraints, including working regulations, time windows of availability of both sides, and matching what the patient required. 

\subsection{Dynamics of HHCRSP} 
Dynamism is a key attribute of any system that involves human participation. Operational planning that neglects to consider this dynamic element is likely to encounter difficulties when implemented in a real-world context \citep{zhou2008towards}.
In the context of HHC, dynamism in workforce planning arises from a variety of sources. Due to the significant influence on workforce planning, several studies have addressed and studied them. 

\citet{moz2004solving} explored the dynamic aspect of caregiver allocation planning, considering scenarios such as nurses being unable to fulfill all their scheduled shifts, the need for additional but unprovided medical attention, or the cancellation of pre-arranged visits.   \citet{erdem2017two}  embedded dynamism in their methodology by accommodating simultaneous on-the-spot visit requests from a cluster of patients. \citet{demirbilek2019dynamically}  translated dynamism into the fluctuating rate of patient arrivals, necessitating swift decision-making about appointment acceptance and scheduling. 
Dynamism caused by unanticipated incidences ranging from cancellation of scheduled appointments to the need for unexpected medical assistance, or broken medical equipment were also addressed in \citep{du2019real,oladzad2022dynamic}.

\subsection{Clustering approach for HHCRSP} 
The process of clustering is a method that is used to proactively organize enormous volumes of data into smaller groups or clusters that have similar characteristics \citep{ezugwu2022comprehensive}. Clustering algorithms may be used to handle a wide variety of data types, ranging from raw data to the results of various simulations and procedures. In the literature of HHCRSP, several studies have leveraged the advantages of clustering techniques in their proposed methodology to realize their planning objectives. 

\citet{rasmussen2012home} harnessed the power of the clustering algorithm within three distinct schemes in the branch-and-price algorithm, thereby accelerating the optimization of home health care crew scheduling significantly.
Subsequently, \citet{quintana2017clustering} put forward an agglomerative greedy clustering algorithm aimed at minimizing the total travel distance in the large-scale, real-world scheduling problem for caregivers in Madrid.
In a novel approach, \citet{riazi2018column} applied the clustering algorithm in their column-generating method, supplemented by the gossip algorithm. This innovative strategy assured the diversification and intensification of the search space, proving to be efficacious in resolving caregivers' vehicle routing problems within specific time windows.
\par
Taking a more comprehensive approach, \citet{pahlevani2022cluster} devised a solution strategy for the simultaneous planning of medical, social, and paramedical service delivery. They designed a multi-step clustering method, incorporating a modified k-means clustering algorithm, Ordering Points to Identify the Clustering Structure, and Agglomerative Hierarchical Clustering.
More recently, \cite{belhor2023multi} proposed a hybrid method that combines the non-dominated sorting GA and the Strength Pareto Evolutionary Algorithm (SPEA) with the k-means clustering algorithm. This hybridization enhances the convergence and diversification of the derived Pareto sets, thereby efficiently solving the bi-objective HHCRSP. This approach is designed to accommodate customer preferences for arrival and departure times, optimizing service duration, and minimizing overall tardiness.

\section{Problem statement}
\label{Problem statement}

\subsection{HHC provider's concerns} 
\label{HHC agency's concerns}
\par
This study was principally driven by a collaboration with a home health and hospice agency, hereinafter referred to as the HHA, that provides care to more than 5,000 patients annually in the Eastern Tennessee region of the United States. The HHA provides in-home clinical, therapeutic, and hospice care and employs workers from a wide range of disciplines including Physical Therapists (PT), Physical Therapy Assistants (PTA), Registered Nurses (RN), Certified Nursing Aides (CNA), Licensed Practical Nurses (LPN), Occupational Therapists (OT), Certified Occupational Therapy Assistants (COTA), Chaplains (CH), Speech-Language Pathologists (SLP or ST for Speech Therapy), and Social Workers (MSW) for its patients.

\par
The goals of the collaboration with the HHA centered on reducing the high attrition rate of skilled clinical workers, improving patient dissatisfaction with visit planning, reducing travel-related costs, and reducing the significant logistics re-work performed by Schedulers. The combination of these issues was resulting in frustrations between staff and leadership while the agency struggled with stunted growth and reduced capacity to serve their patients. Patient satisfaction scores from standardized industry surveys (HHCAHPS and CAHPS Hospice\footnote{https://www.cms.gov/data-research/research/consumer-assessment-healthcare-providers-systems}) are also publicly reported, meaning that sustained difficulties with patient perception of the visit planning domain can result in damage to the agency’s long-term reputation as patients and referral sources consider these ratings when choosing their care providers. The current-state complexities of these issues were investigated by employee interviews and analysis of business data, providing insight into the factors related to each.

Firstly, caregivers consistently expressed concerns about the disparity between their pay and workload. They claimed that responsibilities are frequently distributed unfairly, with some caregivers being assigned additional duties related to their peers without a corresponding difference in their pay or benefits.

Secondly, the deficiencies expressed by patients in their satisfaction surveys related to differences in the stated plans of the agency compared to the actual timing of the visit, or the lack of communication about a plan at all. The current operational framework of the agency relies on caregivers contacting patients a day in advance to schedule visits. However, the unpredictable and dynamic nature of the caregivers’ professional environment often leads to changes in the schedule, as the duration of visits, travel time, and patient availability can be uncertain. When those changes occur on the date of service, pre-planning is barely possible, and patients perceive that the agency is disorganized or not having their care as the first priority.

Thirdly, a review of mileage data showed that inefficient routing and changes in the plan after the clinician had already traveled to the service location resulted in wasted resources. Upstream changes created a chain of events that wasted the time of clinicians, administrative and office staff, caused tension between supervisors and their workers, and wasted gas and car costs.

Lastly, caregivers have expressed dissatisfaction with their rigid schedules. Interviews revealed that they would be more satisfied if they were able to decide the sequence of their visits. This flexibility and autonomy would allow them to adapt their plan to respond to changes in patient availability, handle their personal tasks during the day, between visits, and schedule their next visit based on their current location and schedule.

From an organizational viewpoint, managers at  face challenges in determining the optimal number of caregivers required and the extent to which recruiting additional personnel would improve caregivers’ schedules. These challenges are compounded by varying patient densities across different regions. In regions with low patient density, caregivers may have excessive travel time between patients, which can lead to delays in providing care and burnout. In regions with high patient density, caregivers may be overloaded with patients, which can lead to stress and decreased quality of care. These challenges highlight the need for a more systematic approach to caregiver allocation that takes into account the factors of patient density, travel time, and caregiver workload. Managers at the HHA face the challenge of allocating new patients among the available caregivers. This allocation process is complex and requires careful consideration and decision-making on the part of the managers.

The scheduling process at the HHA has been identified as a weakness within the system, despite the use of commercial home care planning software. The transition to a new electronic medical record and scheduling interface has led to inefficiencies and disrupted workflows, impacting the daily operations of supervisors, schedulers, and clinicians. Consequently, these operational issues have resulted in missed patient visits, increased travel distances, and frustration among both administrative and clinical team members.

\begin{figure}[!h]
    \centering
    \includegraphics[width=1.1\textwidth]{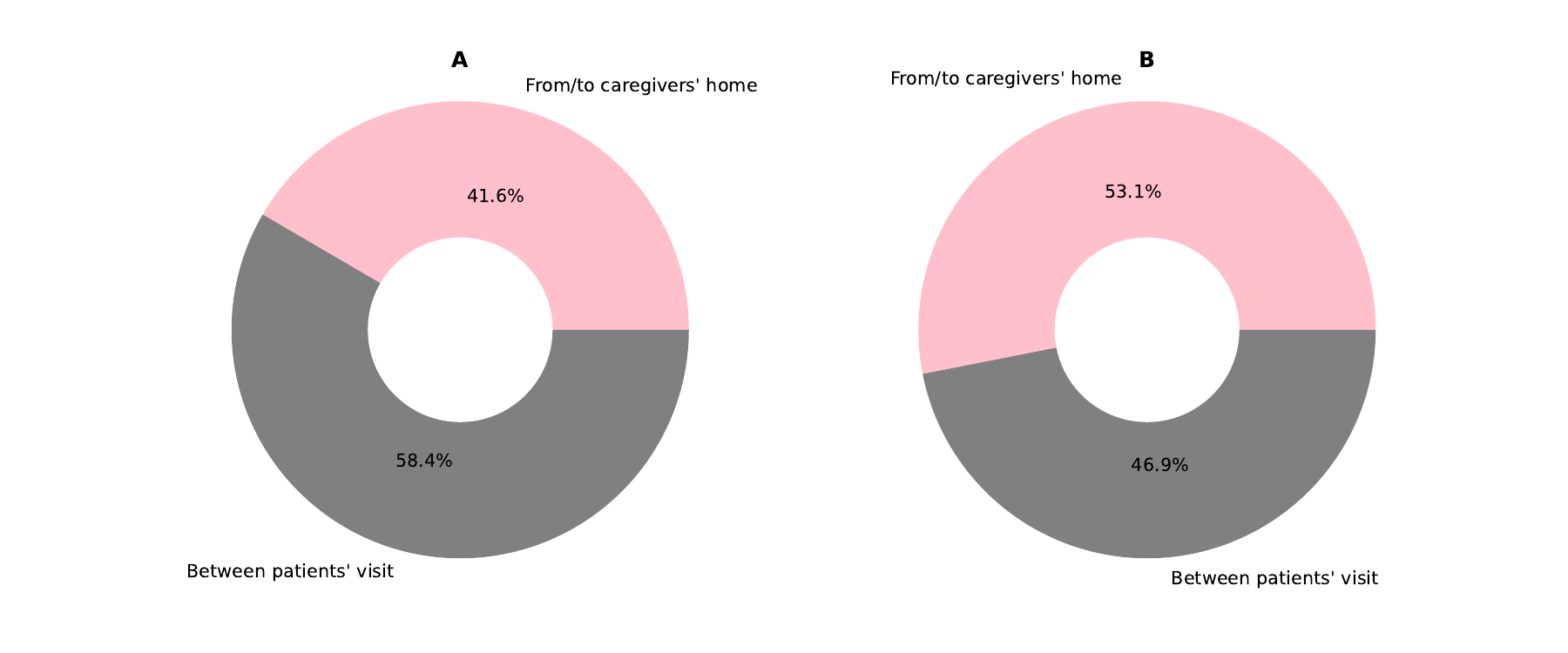}
    \caption{\textbf{A}. Percentage of travel type count in a year, \textbf{B}. Percentage of travel type mileage in a year.}
    \label{fig_3}
\end{figure}

\begin{figure}[!h]
    \centering
    \includegraphics[width=0.9\textwidth]{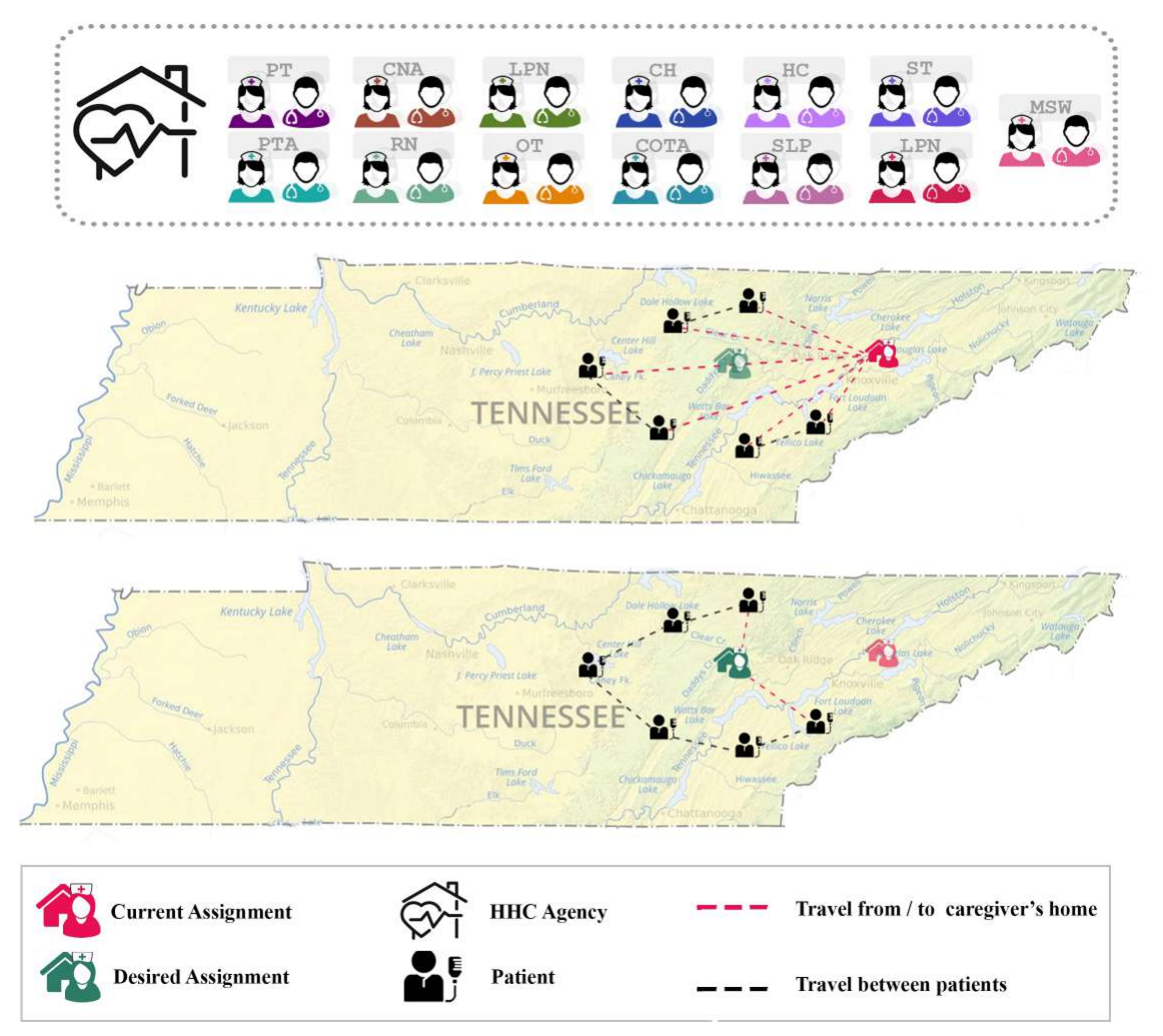}
    \caption{Current caregivers’ assignment \textit{vs.} desired one.}
    \label{fig_4}
\end{figure}

One example of the current visit planning inefficiencies is the ratio of travel from/to the caregiver’s home and from/to the patient's home, as depicted in \textbf{Figure} \ref{fig_3} (A) and (B). High mileage and trip count related to the caregiver’s home suggests that the assignment of patients to caregivers lacks efficiency unless the visiting area has a very high population density and the patients are very close together. The ratio was approximately 40\% home-related to 60\% patient-related. A ratio with higher patient-related miles could mean less wasted travel costs because the caregiver is returning home fewer times during the day and has been assigned patients closer to their point of origin. To illustrate the preferred state compared to the current state, \textbf{Figure} \ref{fig_4} presents a visual comparison.

\par
The scope of this study is to address the challenges faced by the HHA and to develop a decision support framework that can provide efficient caregiver assignments to patients, which provides sustainable and profitable benefits. According to our conversation with the HHA leadership, a multifactorial approach was developed to achieve the stated agency goals.

\par
Firstly, the agency aims to increase the amount of time caregivers spend providing direct care to patients, called 'actual' working time, compared to the time spent traveling to and from the locations of patients. This objective involves minimizing the itineraries that require caregivers to depart or return to their own locations, although strict sequence of visits is not feasible due to operational constraints.

\par
Secondly, the agency seeks to reduce the number of miles traveled per visit, which is measured as a separate key performance index (KPI). To achieve this goal, in addition to minimizing caregivers’ departures from and returns to their homes, patients are assigned in a way that minimizes the average and total travel distance for all caregivers. This optimization of patient allocation takes into account the spatial distribution of patients and aims to minimize travel distances. 

Another important objective is clinician consistency, also known as “continuity of care.” The agency recognizes the importance of patients being visited by the same caregiver throughout their service period. If a new service is required during or after a previous one, the agency prioritizes assigning the same caregiver to ensure continuity of care and familiarity between the caregiver and patient. 

Lastly, the agency places significant emphasis on customer satisfaction and aims to ensure a positive overall experience. This encompasses timely visits, alignment of caregiver skills with patient needs, flexibility in visit scheduling for customers, and the continuity of care. By meeting these objectives, the HHA seeks to not only fulfill the needs of patients but also exceed their expectations, resulting in positive recommendations of the agency to others.

\subsection{Problem formulation}
\label{Problem_formulation}
\par

\par
In this case study, the HHC model is reformulated as an integer linear problem (ILP) formulation. The unique aspects of this model are dependent upon the specific assumptions and factors considered in this distinct case study. For instance, similar works, like the one carried out by \cite{pahlevani2022cluster}, formulated their scenario as an HHC Routing and Scheduling Problem (HHCRSP) model. In their model, they incorporated various factors including the costs associated with caregivers, and the earliest and latest times of services, among other parameters. 

\par
Although the present study is not intended to solve the problem with an exact approach, a mixed-integer linear programming (MILP) model is developed to represent the objective and constraints and to understand how the solution complies with it. Unlike the generic model, this MILP has been customized in alignment with the specific assumptions and concerns previously detailed. The contribution lies in the adaptation and customization of the existing model to better suit our specific study. This customization offers a more accurate understanding of the problem at hand, thereby allowing for more effective and efficient solutions. The notations and variables used in our binary ILP model are presented in \textbf{T}able \ref{parameters}. 
%% table1
%% math model

\begin{table}[!h]
    % \centering
    \footnotesize
    \caption{Indices, parameters, and decision variables notation and their definition}
    \makebox[\textwidth]
    {\begin{tabular}{ll}
    
    \hline
    \textbf{Indices} & \textbf{Definition}\\
    \hline
    $S$ & Set of caregivers' skills\\
    $D$ & Set of patients' disciplines\\
    $s$ & Caregivers’ skill, $s \in S$\\ 
    $d$ & Patient’s demanded discipline, $d \in D$\\
    $P_d$ & Set of patients in discipline $d$\\
    $C_s$ & Set of caregivers of skill $s$\\
    
    \hline
    \textbf{Parameters} & \textbf{Definition}\\
    \hline
    
    $\rho_d$ & Nodes indicating locations of patients demanding discipline \emph{d} , $\rho_d  \in P_d$\\ 
    $\mu_j$ & Number of visits per week required for patient \emph{j}\\
    $\zeta_s$ & Nodes indicating locations of caregivers with skill \emph{s}, $\zeta_s \in C_s$ \\
    $N_{sd}$ & Set of all nodes in a certain skill and discipline, $N_{sd} \in \{P_d \bigcup C_s \}$\\
    $d_{ij}$ & Distance between nodes \emph{i} and \emph{j}, $\forall i,j | i \neq j \in N_{sd}$ \\
    $\overline{t_{ij}}$ & Average travel time per unit distance between nodes \emph{i} and \emph{j}, $\forall i,j | i \neq j \in N_{sd}$ \\
    $W_c^{min}$ & Minimum working hour per week for caregiver \emph{c} \\
    $W_c^{max}$ & Maximum working hour per week for caregiver \emph{c} \\
    $l_i$ & Visit length for patient \emph{i} \\
    
    \hline
    \textbf{Variables} & \textbf{Definition}\\
    \hline
    
    $Z_{ic}$ & binary variable, equals to one if patient $i$ is assigned to caregiver $c$ \\ 
    $X_{ijc}$ & binary variable, equals to one if caregiver $c$ is assigned to patient $i$ \\
     
    \hline
    
    \end{tabular}}
    \label{parameters}
\end{table}

\par
Considering the notations of \textbf{T}able \ref{parameters}, the HHC problem can be formulated as follows:

\begin{equation}
    \begin{aligned}
    % \small
    \textnormal{Min} \; Z = \sum_{i \in N_{sd}} \; \sum_{j \in P_d} \; \sum_{c \in C_s} \; d_{ij} X_{ijc}
    \label{objective}
    \end{aligned}
\end{equation}

\par
\qquad subject to:

\begin{equation}
    \begin{aligned}
    \sum_{i \in N_{sd}} \; \sum_{c \in C_s} X_{ijc} = \mu_j     \qquad      \forall j \in P_d
    \label{constraint_1}
    \end{aligned}
\end{equation}

% \begin{equation}
%     \begin{aligned}
    
%     \sum_{i \in P_d} X_{ijc} = Z_{ic}   \qquad      \forall c \in C_s \; , \; \forall j \in P_d 
    
%     \label{constraint_2}
%     \end{aligned}
% \end{equation}

% \begin{equation}
%     \begin{aligned}
    
%     \sum_{i \in N_{sd}} X_{ijc} = \sum_{j \in N_{sd}} X_{jic}   \qquad      \forall c \in C_s \; , \; \forall j \in P_d
    
%     \label{constraint_3}
%     \end{aligned}
% \end{equation}

% \begin{equation}
%     \begin{aligned}
%     \sum_{i \in P_d} Z_{ic}l_i + \sum_{i \in N_{sd}} \; \sum_{j \in P_d} \; d_{ij} X_{ijc} \overline{t_{ij}} \; \geq W_c^{min}    \qquad \forall c \in C_s
%     \label{constraint_4}
%     \end{aligned}
% \end{equation}

% \begin{equation}
%     \begin{aligned}
%     \sum_{i \in P_d} Z_{ic}l_i + \sum_{i \in N_{sd}} \; \sum_{j \in P_d} \; d_{ij} X_{ijc} \overline{t_{ij}} \; \leq W_c^{max}    \qquad \forall c \in C_s
%     \label{constraint_5}
%     \end{aligned}
% \end{equation}

% \begin{equation}
%     \begin{aligned}
    
%     Z_{ic} \; , X_{ijc} \in \{0,1\}     \qquad      \forall i,j \in N_{sd} \; , \; \forall c \in C_s 
%     \label{constraint_6}
%     \end{aligned}
% \end{equation}

\begin{equation}
    \begin{aligned}
    \sum_{i \in P_d} X_{ijc} = Z_{ic}   \qquad      \forall c \in C_s, \; \forall j \in P_d 
    \end{aligned}
    \label{constraint_2}
\end{equation}

\begin{equation}
    \begin{aligned}
    \sum_{i \in N_{sd}} X_{ijc} = \sum_{j \in N_{sd}} X_{jic}   \qquad      \forall c \in C_s, \; \forall j \in P_d
    \end{aligned}
    \label{constraint_3}
\end{equation}

\begin{equation}
    \begin{aligned}
    \sum_{i \in P_d} Z_{ic}l_i + \sum_{i \in N_{sd}} \sum_{j \in P_d} d_{ij} X_{ijc} \overline{t_{ij}} \geq W_c^{\text{min}}    \qquad \forall c \in C_s
    \end{aligned}
    \label{constraint_4}
\end{equation}

\begin{equation}
    \begin{aligned}
    \sum_{i \in P_d} Z_{ic}l_i + \sum_{i \in N_{sd}} \sum_{j \in P_d} d_{ij} X_{ijc} \overline{t_{ij}} \leq W_c^{\text{max}}    \qquad \forall c \in C_s
    \end{aligned}
    \label{constraint_5}
\end{equation}

\begin{equation}
    \begin{aligned}
    Z_{ic}, X_{ijc} \in \{0,1\}     \qquad      \forall i,j \in N_{sd}, \; \forall c \in C_s 
    \end{aligned}
    \label{constraint_6}
\end{equation}

\par
In this model, the \textbf{O}bjective \ref{objective} is the minimization of the distance traveled by all caregivers. Constraint \ref{constraint_1} is to ensure all patients have their visits completed during the week. Constraint \ref{constraint_2} guarantees each patient is visited by just one caregiver. Constraint \ref{constraint_3} is for node flow balance for each caregiver. Constraint \ref{constraint_4} and \ref{constraint_5} and are the constraints for the working times of each caregiver during the week. Constraint \ref{constraint_6} is the binary condition of decision variables.
The presented model is Np-Hard \citep{rasmussen2012home}. According to the average number of demands for the HHA, which in our case is about 5,000 patients for a time window of nine months, the total number of our variable’s permutation would be striking high. Following this, achieving the solution through solving a mathematical model for the real-size problem would be time-consuming, and as a consequence, utilizing it would be no longer reasonable.

\section{Refined allocation approach}
\label{Decision_support_framework}

\par

The main goal of this study is to improve the decision-making process for HHA managers and planners by addressing their concerns and aspirations. Many previous HHC studies treat caregivers more like machines in discrete processes or WIPs on an assembly line rather than human beings. However, this study proposes a new paradigm that emphasizes the importance of employees by giving them more autonomy over their personal preferences.

\par
The proposed framework aims to provide HHAs with a comprehensive tool for making two crucial decisions. Firstly, it facilitates the allocation of caregivers by considering factors such as their location and the specific needs of patients. This approach emphasizes the importance of ensuring continuity of care across different services, with the objectives of reducing costs and maximizing the quality of care provided. Secondly, the framework addresses the complex task of managing employee recruitment and layoffs, allowing manipulation of the available workforce to determine the most efficient allocation of resources. To address these challenges, the framework adopts a clustering approach as the underlying methodology. This selection is based on the recognition that clustering techniques offer valuable insights and solutions in the context of HHC resource allocation.

\subsection{Clustering method}
\par

The purpose of clustering methods in data mining is to divide a set into subsets by considering their similarities in predefined metrics. Geolocation data clustering is generally meant to associate occurrence patterns with goods and services movements. In this research, a clustering scheme is employed in a multi-step method to map patients’ locations to caregivers’ locations and provide an assignment baseline for planners regarding the workload balance between all caregivers of a certain type for the long term.

\par
There are a few well-known clustering methods in the literature used for geolocation data clustering in different disciplines, including HHC. Some of the most applied methods are k-means clustering, affinity propagation, self-organizing map, DBSACN (density-based spatial clustering of applications with noise), and HDBSCAN (Hierarchical DBSACN) \citep{quintana2017clustering,riazi2018column,belhor2023multi}. Although any of these methods can be used to categorize data points into nearby groups, not all of them result in a proper and rational classification for home health applications. Many of these methods yield excessive noise points together with highly unequal clusters, which are not capable of assigning patients. Besides, in some methods, such as HDBSCAN, the number of clusters is not an input parameter. Thus, it is not possible to specify a certain number of caregivers as the centroid of each cluster \citep{campello2013density}. 

\par
\begin{figure}[!h]
\centering
\begin{subfigure}{.8\textwidth}
  % \raisebox{0.2cm}{%
    \centering
    \includegraphics[width=\textwidth]{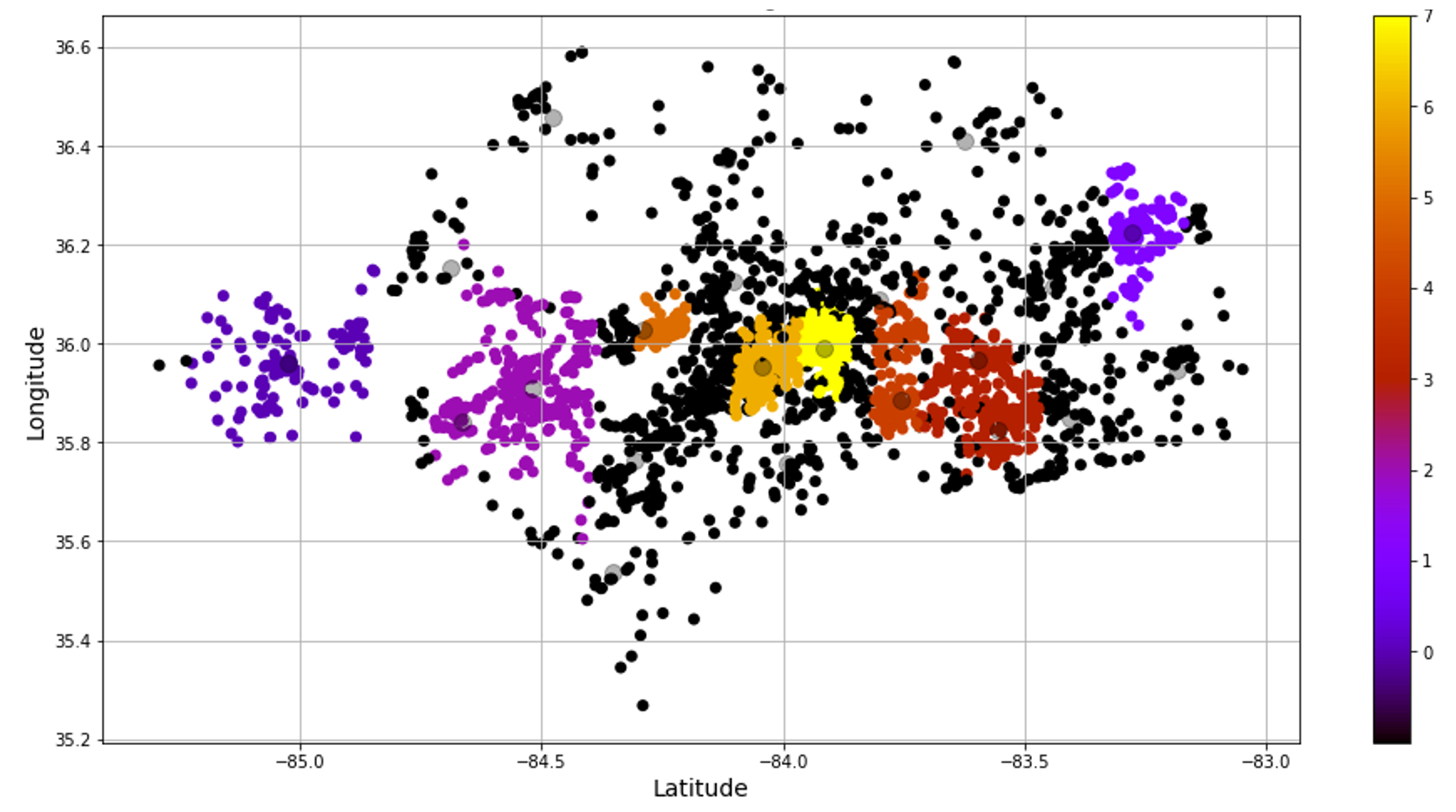}
    \caption{Mapping of RN caregivers allocated patients}
    \label{fig-5-a}
\end{subfigure}
\begin{subfigure}[]{0.8\textwidth}
  \centering
  \includegraphics[width=\textwidth]{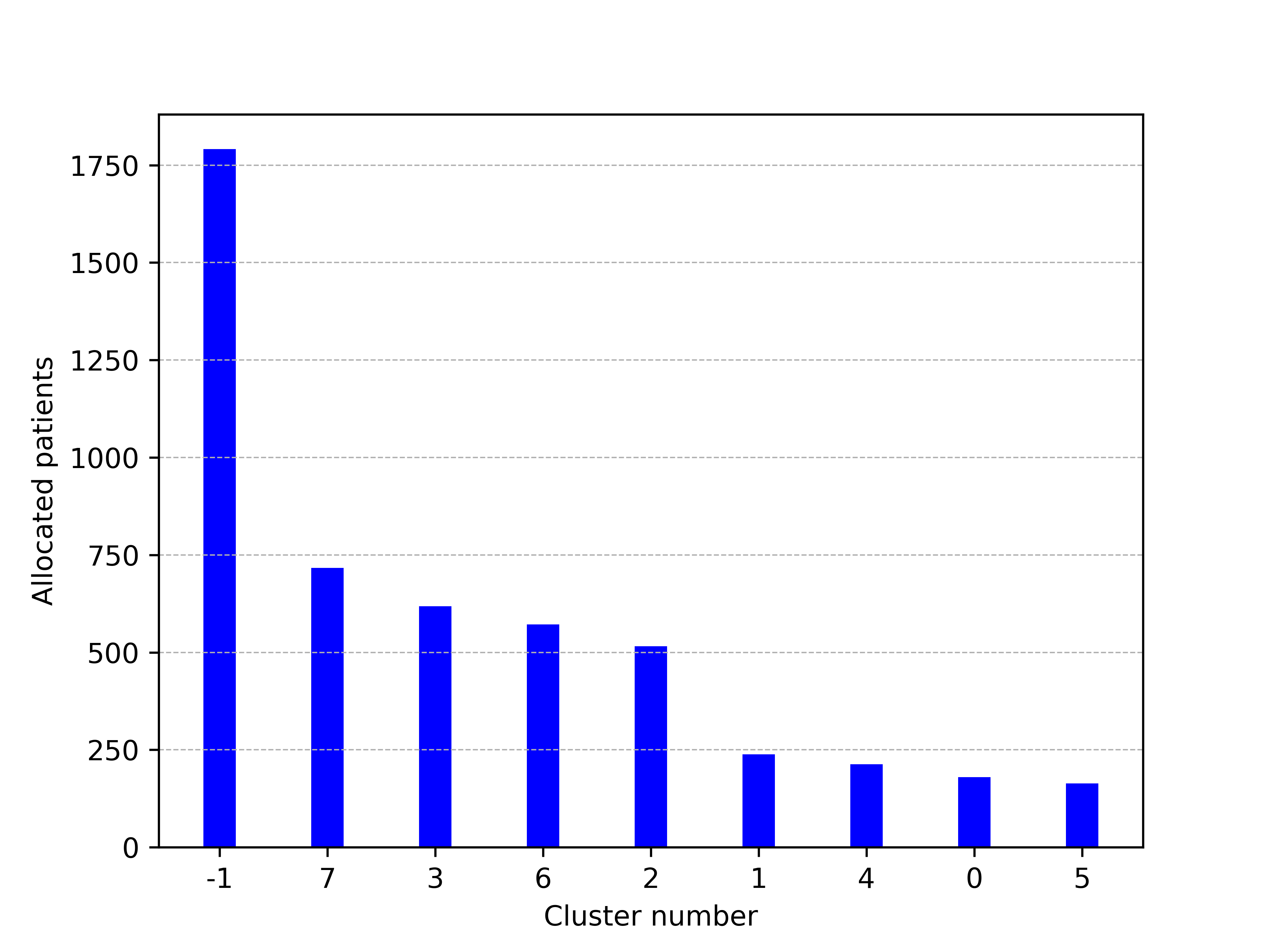}
  \caption{Distribution of allocated patients to caregivers}
  \label{fig-5-b}
\end{subfigure}

\caption{Clustering of RN data using HDBSCAN}
\label{fig_5}
\end{figure}

In this study, we investigated the test results of several clustering algorithms, including the aforementioned methods, to determine which one is most suitable for integration into the proposed framework. To illustrate the strengths and weaknesses of these algorithms, we discuss two of them in detail in this section: HDBSCAN and k-means. HDBSCAN is a density-based clustering algorithm that is well-suited for clusters with arbitrary shapes and sizes. However, it can be computationally expensive for large datasets. K-means is a simpler algorithm that is more efficient for large datasets, but it is not as effective at clustering data with non-convex shapes. \textbf{F}igure \ref{fig_5} shows the results of these algorithms for a randomly selected set of locations.

\par 
HDBSCAN is an extended version of DBSCSN that tries to specify the most significant clusters regarding levelized density thresholds \citep{campello2013density}. The results of HDBSCAN in \textbf{F}igure \ref{fig_5} show the number of clusters is not close to the desired number in that it is not an input parameter of this method. In this case, the desired number is equivalent to the number of available caregivers. Another drawback is that this method labels a large portion of the locations as $noise$ data (black locations with the cluster number $-1$ ). While it is possible to use heuristic algorithms to force HDBSCAN to make the desired number of clusters, a significant portion of locations labeled as noise will still remain a major problem. The advantage of this method is that the locations that are not detected as noise are relatively distributed in the clusters as desired (cluster numbers 0 to 7 on the top-right chart of \textbf{F}igure \ref{fig_5}).

\begin{figure}[!h]
\centering
\begin{subfigure}{.8\textwidth}
  % \raisebox{0.2cm}{%
    \centering
    \includegraphics[width=\textwidth]{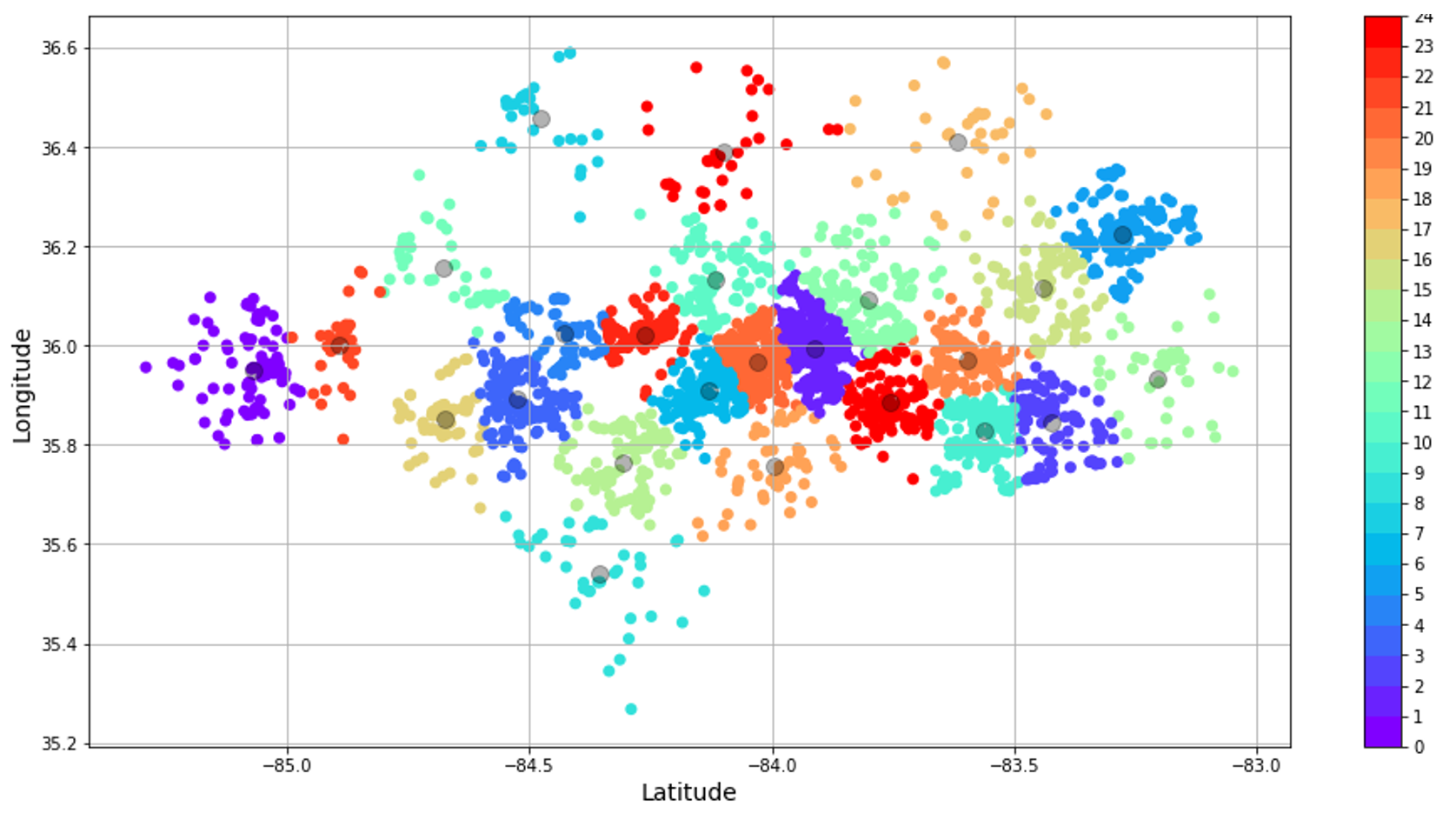}
    \caption{Mapping of RN caregivers allocated patients}
    \label{fig-5-1-a}
\end{subfigure}
\begin{subfigure}[]{0.8\textwidth}
  \centering
  \includegraphics[width=\textwidth]{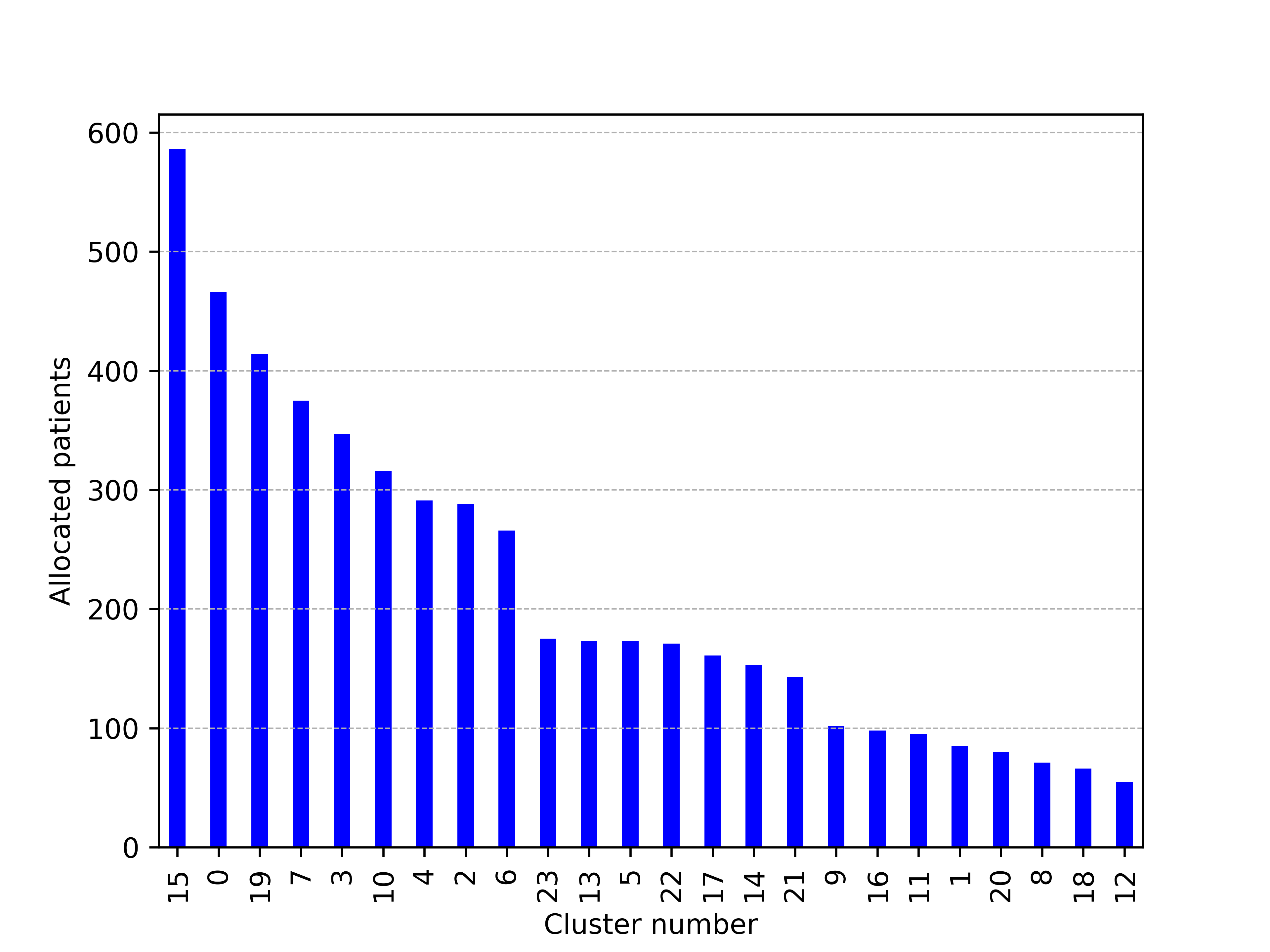}
  \caption{Distribution of allocated patients to caregivers}
  \label{fig-5-1-b}
\end{subfigure}

\caption{Clustering of RN data using K-Means}
\label{fig_5_1}
\end{figure}

Another method that can be explored in this study is the application of \textit{K-means}, an unsupervised centroid-based algorithm that aims to minimize the distances between data points and a predetermined number of central points. While the capability of \textit{K-means} to classify geolocation data into a specific number of clusters is advantageous for home health assignment purposes, as depicted in \textbf{F}igure \ref{fig_5_1}, it tends to create relatively large clusters in densely populated regions like urban districts. This poses a challenge in terms of assigning a single caregiver to such large clusters, making it impractical in practice.

\par
To tackle the aforementioned errors, another method named Spectral Clustering \citep{von2007tutorial} is employed here. Spectral clustering is helpful in cases where the topology of each cluster is non-convex. In these cases, measuring the centroid and distribution of data points cannot properly describe each cluster. In geolocation data clustering, Spectral clustering creates the affinity matrix of the graph by using either a \emph{k-nearest} neighbors (aka \emph{K-NN}) matrix or another kernel function known as RBF (radial basis function), in which the Gaussian kernel $k(X,X)$ is calculated using the Euclidean distance matrix $d(X,X)$ as:

\begin{equation}
    \begin{aligned}
    k(X, X) = \exp (- \psi \times d(X, X))
    \label{gaussian_kernel}
    \end{aligned}
\end{equation}

\par
Where $\psi$ is the kernel hyperparameter that is inversely proportional to the dispersion level of data points \citep{damle2019simple}.

\begin{figure}[!h]
\centering
\begin{subfigure}{.8\textwidth}
  % \raisebox{0.2cm}{%
    \centering
    \includegraphics[width=\textwidth]{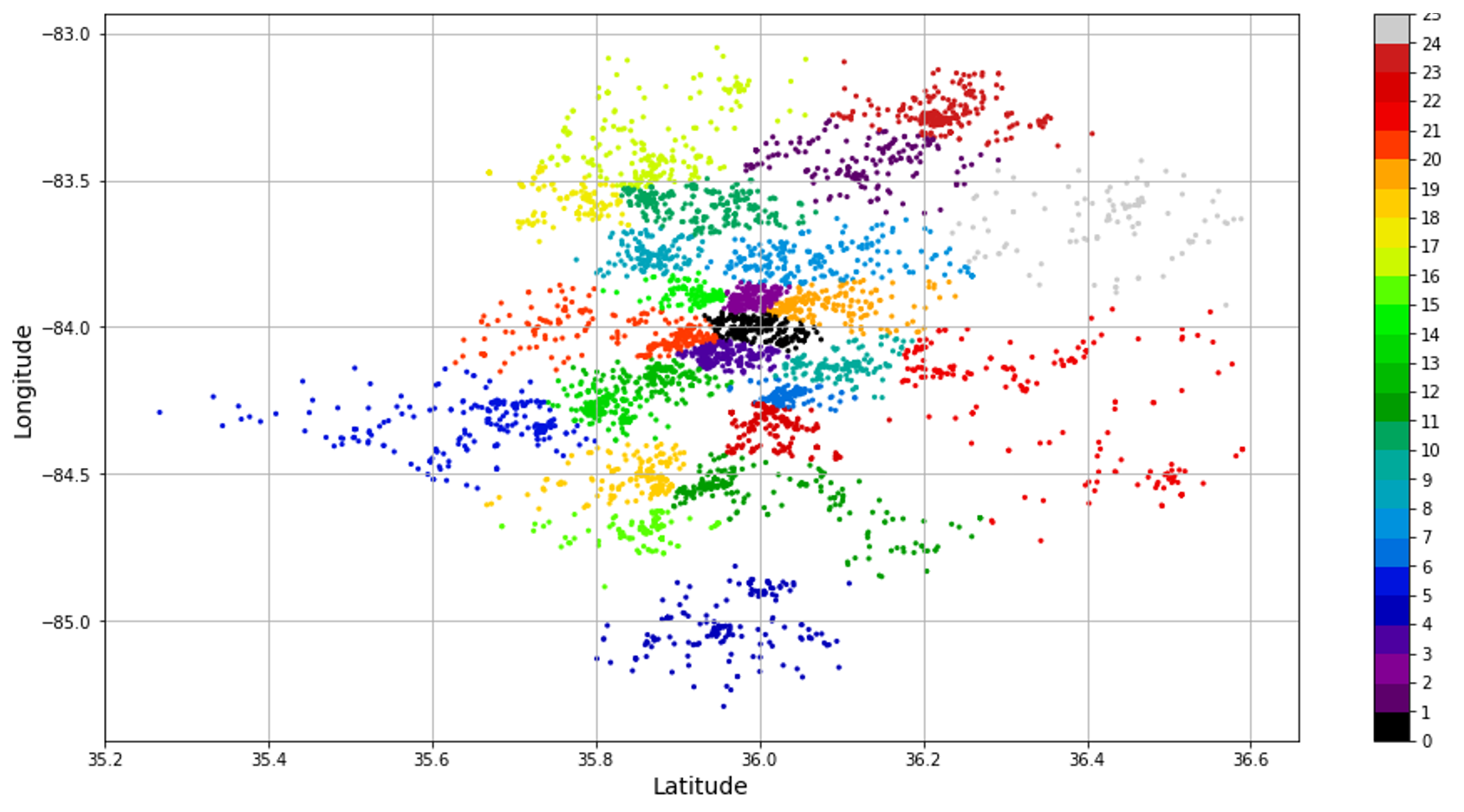}
    \caption{Mapping of RN caregivers allocated patients}
    \label{fig-5-a}
\end{subfigure}
\begin{subfigure}[]{0.8\textwidth}
  \centering
  \includegraphics[width=\textwidth]{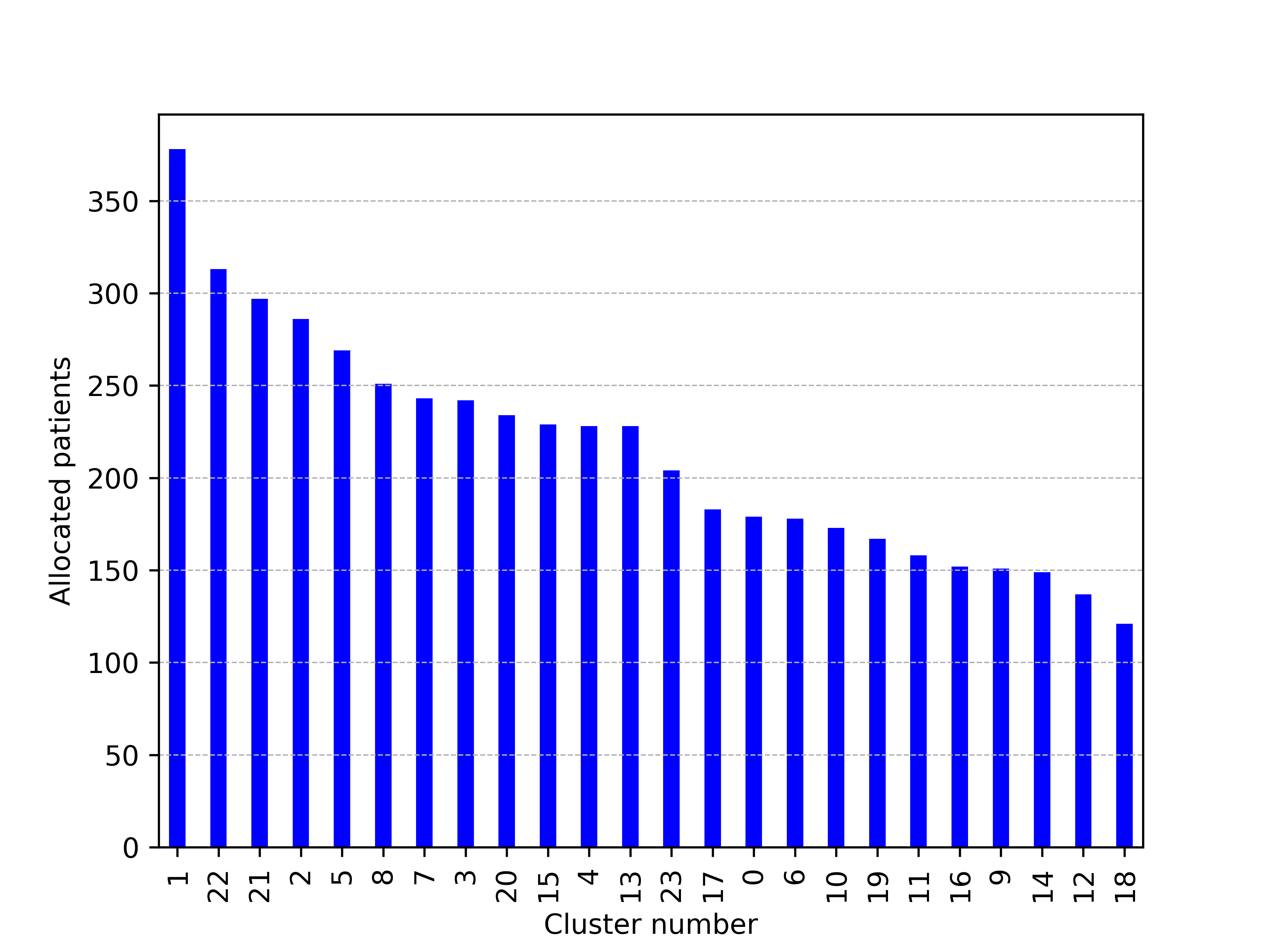}
  \caption{Distribution of allocated patients to caregivers}
  \label{fig-5-b}
\end{subfigure}

\caption{Clustering of RN data using the modified spectral clustering}
\label{fig_6}
\end{figure}

\par
\textbf{F}igure \ref{fig_6} illustrates the results of clustering the sample data points with Spectral clustering. Since this method considers the non-convexity of the desired clusters and the relative proportion of the distances between points in each cluster, the population distribution of the resulting clusters is proportional to their distances, meaning that if the data points are more dispersed, the number of data points corresponding to that cluster would be lower. This property makes the Spectral method appropriate for home health applications. Since the traveling time of the caregiver is a factor of their workload, assigning of fewer patients to a caregiver that covers a dispersed area is desirable.

\par
To evaluate the performance of clustering techniques, various measures have been introduced in the literature. Among them, two popular ones are the \textit{Calinski–Harabasz} \citep{calinski1974dendrite} and the \textit{Davies–Bouldin} \citep{davies1979cluster} indexes. These measures have gained significant recognition and are widely used for assessing the effectiveness of clustering algorithms. The \textit{Calinski–Harabasz} evaluates the quality of clusters by assessing their centroids’ dispersion and within-cluster distances. In our problem, this index is calculated for each discipline $d$ as a score $S_{CH}^d$ using the following equation:

\begin{equation}
    \begin{aligned}
    S_{CH}^d = \left( \frac{\sum_{k \in CL_d} |CL_d^k| d_{ic_n^k}}{|CL_d|-1} \right) \left( \frac{|D| - |CL_d|}{\sum_{i \in CL_d^k} \; \sum_{j \neq i \in CL_d^k} d_{ij}} \right)
    \label{CH_index}
    \end{aligned}
\end{equation}

Where $CL_d = \{ CL_d^1, CL_d^2, \cdots, CL_d^n \}$ is the set of all clusters of patients in discipline $d$, thus the size of this set, $|CL_d|$, is equal to the number of eligible caregivers for discipline $d$. $k$ is the cluster indicator $(k \in CL_d)$, $c_n^k$ is the centroid of cluster $k$, that is the location of the caregiver assigned to that cluster. Thus, $d_{ic_n^k}$ is the distance between this centroid and any location $i \in CL_d^k$.
Finally, $|D|$ is the size of the set of all patients demanding discipline $d$.
Another performance measure, which is the Davies–Bouldin Index, calculates clustering scores by evaluating their similarities while maximizing the distances between cluster centroids. To calculate this index as a score for each discipline $d$, a similarity measure $R_{}$ for all , is first defined as follows:

\begin{equation}
    \begin{aligned}
    R_{uv} = \frac{\sum_{i \in u} \frac{d_{ic_n^u}}{|u|} + \sum_{i \in v} \frac{d_{ic_n^v}}{|v|}}{d_{uv}}  \qquad  u \neq v \in CL_d
    \label{DB_index_R}
    \end{aligned}
\end{equation}

Where $|u|$ and $|v|$ are respectively the size of clusters $u$ and $v$, and $d_{uv}$ is the distance between their centroids. This similarity measure actually trades off the average distance between each point in each cluster and their centroids in one hand, and the distance between their centroids in another. The Davies–Bouldin Index is then is calculated as follows:

\begin{equation}
    \begin{aligned}
    S_{DB}^d = \frac{1}{|CL_d|} \sum_{u \in CL_d} \max_{u \neq v} R_{uv}
    \label{DB_index}
    \end{aligned}
\end{equation}

\par
Although both measures give practical results in many geolocation clustering cases, there are two drawbacks that make them irrelevant in this case. First, both bring into account the centroids’ dispersion, while we assume the centroids are pre-defined by caregivers’ locations. Second, they prioritize the average distance between cluster centroids and their members when scoring inter-cluster dispersion \citep{halkidi2001clustering}. As mentioned earlier, here, the caregivers are mostly traveling between patients’ locations rather than from/to their own locations (centroids), thus it is better to bring into account the distances between all members of each cluster. 

To address these issues in our problem, we proposed two novel performance measures, which are the Average of the Mean Pairwise Mileage (AMPM) within clusters  and the Average of the Total Pairwise Mileage(ATPM) within clusters. These two measures are computed for each discipline $d$. These measures are computed by \textbf{E}quations \ref{AMPM} and \ref{ATPM}.

% \begin{equation}
%     \begin{aligned}
%     AMPM_d = \\
%     \frac{1}{|CL_d|} \left[ \sum_{k \in CL_d} \left( \gamma_k  \sum_{i \in CL_d^k} \frac{d_{ic_n^k}}{|CL_d^k|}
%     +(1+\gamma_k) \sum_{i \in CL_d^k} \; \sum_{j \neq i \in CL_d^k} \frac{d_{ij}}{|CL_d^k|(|CL_d^k|-1)} \right) \right]
%     \label{AMPM}
%     \end{aligned}
% \end{equation}

% \begin{equation}
%     \begin{aligned}
%     ATPM_d = \frac{1}{|CL_d|} \left[ \sum_{k \in CL_d} \left( \gamma_k \sum_{i \in CL_d^k} d_{ic_n^k} + 
%     (1+\gamma_k) \sum_{i \in CL_d^k} \; \sum_{j \neq i \in CL_d^k} d_{ij} \right) \right]
%     \label{ATPM}
%     \end{aligned}
% \end{equation}

{\scriptsize
\begin{flalign}
    &AMPM_d = 
    \frac{1}{|CL_d|} \left[ \sum_{k \in CL_d} \left( \gamma_k  \sum_{i \in CL_d^k} \frac{d_{ic_n^k}}{|CL_d^k|}
    +(1+\gamma_k) \sum_{i \in CL_d^k} \; \sum_{j \neq i \in CL_d^k} \frac{d_{ij}}{|CL_d^k|(|CL_d^k|-1)} \right) \right]&
    \label{AMPM}
\end{flalign}
}

{
\small
\begin{flalign}
    &ATPM_d = \frac{1}{|CL_d|} \left[ \sum_{k \in CL_d} \left( \gamma_k \sum_{i \in CL_d^k} d_{ic_n^k} + 
    (1+\gamma_k) \sum_{i \in CL_d^k} \; \sum_{j \neq i \in CL_d^k} d_{ij} \right) \right]&
    \label{ATPM}
\end{flalign}
}

\par
Where $\gamma_k$ is defined as a coefficient of visitation consistency, which shows the ratio of the caregivers’ “from/to caregiver’s home” travels to their total number of travels. As we explained in \textbf{S}ection \ref{Problem statement} and depicted in \textbf{F}igures \ref{fig_3}, although it is desired to lower this coefficient as much as possible, it is not practical to impose a strict certain level of  $\gamma$ to the caregivers. Thus, to avoid putting extra pressure on the caregivers and to give them enough flexibility in sequencing their visitations, values of $\gamma$ are calculated using the historical data and considered as input parameters for the problem. 

\par
The interpretation of \textit{$AMPM_d$} is more straightforward when we note that the first term in the parentheses of \textbf{E}quation \ref{AMPM} calculates the expected average mileage that caregivers travel on “from/to caregiver’s home” routes and the second term is the expected “between patient visits” mileage. In fact, \textit{$AMPM_d$} and \textit{$ATPM_d$} are evaluating the average and total distances between all locations in all clusters, respectively. Hence, these measures give more realistic criteria in terms of the expected mileage each caregiver needs to travel, giving all possible visitation sequences. In this study, clustering is assigned based on the minimization of \textit{$AMPM_d$}. Nevertheless, to assess the quality of solutions, \textit{$ATPM_d$} is calculated to be able to compare improved scenarios with the historical data of total travel mileage.

\subsection{Genetic optimization of clustering}
\label{Genetic_optimization_of_clustering}

To integrate the clustering hyperparameter optimization into the framework, we employed the Genetic algorithm (GA) to solve for an integer optimization model in which the integer variables correspond to the set of clustering algorithm's hyperparameters (i.e., spectral clustering), with the objective function of $AMPM$. GA is a type of evolutionary algorithm and a branch of artificial intelligence inspired by the process of natural selection. It is used to solve optimization and search problems by mimicking the principles of biological evolution. A genetic algorithm operates through a cycle of selection, crossover (recombination), and mutation on a population of individuals, each representing a possible solution to the problem at hand. These individuals are encoded as strings, often resembling chromosomes \citet{RN69}. A generalized formulation of the hyperparameter optimization model is presented in \textbf{M}odel \ref{hp_opt}.

\begin{equation}
\begin{aligned}
& \underset{\theta}{\text{Minimize}}
& & \textit{AMPM}(\theta,x) \\
& \text{subject to}
& & C_i^E (\theta,x), \quad i=1,\ldots,n \\
& & & C_j^I (\theta,x), \quad j=1,\ldots,m \\
\label{hp_opt}
\end{aligned}
\end{equation}

In this model, $\theta$ is the set of clustering hyperparameters, $x$ is the geolocation setpoints,
$AMPM(\theta,x)$ is the objective function, $C_i^E (\theta,x) \in C^E$ are the set of equality constraints (i.e. constraint \ref{constraint_1}, \ref{constraint_2} and \ref{constraint_3}), and $C_j^I (\theta,x) \in C^I$ are set of inequality constraints (i.e. constraints \ref{constraint_4} and \ref{constraint_5}). The goal is to minimize the objective $AMPM$ by selecting the best combination of spectral hyperparameters. Note that this kind of improvement can also be performed for other clustering methods rather than spectral clustering described here by customizing the appropriate set of hyperparameters and their possible values.

\subsection{Decision support framework}
\label{Decision_support_framework}
\par
\begin{figure}[!h]
    \centering
    \includegraphics[width=1.25\textwidth, angle = 90]{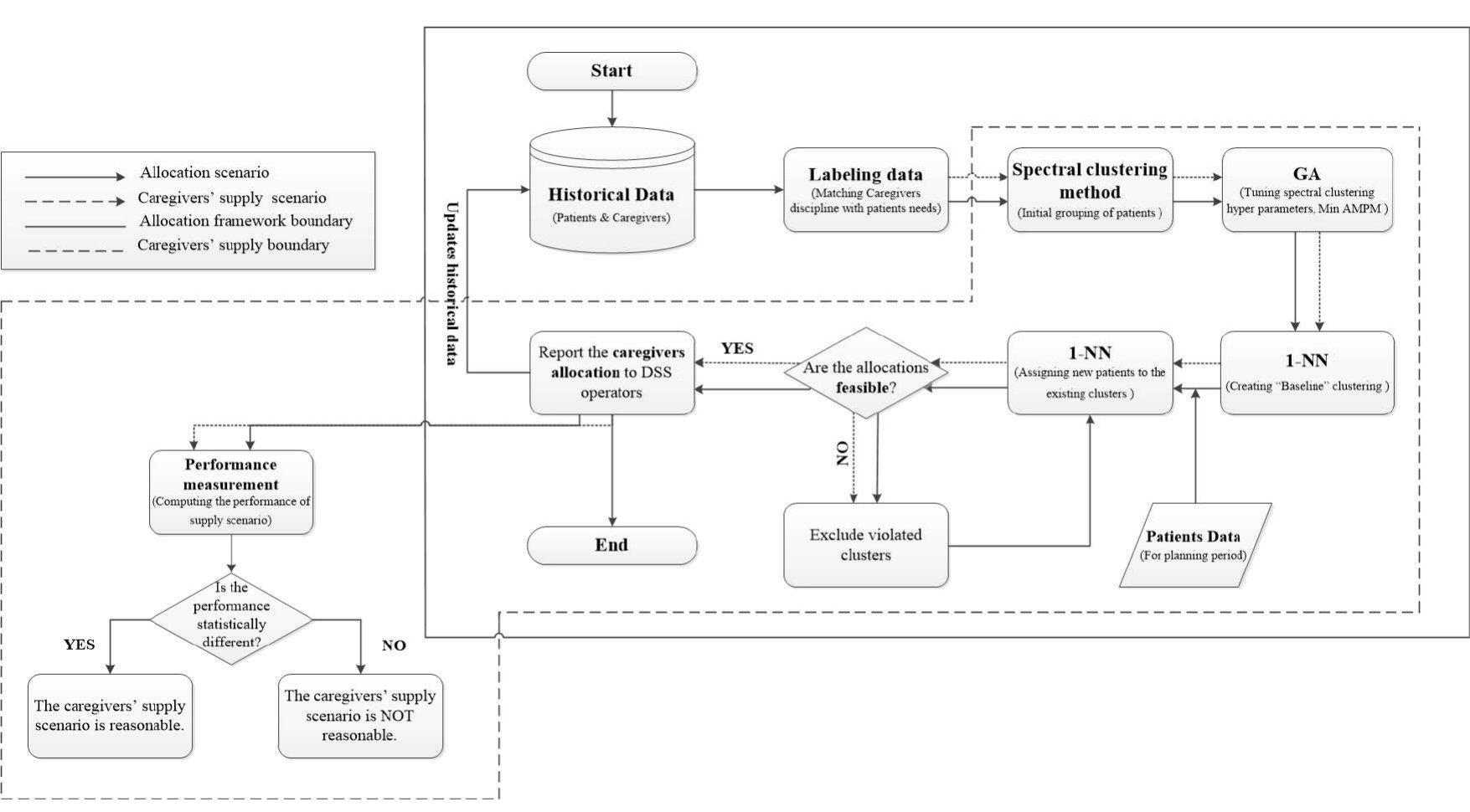}
    \caption{Flowchart of the background approach used in the proposed decision support framework.}
    \label{fig_7}
\end{figure}

After selecting the proper clustering method and defining the performance metric for the problem, we construct the required steps of the proposed allocation framework. This framework is presented as a flowchart in \textbf{F}igure \ref{fig_7}. The main scheme of the framework consists of the following steps. A more detailed description of the process is provided in \textbf{S}ection \ref{Computational_Analysis}.

\par
\textit{Importing historical data:} In the first step, historical data about the locations of the patients and caregivers and the dates of visits are cleansed. Then they are labeled according to their corresponding required skills and disciplines. To prepare these data in a proper format for processing, we need the exact coordinates, latitudes, and longitudes of the location addresses. These locations are derived from a governmental database.

\textit{Patients clustering:} The coordinates of the labeled data are given to the spectral clustering algorithm with an initial setting for hyperparameters that returns a reasonable cluster of patients in terms of the size of each cluster and their relative sparseness so that it provides an initial scheme for allocating the patients to caregivers.

\textit{Clustering tunning:} For tuning the hyperparameters of the spectral clustering, GA is employed as our metaheuristic optimization method. For each discipline, assumable ranges and values of clustering hyperparameters are determined based on the initial algorithm settings in the previous step. These ranges and values are used as the inputs for specifying the variable constraint in GA. The optimization process is performed in GA iterations to minimize AMPM and ATPM.

\textit{Allocating caregivers to clusters:} The tunned clusters are used to create the “baseline” allocation. Since in spectral clustering, the location of the centroids cannot be determined as an input parameter, to create the baseline, we need to locate the nearest centroid of each cluster. To do so, we perform a \textit{k-NN} clustering while setting $K=1$, which results in the unique nearest centroid of each caregiver’s location.

\textit{Allocating patients to clusters:} When the baseline clustering is ready, it is employed to allocate patients in each planning period to the caregivers of the corresponding discipline. To determine the cluster of each patient, the nearest clustered location of that patient in the baseline is determined by performing a \textit{1-NN} clustering. Thus, the new patient is allocated to the caregiver of its corresponding nearest neighbor’s cluster.

\textit{Allocation feasibility check:} To check the feasibility of the resulting allocation, the constraint of \textbf{E}quations \ref{constraint_4} and \ref{constraint_5} regarding the minimum and maximum working hours of each caregiver in the planning period must be checked. The logic of the clustering approach guarantees that all other constraints in the mathematical model (\textbf{E}quations \ref{constraint_2}, \ref{constraint_3}, \ref{constraint_6}) are automatically satisfied. If the allocation is feasible, it is referred to by the planning operator as the DSS suggestion. If the allocations are not feasible, caregivers of the clusters that correspond to the violated constraints will be excluded from the baseline allocation, and the patient allocation process will repeat.

\textit{Caregivers' supply analysis:} This part functions as the add-on to the main framework to provide insights about recruiting or laying off caregivers to the managers, in order to optimally allocate the company's resources. By defining alternative scenarios regarding the number of caregivers available, we can test the changes in the performance of allocation and the service level of the company. More details and sample results are provided in \textbf{S}ection \ref{Sensitivity_analysis}. 

\section{Computational Analysis}
\label{Computational_Analysis}
To implement the framework for the HHA case study, some assumptions needed to be adopted. First, the distances between addresses are calculated by multiplying Euclidean (Haversine) distances to a correction coefficient proposed by \citet{boyaci2021vehicle}, which is approximately equal to 1.285. The second assumption is that for data points (addresses) that the exact location coordinates were not found by searching the US Census references, zip-code centers are replaced and used in the algorithm. Last but not least due to operational obstacles and the fact that visitation sequences cannot be imposed on the caregivers, working normal and extra hours are tuned by the caregivers with the agency's permission.

\par
The scope of our case study encompasses the visiting records spanning a nine-month period, from July $2019$ to March $2020$ including more than $100,000$ visiting records.

The computational analysis of this case study has been conducted in three main stages: data preprocessing, obtaining improved results using the proposed framework, and analysis of the sensitivity of results to changes in caregivers' supply. The data preprocessing stage involves necessary data cleaning, transformation, and preparation for subsequent analysis. Then we use the prepared data as the input of the proposed framework to obtain improved results which provide flexible allocations for the decision-making process. The overall performance of the implemented approach is then evaluated. Lastly, the caregiver supply analysis focuses on assessing different scenarios for the availability and distribution of caregivers and comparing the results derived from these scenarios with the current situation. This analysis offers valuable insights into resource allocation and management within the studied context.

\subsection{Data pre-processing}
\label{Data_pre-processing}

The historical data used in this case study consists of more than 100,000 visiting records over the course of nine months. Each record contains entries specifying the caregiver, patient, length of visits, time of departure, locations of start and end of each departure, discipline, calculated mileage, branch, insurance, agency info, etc.

\par
To be able to use the data in the proposed approach, it was necessary to convert them to the right format and clean them. After this operation, the addresses queried from the HHA database are imported to the \textit{geo-coder} service provided by the United States Census Bureau \footnote{https://www.census.gov/data/developers/data-sets/Geocoding-services.html}. This geo-coding service returns the exact coordinates of longitude and latitude for most of the addresses. For around 20\% of the data, the coordinates are not available in the Census Bureau’s database. these coordinates are estimated by their zip code centers. Due to the relatively small area that each zip code encompasses, this estimation is considered reasonable. It also prevents the exclusion of valuable data in the cleansing stage.

\par
The prepared data is then split into two parts. The first part includes the first 6 months and is used as the historical data to prepare the baseline allocations via tuned clustering described in Section \ref{Decision_support_framework}. The second part, including the remaining 3 months, is used as the test scenarios in weekly planning periods through the $1-NN$ allocation and feasibility check process. In each week of the test, the validity and performance of the baseline allocation are evaluated.
As discussed in Section \ref{Decision_support_framework}, $\gamma$ ratios are calculated using the training data and considered as given input parameters for the problem. In this case study, we can derive the current values of $\gamma$ from the historical data for each discipline. This value is called $\gamma_d^{curr}$ and is calculated by the following ratio:

\begin{equation}
    \begin{aligned}
    \gamma_d^{curr} = \frac{N_d^{home}}{N_d^{total}} =  \frac{\sum_{k \in CL_d} \sum_{i \in CL_d^k} \; X_{ic_n^k} + \sum_{k \in CL_d} \sum_{i \in CL_d^k} \; X_{c_n^k i}}
    {\sum_{k \in CL_d} \sum_{i \in CL_d^k} \sum_{j \neq i \in CL_d^k} \; X_{ijk}}
    \label{gamma_curr}
    \end{aligned}
\end{equation}

Where $N_d^{total}$ is the total number of travels, and $N_d^{home}$ is the number of “from/to caregivers’ home” travels in a certain discipline $d$. Remind that $X_{ijk}$ is the binary variable indicating the traveling of caregiver $k$ from location $i$ to $j$. 
We argued in Section \ref{Decision_support_framework} that to provide a flexible allocation of patients to caregivers, this $\gamma_d^{curr}$ is used as an average in \textbf{E}quations \ref{AMPM} and \ref{ATPM}. Nevertheless, as mentioned in Sub-section \ref{HHC agency's concerns}, it is desired for the agency to reduce this ratio. Therefore, we can assume that caregivers would be able to reduce their home travels if they had more freedom in their visitation scheduling during a period. As a result, this restricted ratio called $\gamma_d^{lim}$, is arbitrarily assumed to be $\%20$ less than  $\gamma_d^{curr}$, and can be used in clustering evaluation in \textbf{E}quations \ref{AMPM} and \ref{ATPM}. Table \ref{Gamma_table} shows these current and limited values of $\gamma$ ratio for each discipline.

\begin{table}[ht]
\centering
\caption{Current and limited values of $\gamma$ ratio for each discipline.}
\label{Gamma_table}
\small
\scalebox{0.9}{
\begin{tabular}{lllll}
\toprule
\textit{Discipline} & \textbf{$N^{total}$} & \textbf{$N^{home}$} & \textbf{$\gamma^{curr}$} & \textbf{$\gamma^{lim}$} \\ 
\midrule
BSW  & 2332  & 944   & 0.40 & 0.32 \\ 
CH   & 3635  & 1545  & 0.43 & 0.34 \\ 
CNA  & 12554 & 4548  & 0.36 & 0.29 \\ 
COTA & 1578  & 554   & 0.35 & 0.28 \\ 
LPN  & 4983  & 1845  & 0.37 & 0.30 \\ 
MSW  & 3641  & 1579  & 0.43 & 0.34 \\ 
OT   & 6372  & 3133  & 0.49 & 0.39 \\ 
PT   & 21488 & 9655  & 0.45 & 0.36 \\ 
PTA  & 10922 & 4304  & 0.39 & 0.31 \\ 
RN   & 56360 & 23505 & 0.42 & 0.34 \\ 
SLP  & 846   & 476   & 0.56 & 0.45 \\ 
\bottomrule
\end{tabular}
}
\end{table}

\subsection{Case study results}
\label{Results_for_improved_clustering}

\par

As described in Section \ref{Genetic_optimization_of_clustering}, the proposed allocation framework uses a hybrid clustering approach that goes through a hyperparameter tuning process to improve its results regarding the AMPM and ATPM metrics. This improvement process uses GA. The parameters of the GA itself are selected through an exhaustive search. \textbf{T}able \ref{ga_parameters} presents these parameters used for the case study data.

\begin{table}[ht]
\centering
\caption{Genetic Algorithm Parameters}
\label{ga_parameters}
\small
\scalebox{0.9}{
\begin{tabular}{lp{8cm}l}
\toprule
\textit{GA Parameter} & \textit{Description} & \textit{Value} \\
\midrule
Population Size & Number of individuals in each generation & 40 \\

Selection Mechanism &  Determines how individuals are chosen for reproduction & roulette wheel \\

Crossover Rate & Controls how often pairs of individuals undergo crossover to produce offspring & 0.5\\

Mutation Rate & Probability of an allele (a part of an individual's chromosome) changing & 0.1 \\

Termination Criteria & Defines when the algorithm should stop & 100 max iteration \\

\bottomrule
\end{tabular}
}
\end{table}

% which is a well-known metaheuristic optimization algorithm. GA is a random-based classical evolutionary method, which creates a population of different random-selected chromosomes (features) at each iteration and passes the best-fit solutions into the next iterations using two evolutionary processes known as “crossover” and “mutation” \citep{RN69}. 
\par
The decision variables of GA in this study are the spectral clustering hyperparameters, namely, the strategy for eigenvalue decomposition, size of eigenvectors used for the embedding, number of iterations of running the k-means algorithm with replaced centroid seeds, type of the kernel function to create the affinity matrix, kernel coefficient, number of neighbors to use by the nearest-neighbors kernel to construct the affinity matrix, termination criteria for eigenvalue decomposition, and the specialized kernel parameters. \textbf{T}able \ref{spectral_params} shows the optimized hyperparameters of the modified spectral clustering selected by GA. To maintain brevity, the description of these parameters will not be discussed here, but an interested reader can find out about them in detail in \cite{von2007tutorial}.

\begin{table}[ht]
\centering
\caption{Optimized hyperparameters of the modified spectral clustering}
\label{spectral_params}
\small
\scalebox{0.9}{
\begin{tabular}{ll}
\toprule
\textit{Hyperparameter} & \textit{Value} \\
\midrule
Strategy for Eigenvalue Decomposition & Algebraic multigrid methods \\

Size of embedding Eigenvectors & Number of desired clusters\\

Number of K-Means iterations with replaced centroid seeds & 10\\

Type of kernel function to create the affinity matrix & radial basis function \\

Coefficient for the kernel function & 1.0 \\

Number of neighbors used by nearest-neighbors kernel & 10 * Number of desired clusters\\

Termination criteria for Eigen decomposition & 100 max iteration \\

\bottomrule
\end{tabular}
}
\end{table}

\par
\begin{figure}[]
\centering
\begin{subfigure}{.4\textwidth}
\raisebox{0.2cm}{%
  \centering
  \includegraphics[width=\textwidth, height=5cm]{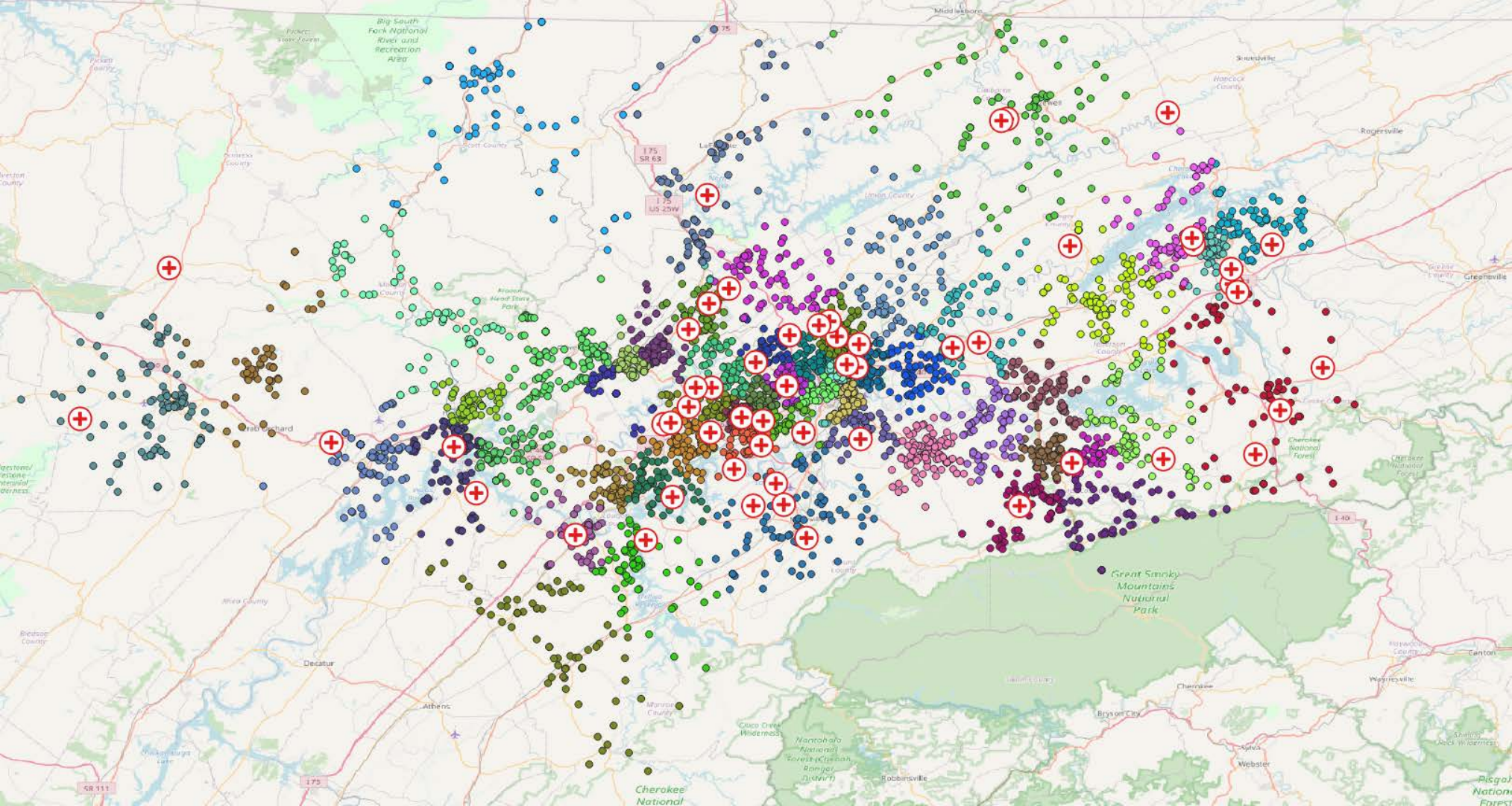}}
  \caption{Mapping of RN caregivers allocated patients}
  \label{fig:sub1_rn}
\end{subfigure}
\begin{subfigure}[b]{.23\textwidth}
  \centering
  \includegraphics[width=\textwidth]{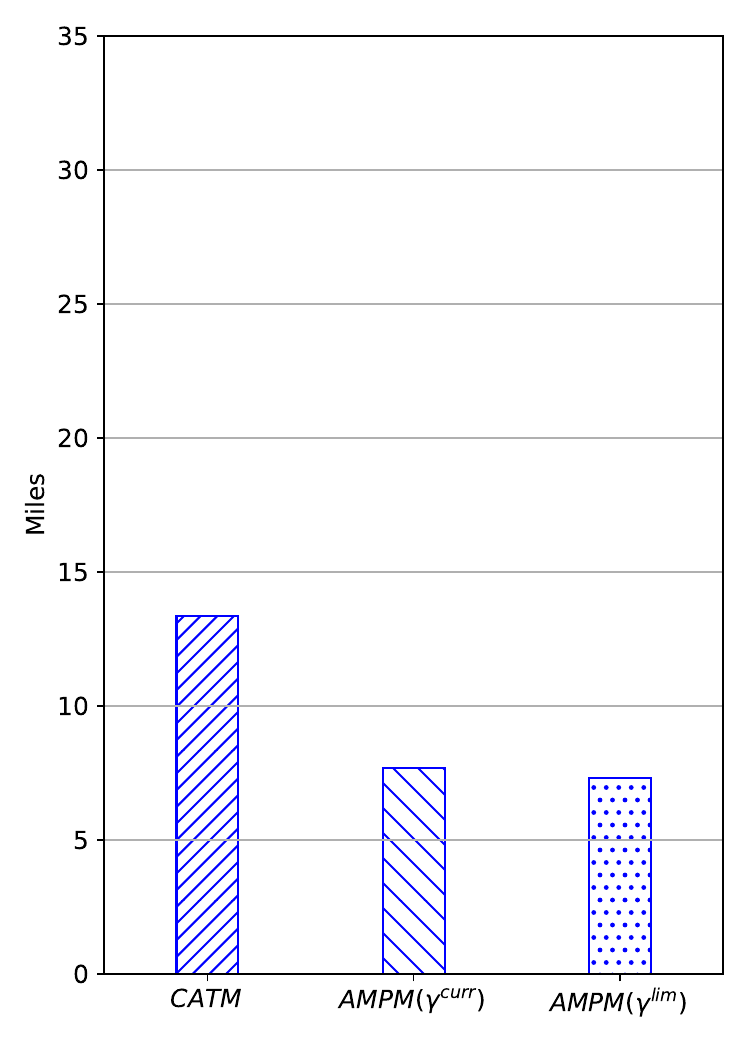}
  % \caption{AMPM}
  \label{fig:sub2_rn}
\end{subfigure}
\begin{subfigure}[b]{.23\textwidth}
  \centering
  \includegraphics[width=\textwidth]{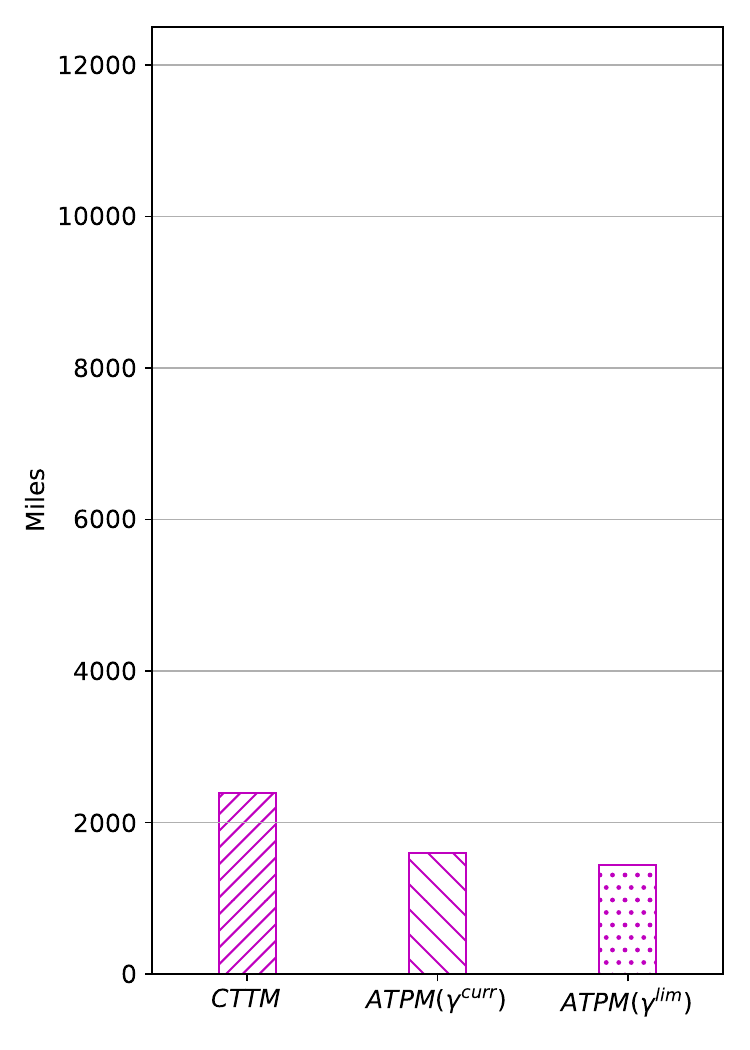}
  % \caption{ATPM}
  \label{fig:sub3_rn}
\end{subfigure}

\begin{subfigure}{.4\textwidth}
\raisebox{0.2cm}{%
  \centering
  \includegraphics[width=\textwidth, height=5cm]{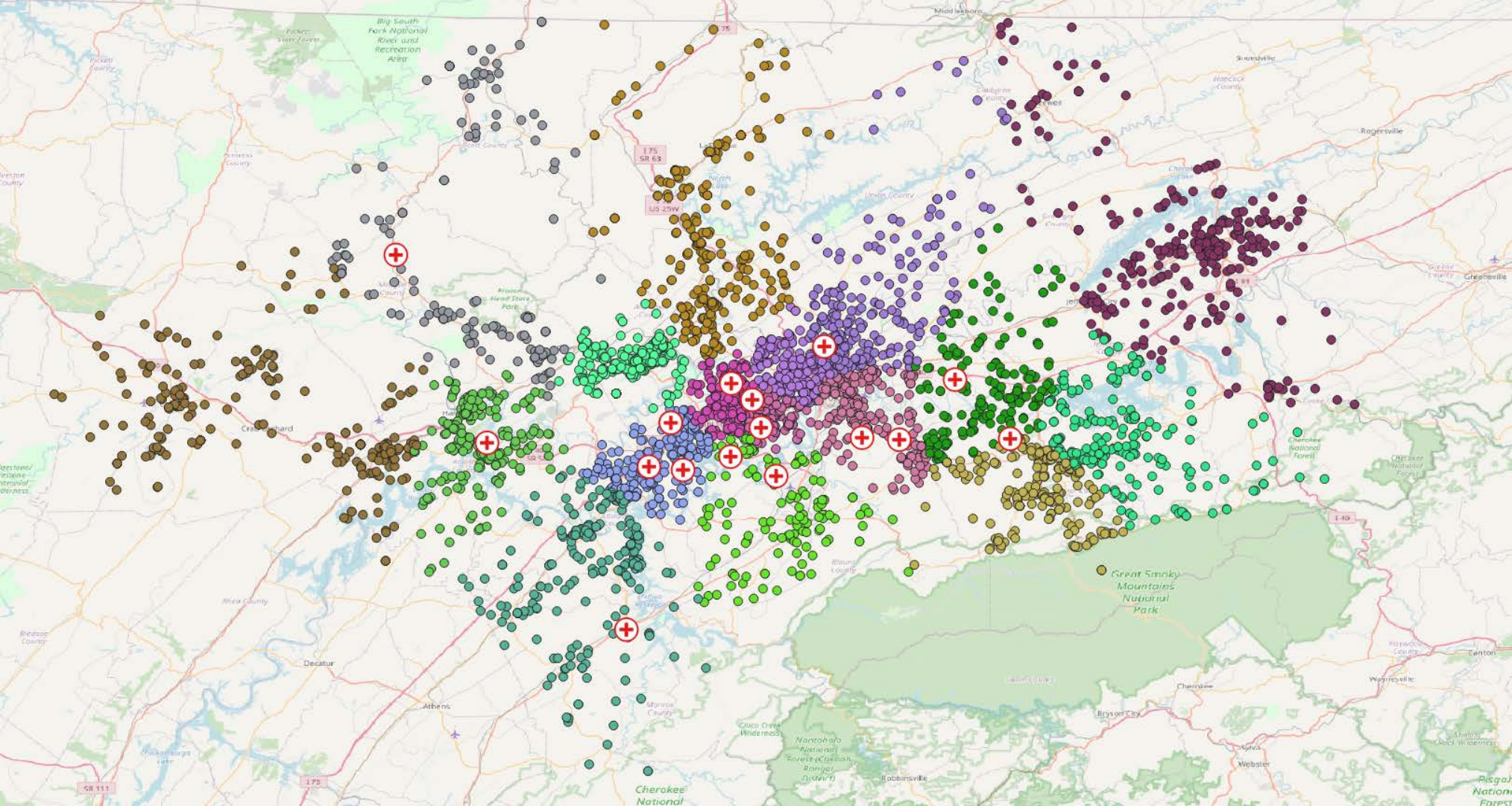}}
  \caption{Mapping of PT caregivers allocated patients}
  \label{fig:sub1_pt}
\end{subfigure}
\begin{subfigure}[b]{.23\textwidth}
  \centering
  \includegraphics[width=\textwidth]{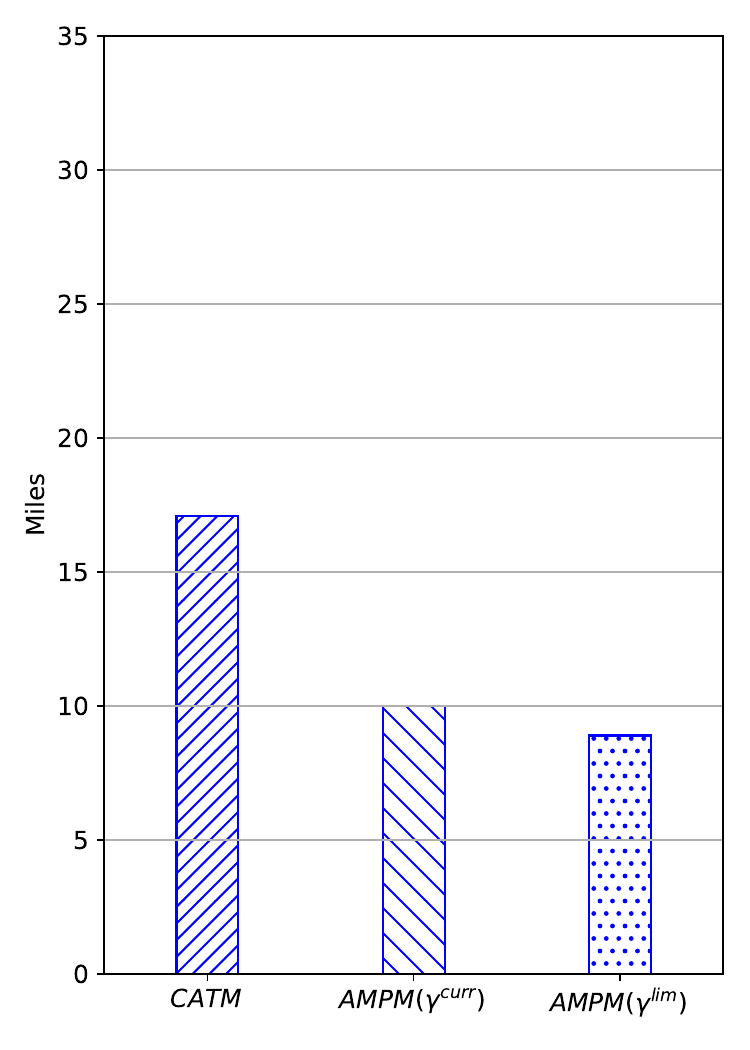}
  % \caption{AMPM}
  \label{fig:sub2_pt}
\end{subfigure}
\begin{subfigure}[b]{.23\textwidth}
  \centering
  \includegraphics[width=\textwidth]{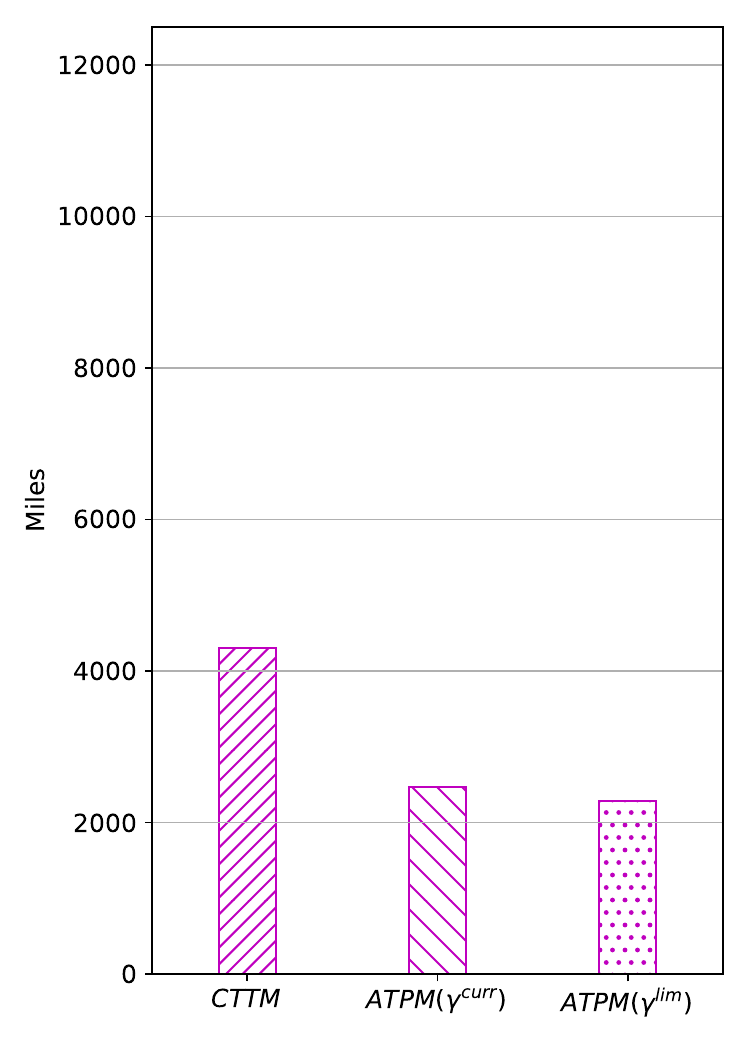}
  % \caption{ATPM}
  \label{fig:sub3_pt}
\end{subfigure}

\begin{subfigure}{.4\textwidth}
\raisebox{0.2cm}{%
  \centering
  \includegraphics[width=\textwidth, height=5cm]{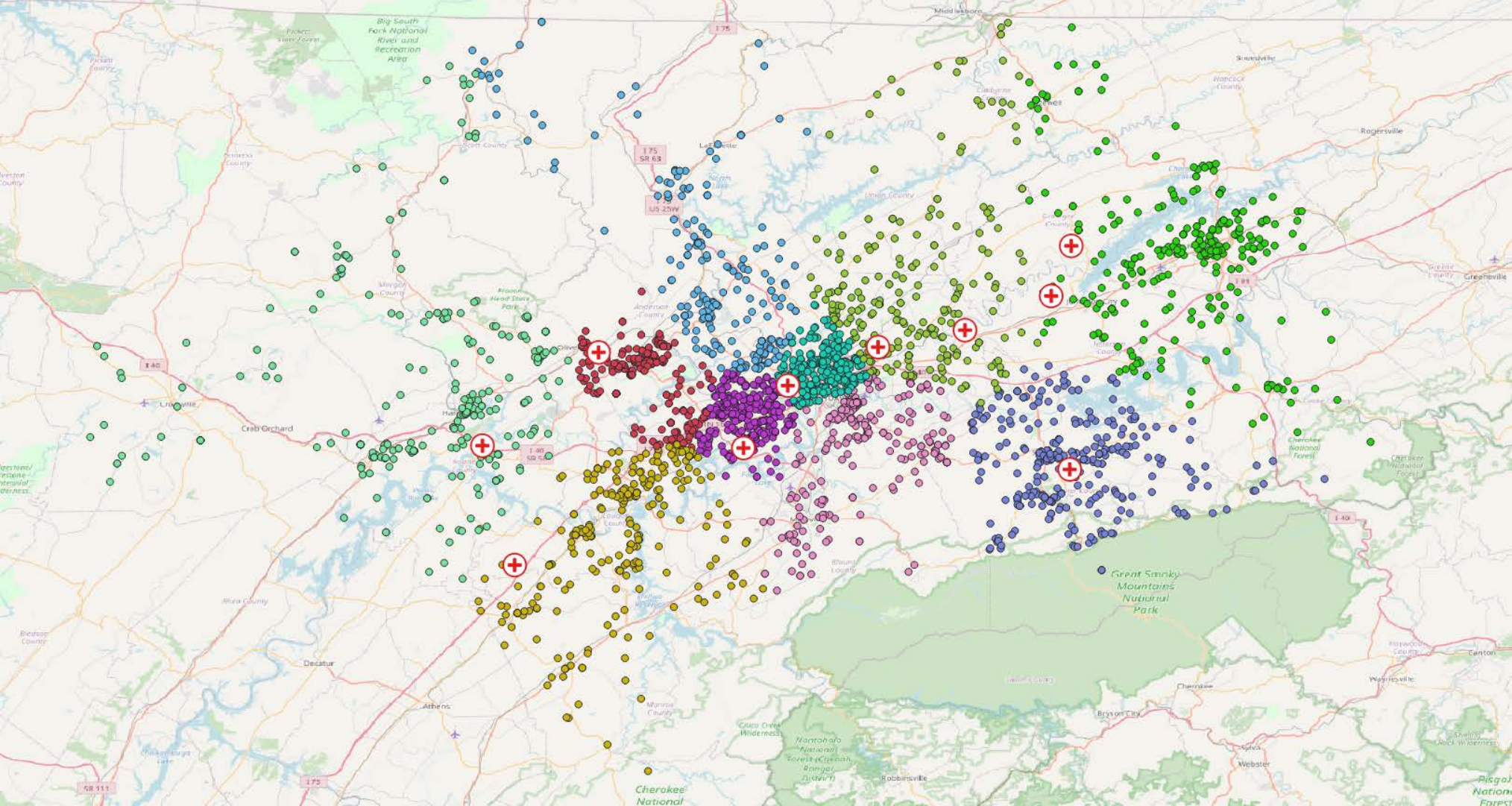}}
  \caption{Mapping of PTA caregivers allocated patients}
  \label{fig:sub1_cna}
\end{subfigure}
\begin{subfigure}[b]{.23\textwidth}
  \centering
  \includegraphics[width=\textwidth]{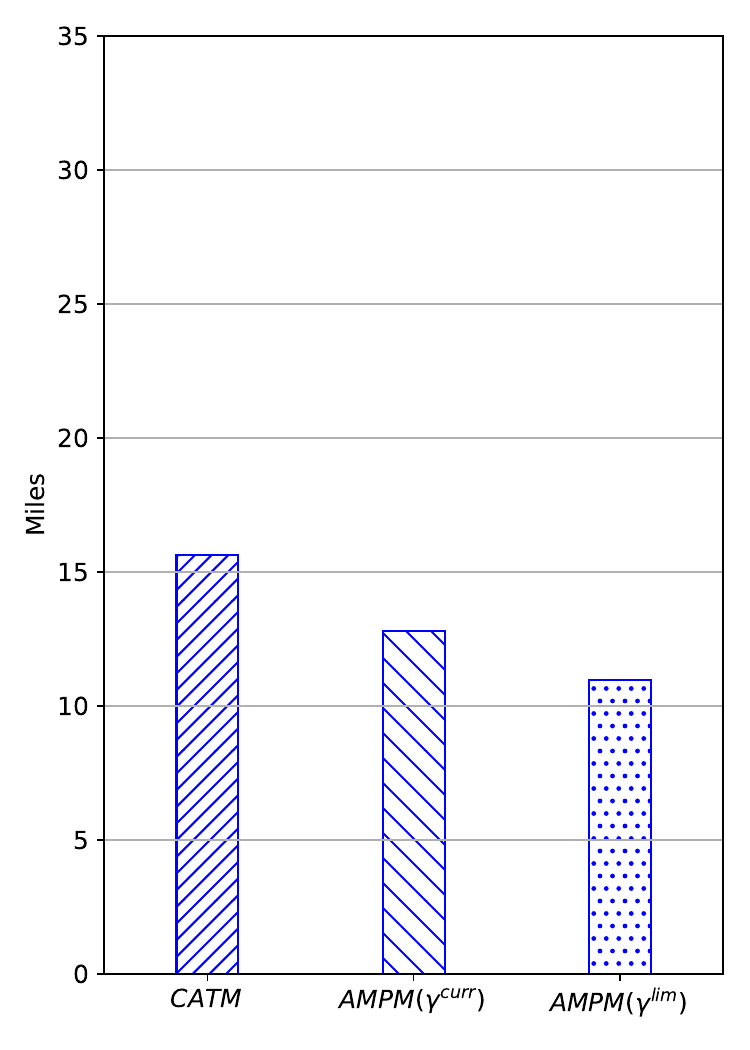}
  % \caption{AMPM}
  \label{fig:sub2_cna}
\end{subfigure}
\begin{subfigure}[b]{.23\textwidth}
  \centering
  \includegraphics[width=\textwidth]{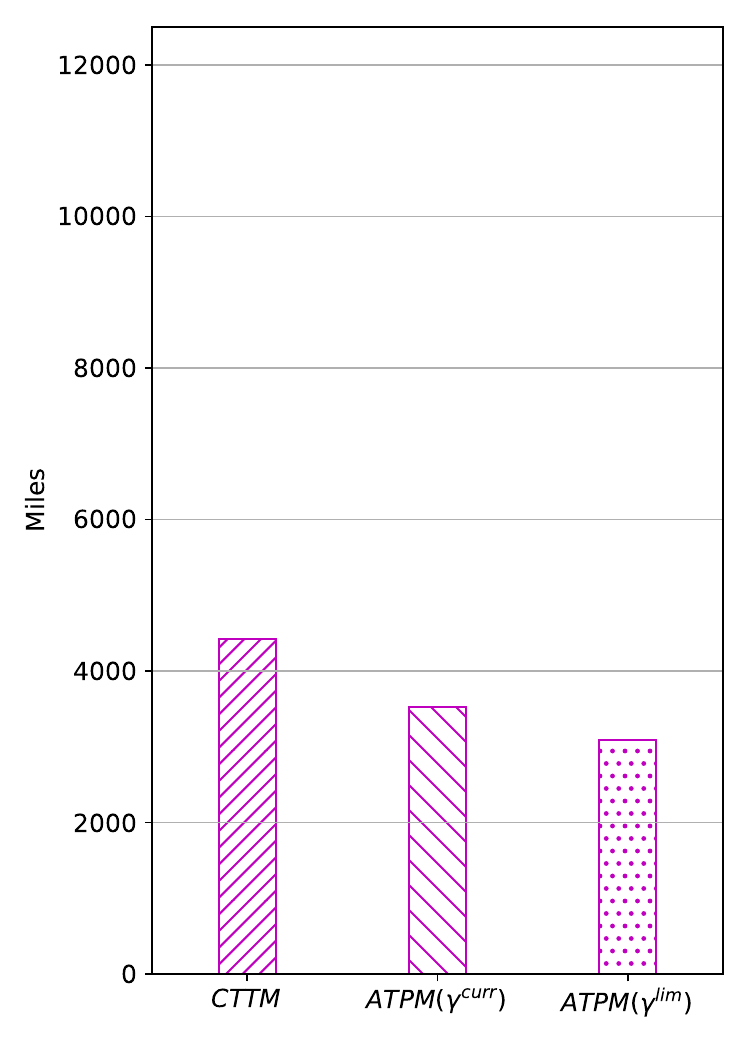}
  % \caption{ATPM}
  \label{fig:sub3_cna}
\end{subfigure}

\caption{“Left side: Visualization of geo-clustering”, "Middle: CATM compared with AMPM scores for $\gamma^{curr}$ and $\gamma^{lim}$" “ CTTM compared with Right side: ATPM improvements for $\gamma^{curr}$ and $\gamma^{lim}$.}
\label{improved_results}
\end{figure}

To benchmark the results of the proposed framework, \textbf{F}igure \ref{improved_results} depicts the mapping of clustering results from the proposed framework along with their performance plots for three sample disciplines, namely RN, PT, and PTA. For the sake of brevity, the extended version of the results and mappings for all other disciplines are provided in \hyperref[Appendix A]{Appendix A}. In \textbf{F}igure \ref{improved_results}, the dots on each map shows the locations of patients of each discipline during the period of study, and the red crosses indicate the locations of caregivers. The bar charts corresponding with each map compares the AMPM and ATPM for both $\gamma^{curr}$ and $\gamma^{lim}$ with the current average traveled mileage ($CATM$) and current total traveled mileages ($CTTM$) in each discipline, respectively.

\par
\textbf{T}able \ref{improved_table} indicates that the inconsistency of the improvement between disciplines comes from the fact that they are different in terms of the number of patients, number of caregivers, location of caregivers, and frequency of demands. Since the number and locations of the caregivers and their skills are pre-determined inputs of the problem, manipulating them is dependent on the policy of the company. 

\begin{table}[ht]
\caption{Clustering scores using different types of $\gamma$ compared to $CATM$.}
\label{improved_table}
\scalebox{0.9}{
\begin{tabularx}{\textwidth}{X X X X X X}
\hline
& & \multicolumn{2}{c}{$AMPM_d (\gamma^{curr})$} & \multicolumn{2}{c}{$AMPM_d (\gamma^{lim})$} \\ \cline{3-6}
\textit{Discipline} & \textit{$CATM_d$} & \textit{score} & \textit{\%decrease} & \textit{score} & \textit{\%decrease} \\
\hline
RN & 13.3700 & 7.7048 & 42.37\% & 7.3092 & 45.33\% \\
COTA & 29.6101 & 21.2134 & 28.36\% & 19.4718 & 34.24\% \\
CH & 21.5143 & 19.0779 & 11.32\% & 18.1845 & 15.48\% \\
PTA & 15.6334 & 12.7911 & 18.18\% & 10.9621 & 29.88\% \\
SLP & 28.1106 & 27.0158 & 3.89\% & 24.3492 & 13.38\% \\
LPN & 19.3389 & 17.4936 & 9.54\% & 16.8504 & 12.87\% \\
MSW & 24.0817 & 23.6812 & 1.66\% & 23.5084 & 2.38\% \\
OT & 23.9360 & 14.0688 & 41.22\% & 12.1524 & 49.23\% \\
PT & 17.0863 & 10.0065 & 41.44\% & 8.9023 & 47.90\% \\
CNA & 20.3114 & 14.8352 & 26.96\% & 12.7514 & 37.22\% \\
\hline
\end{tabularx}
}
\end{table}

%%%%%%%%%%%%%%%%%%%%%%

\subsection{Caregivers' supply analysis}
\label{Sensitivity_analysis}

\par
Sensitivity analysis (SA) allows us to understand how the suggested framework responds to a collection of fluctuations and modifications. One of the main concerns stated by HHA managers was how much they could improve the service level by hiring more caregivers. SA is the proper way to answer this concern. To quantify the improvement and retrogression caused by the number of caregivers, we defined a measurement index called the \textit{"Average Percentage Change"}, or APC. \textbf{E}quations \ref{APC_AMPM} and \ref{APC_ATPM} show how this index is calculated for each clustering metric.  APC is formulated on the basis of changes in the desires of HHA managers, which have been translated into AMPM and TMPM, which are $APC_d^{AM}$ and $APC_d^{AT}$, respectively. 

\begin{equation}
    \begin{aligned}
    APC_d^{AM} = \frac{AMPM_d^{curr} - AMPM_d^{alt}}{\left| |C_d^{curr}| - |C_d^{alt}| \right|}
    \label{APC_AMPM}
    \end{aligned}
\end{equation}

\begin{equation}
    \begin{aligned}
    APC_d^{AT} = \frac{ATPM_d^{curr} - ATPM_d^{alt}}{\left| |C_d^{curr}| - |C_d^{alt}| \right|}
    \label{APC_ATPM}
    \end{aligned}
\end{equation}

Where $|C_d^{curr}|$ and $|C_d^{alt}|$ are the current and alternative number of caregivers for discipline $d$, respectively, and
$AMPM_d^{curr}$ and $AMPM_d^{alt}$ are their corresponding evaluation metrics related to the current and alternative scenarios for discipline $d$.

\par
In this section, by manipulating the number of available caregivers for each discipline and doing re-allocation accordingly, we are going to manifest (I) How much improvement (retrogression) could be achieved and (II) which discipline is more sensitive to modification. In this sense, APC can supplement SA as a clear and comparable method of measuring findings. Amelioration in caregivers’ allocation could be realized when AMPM and ATPM are reduced. 

\par
In our research, the SA is conducted by randomly placing a group of caregivers in the vicinity of current patients and caregivers. The quantitative SA findings for all skills are shown in \textbf{T}ables \ref{SA_AMPM} and \ref{SA_ATPM}. The findings regarding AMPM and ATPM reveal that the most progress is made for MSW and COTA, while the most improvement occurs in COTA. The cause of their reaction is because of their geographical location. The caregivers with the above-mentioned skills have been dispersed at a significant distance from the patients. Thus, they react more intensely to changes of caregivers.

\begin{table}[!h]
\centering
\caption{AMPM clustering performance for alternative scenarios.}
\small % for smaller size
\scalebox{0.9}{
\begin{tabular}{cccccc}
\toprule
\textbf{Discipline} & \textbf{$Alt^{-}$} & \textbf{$Baseline$} & \textbf{$Alt^{+}$} & \textbf{$APC^{-}$} & \textbf{$APC^{+}$} \\
\midrule
RN & 20 & 25 & 30 & -1 & +1 \\
 & 6.880 & 5.982 & 5.449 & 3.0\% & -1.8\% \\
\midrule
COTA & 1 & 2 & 3 & -1 & +1 \\
 & 23.773 & 16.470 & 13.903 & 44.3\% & -15.6\% \\
\midrule
CH & 3 & 4 & 5 & -1 & +1 \\
 & 17.441 & 14.812 & 12.627 & 17.7\% & -14.8\% \\
\midrule
PTA & 7 & 10 & 13 & -1 & +1 \\
 & 12.078 & 9.931 & 8.575 & 7.2\% & -4.6\% \\
\midrule
SLP & 1 & 2 & 3 & -1 & +1 \\
 & 28.041 & 20.975 & 18.544 & 33.7\% & -11.6\% \\
\midrule
LPN & 3 & 4 & 5 & -1 & +1 \\
 & 15.492 & 13.582 & 12.208 & 14.1\% & -10.1\% \\
\midrule
MSW & 2 & 3 & 4 & -1 & +1 \\
 & 20.414 & 18.386 & 14.963 & 11.0\% & -18.6\% \\
\midrule
OT & 6 & 8 & 10 & -1 & +1 \\
 & 12.719 & 10.923 & 9.731 & 8.2\% & -5.5\% \\
\midrule
PT & 14 & 17 & 20 & -1 & +1 \\
 & 8.322 & 7.769 & 7.252 & 2.4\% & -2.2\% \\
\midrule
CNA & 4 & 6 & 8 & -1 & +1 \\
 & 14.167 & 11.518 & 10.138 & 11.5\% & -6.0\% \\
\bottomrule
\label{SA_AMPM}
\end{tabular}
}
\end{table}

\begin{table}[]
\centering
\caption{ATPM clustering performance for alternative scenarios.}
\small % for smaller size
\scalebox{0.9}{
\begin{tabular}{cccccc}
\toprule
\textbf{Discipline} & \textbf{$Alt^{-}$} & \textbf{$Baseline$} & \textbf{$Alt^{+}$} & \textbf{$APC^{-}$} & \textbf{$APC^{+}$} \\
\midrule
RN & 20 & 25 & 30 & -1 & +1 \\
 & 1784.978 & 1241.153 & 902.377 & 8.8\% & -5.5\% \\
\midrule
COTA & 1 & 2 & 3 & -1 & +1 \\
 & 19303.682 & 6683.827 & 3597.113 & 188.8\% & -46.2\% \\
\midrule
CH & 3 & 4 & 5 & -1 & +1 \\
 & 3283.031 & 2026.515 & 1390.022 & 62.0\% & -31.4\% \\
\midrule
PTA & 7 & 10 & 13 & -1 & +1 \\
 & 4506.949 & 2735.015 & 1887.609 & 21.6\% & -10.3\% \\
\midrule
SLP & 1 & 2 & 3 & -1 & +1 \\
 & 5608.371 & 2147.245 & 1209.620 & 161.2\% & -43.7\% \\
\midrule
LPN & 3 & 4 & 5 & -1 & +1 \\
 & 6777.348 & 4521.608 & 3275.039 & 49.9\% & -27.6\% \\
\midrule
MSW & 2 & 3 & 4 & -1 & +1 \\
 & 12287.087 & 7045.905 & 4187.077 & 74.4\% & -40.6\% \\
\midrule
OT & 6 & 8 & 10 & -1 & +1 \\
 & 4653.055 & 2945.684 & 2184.726 & 29.0\% & -12.9\% \\
\midrule
PT & 14 & 17 & 20 & -1 & +1 \\
 & 2595.075 & 1913.758 & 1474.502 & 11.9\% & -7.7\% \\
\midrule
CNA & 4 & 6 & 8 & -1 & +1 \\
 & 2029.576 & 1053.051 & 707.303 & 46.4\% & -16.4\% \\
\bottomrule
\label{SA_ATPM}
\end{tabular}
}
\end{table}

\par
For a better examination of the sensitivity with respect to variations in the number of available caregivers, we leverage the paired t-test. This statistical methodology is well-suited to compare the means of two related groups to discern if a significant difference exists \citep{emmert2019understanding}. The two groups here will be constituted by replicating the allocation algorithm 100 times each under the baseline and alternative scenario for increasing the number of available caregivers by 1 unit. Another t-test is done in the same way with the alternative scenario for decreasing by 1 unit. Each replication generates a different value for AMPM, resulting in a population of 100 values for each group.
The t-statistic and its associated $p-value$ are calculated from the mean and standard deviation of AMPM for both sets of 100 runs. This t-statistic measures the magnitude of the difference between the two groups relative to the variability within the data. If the t-statistic is significant, it signifies a substantial change in AMPM's performance due to the fluctuation in the number of caregivers. 

\par
\begin{figure}[!h]
    \centering
    \includegraphics[width=0.8\textwidth]{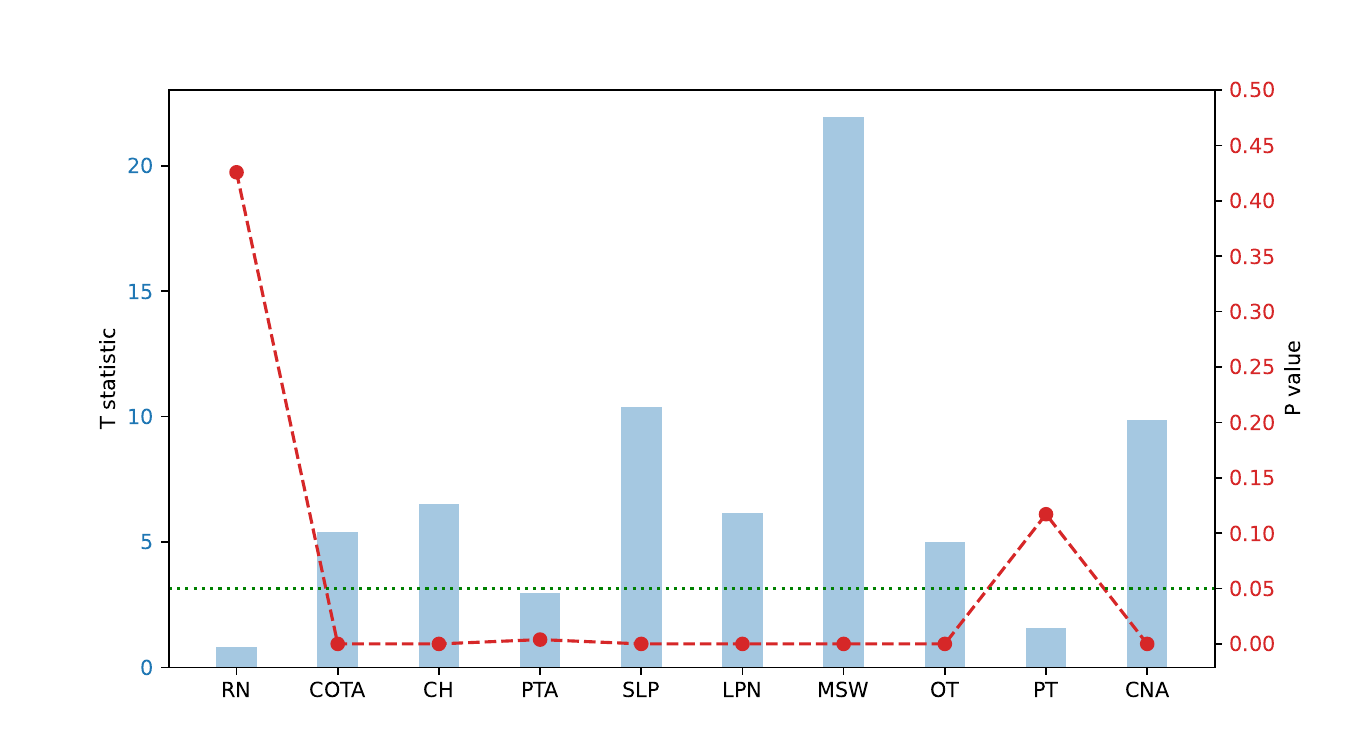}
    \caption{Paired t-test results to compare \textbf{$Baseline$} with \textbf{$APC^{-}$} scenario.}
    \label{t-test_minus}
\end{figure}
\begin{figure}[!h]
    \centering
    \includegraphics[width=0.8\textwidth]{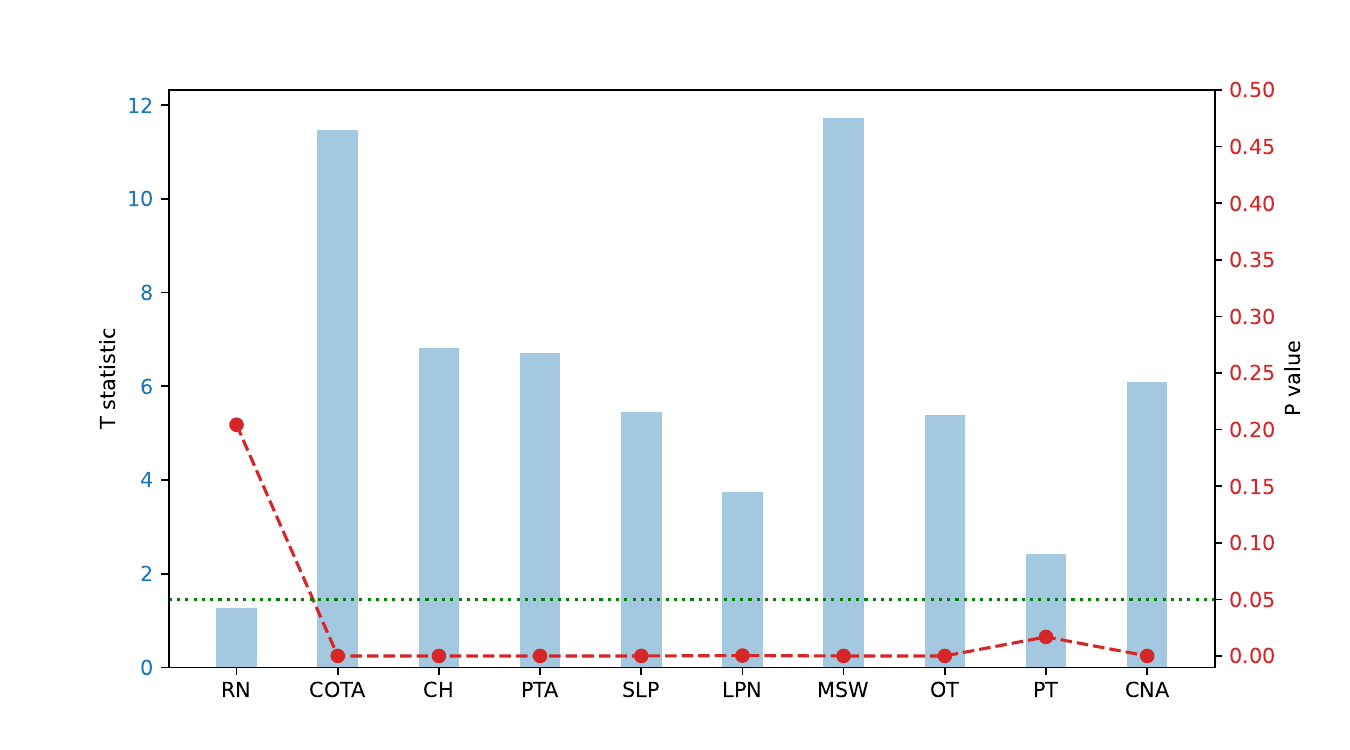}
    \caption{Paired t-test results to compare \textbf{$Baseline$} with \textbf{$APC^{+}$} scenario.}
    \label{t-test_plus}
\end{figure}

Figures \ref{t-test_minus} and \ref{t-test_plus} show these results for a significance level of 0.95 confidence interval. When the p-value is under 0.05, we can reject the null hypothesis that there is no significant difference between the two groups. This situation happens for all changes in all disciplines except RN and PT in Figure \ref{t-test_minus} and RN in Figure \ref{t-test_plus}, implying that in these cases changing the number of caregivers by 1 unit will not result in significant changes in AMPM. This analysis gives us valuable insights, which can serve as a critical basis for strategic planning and optimization of resources.

%%%%%%%%%%%%%%%%%%%%%%%%%%%%%%%%%%%%%

\subsection{Managerial implications}
\label{Discussion_and_conclusion}

From a pragmatic point of view, the results shed light on several key insights. First, the proposed caregiver allocation is flexible, allowing caregivers to arrange the sequence of visits based on their real-time location and the available patients. This flexibility is beneficial for caregivers, patients, and the HHA. It allows all parties to rearrange their schedules and re-plan resources as needed. In addition, the flexibility of the system ensures that caregivers do not have to take an unfair share of the workload. In general, the study's results suggest that the proposed caregiver allocation system is a practical and effective way to improve caregiver scheduling efficiency and work-life balance.

Second, data analysis reveals that the traveled mileage from/to caregivers' homes in historical data was more than between patients' visits, confirming the weakness of their planning. Our simulation results indicate that our proposed framework has resolved this weakness, reducing the average traveled mileage. While the minimum reduction is around 4\%, it goes up to 42\% depending on discipline.

The third implication is considering continuity of care as a pivotal criterion of patient satisfaction. The proposed solution takes this into account by naturally prioritizing the same caregivers to an existing patient's location unless the allocation is not feasible, thus providing a reliable caregiver planning system that can guarantee maximum continuity of care.

Last but not least insight proposed by the framework could illuminate the recruiting process for human resource managers, helping them to determine which skills are in demand and which skills are in decline. This information could be used to improve the recruiting process by ensuring that the right people are hired for the right jobs. Additionally, the framework could be used to track the effectiveness of the recruiting process over time, providing insight into how it could gradually improve.

When the HHA applied a prototype of this decision-making framework, several key practical improvements were achieved. The geospatial analysis of historical visits produced a starting place to assign territory coverage to current workers for each caregiving discipline. While the simulation projected higher reductions in total travel, the early real-world application was closer to 5\% to 10\% improvement in miles driven, which translated to thousands of dollars in annual travel costs. The agency’s implementation of the model suggestions was limited by current operational structures of care planning and worker dispatch, clinician and supervisor preferences based on experience and established habits, and the difficulties of communicating the highly technical decision-support algorithm across the organization. Real-world improvement could be closer to simulated improvement when businesses implement the framework before or during a large shift in culture or technology.

One unexpected benefit of applying the framework was the ability to justify requests for additional clinical positions from corporate leadership with objective data from the optimization model. Corporate structures can sometimes use productivity models for workforce requisitions that are prohibitive to the growth of their business units unless upper management can be persuaded that more positions are needed. The optimization model provided supportive evidence that certain disciplines required more workers, and helped with future planning by projecting discipline counts given different circumstances.

From a practical perspective, the proposed framework could be beneficial for businesses other than HHC which have two main characteristics:
\textit{(I)} They have a set of employees who need to be assigned to a set of customers. \textit{(II)} There are various sources of uncertainty that make it impractical to impose a strict sequence of visits. The proposed framework has been designed in a way that even if some unpredictable circumstances occur at the last minute, the plan will remain valid. Besides HHAs, sales companies, consulting companies, and marketing agencies where the employee needs to travel to visit a set of clients are examples of businesses that can take advantage of our research study in their planning. 

\section{Conclusion and future direction}
\label{Conclusion}

\par
In addressing the pressing challenges faced by a home healthcare service provider located in Tennessee, US, the proposed caregiver allocation framework has proven to be a robust method that employs meticulous and data-driven stages. Beginning with the cleansing and labeling of historical data, this process ensures that the information at hand is reliable and formatted aptly for seamless processing. From here, the advantages of the optimally tuned spectral clustering method via genetic algorithms take center stage, diligently clustering patients and adjusting these clusters to achieve optimal results. By employing k-NN clustering for both caregiver and patient allocation, the method facilitates the creation of a baseline that is both intuitive and practical.

\par
An essential feature of this system is its ability to continually validate its results. Through the allocation feasibility check, we ensure that the generated allocations are not only mathematically sound but also meet the real-world requirements of caregiver working hours. Such rigorous validation is paramount for real-world applicability, as it fosters trust and eliminates the potential for oversights. If any inconsistencies arise, the framework is dynamic enough to recalibrate its allocations, guaranteeing the most efficient outcomes without compromising feasibility.

\par
Beyond the core steps, the incorporation of caregivers' supply analysis underscores the holistic nature of the framework. This forward-looking component equips management with the vital insights needed to make informed decisions about recruitment and resource allocation. It's not merely about meeting current demands but also about anticipating and preparing for future challenges and fluctuations. This level of foresight is essential for maintaining the high service standards that patients expect and deserve.

\par
In conclusion, the caregiver allocation framework offers a comprehensive and dynamic solution to the complexities of home healthcare service provision. By merging sophisticated algorithms with practical application checks and strategic foresight, it paves the way for a future where patients receive the care they need when they need it, and where caregivers are deployed with precision and efficiency. It's a promising stride towards a more optimized and patient-centered healthcare landscape.

\par
In the research concerning the HHA case study, certain assumptions and limitations, such as using a correction coefficient of approximately 1.285 for calculating distances and substituting exact location coordinates with zip-code centers, may introduce approximation and spatial inaccuracies to the results. Additionally, the flexibility in visitation sequences and caregivers' ability to adjust their working hours can lead to deviations from predicted schedules, emphasizing the importance of considering these factors when interpreting or applying the study's findings.

\par
Future research endeavors in this area can significantly enhance the efficacy and precision of worker allocation in home healthcare settings. A pressing direction would be the development of a real-time platform allowing decision-makers to predict worker-to-address allocations, leveraging the potential of Clustering and other Machine Learning methodologies. Furthermore, delving deep into the statistical analysis will be invaluable, particularly in discerning variances in visitation and travel times based on factors like location, discipline, workers involved, and the patient's condition. 

\par
As the healthcare sector is often influenced by seasonal trends, it is imperative to examine how visitation metrics, workers' demand, and patient counts oscillate seasonally. Moreover, understanding the implications of external factors, such as the COVID-19 pandemic, on these metrics can offer pivotal insights. Lastly, optimizing the clustering of patients can be achieved by introducing a Multi-Objective goal function. This would aim to minimize disparities in patient allocations across workers and standardize measures like AMPM and ATPM for each caregiver's allocation, ensuring more balanced and efficient service delivery.

\section*{Declaration of generative AI and AI-assisted technologies in the writing process}
\par
During the preparation of this work, the authors used ChatGPT in order to improve the readability of the text. After using this service, the authors reviewed and edited the content as needed and they take full responsibility for the content of the publication.

\newpage
\section*{Appendix A}
\label{Appendix A}
\appendix
Benchmarking of the results for all disciplines: “Top: Visualization of geo-clustering”, "Down left: CATM compared with AMPM scores for $\gamma^{curr}$ and $\gamma^{lim}$" “Down right: CTTM compared with ATPM improvements for $\gamma^{curr}$ and $\gamma^{lim}$.

\setcounter{figure}{0}                       % <---------------
\renewcommand\thefigure{A.\arabic{figure}}   % <---------------

\begin{figure}[]
\centering
\begin{subfigure}{.8\textwidth}
\raisebox{0.2cm}{%
  \centering
  \includegraphics[width=\textwidth, height=8cm]{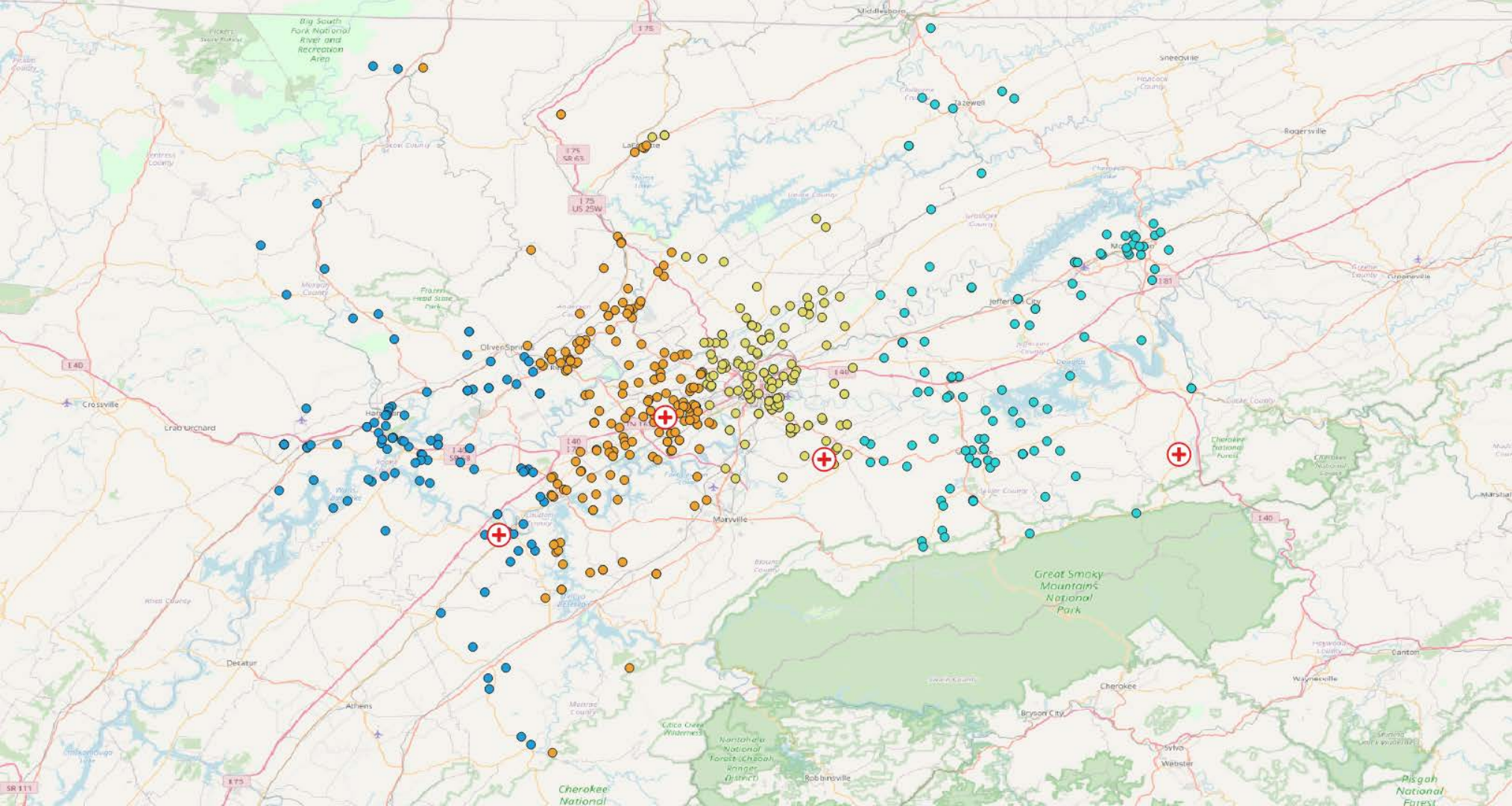}}
  \caption{Mapping of CH caregivers allocated patients}
  \label{fig:sub1_ch}
\end{subfigure}
\begin{subfigure}[b]{.4\textwidth}
  \centering
  \includegraphics[width=\textwidth]{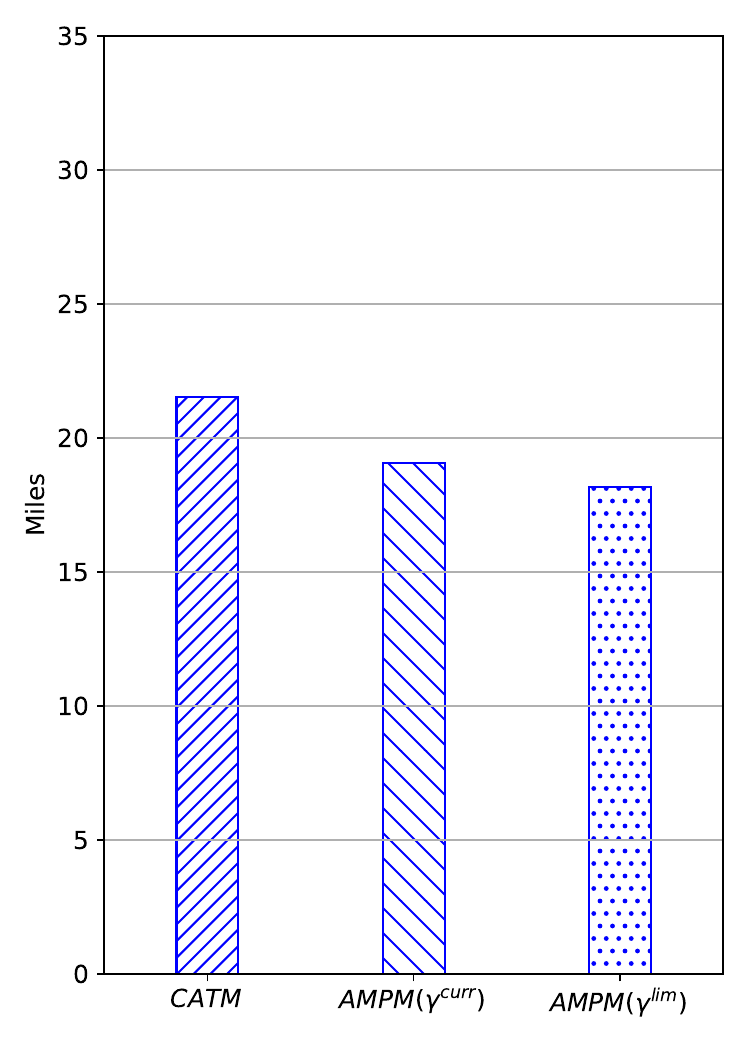}
  \caption{Current and expected average mileage of CH caregivers}
  \label{fig:sub2_ch}
\end{subfigure}
\begin{subfigure}[b]{.4\textwidth}
  \centering
  \includegraphics[width=\textwidth]{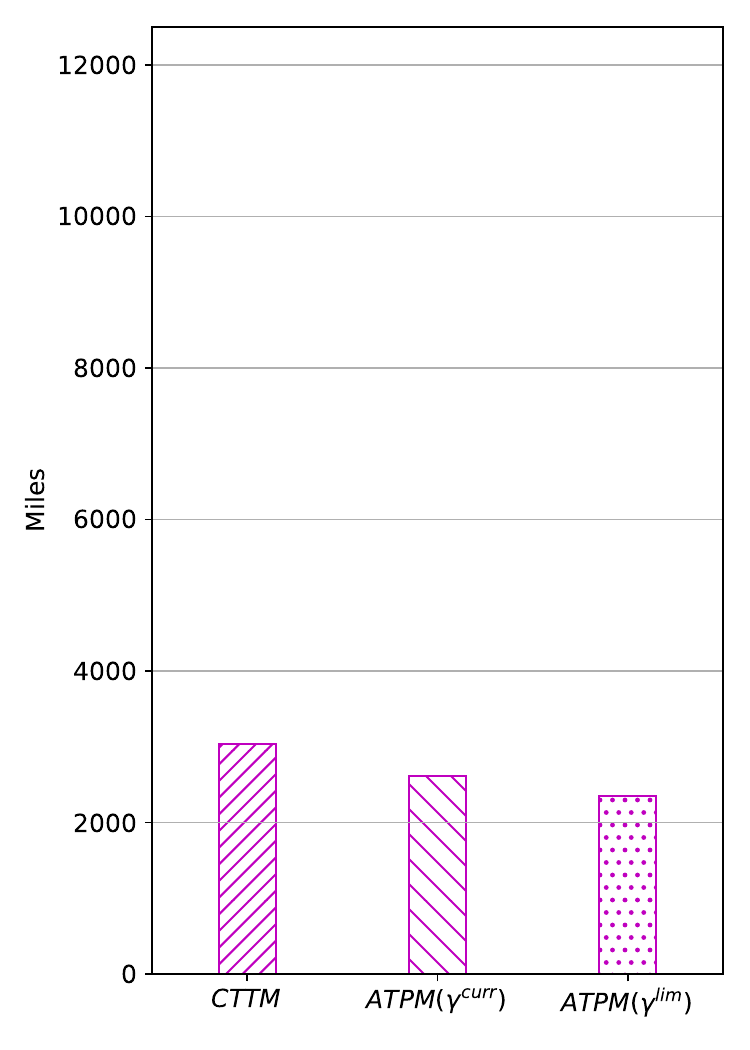}
  \caption{Current and expected total mileage of CH caregivers}  
  \label{fig:sub3_ch}
\end{subfigure}
\caption{Results for CH discipline.}
\end{figure}

%%%%%%%%%%%%%%%%%%%%%%%%%%%%%%%%%%%%%%%%%
\begin{figure}[]
\centering
\begin{subfigure}{.8\textwidth}
  \raisebox{0.2cm}{%
    \centering
    \includegraphics[width=\textwidth, height=8cm]{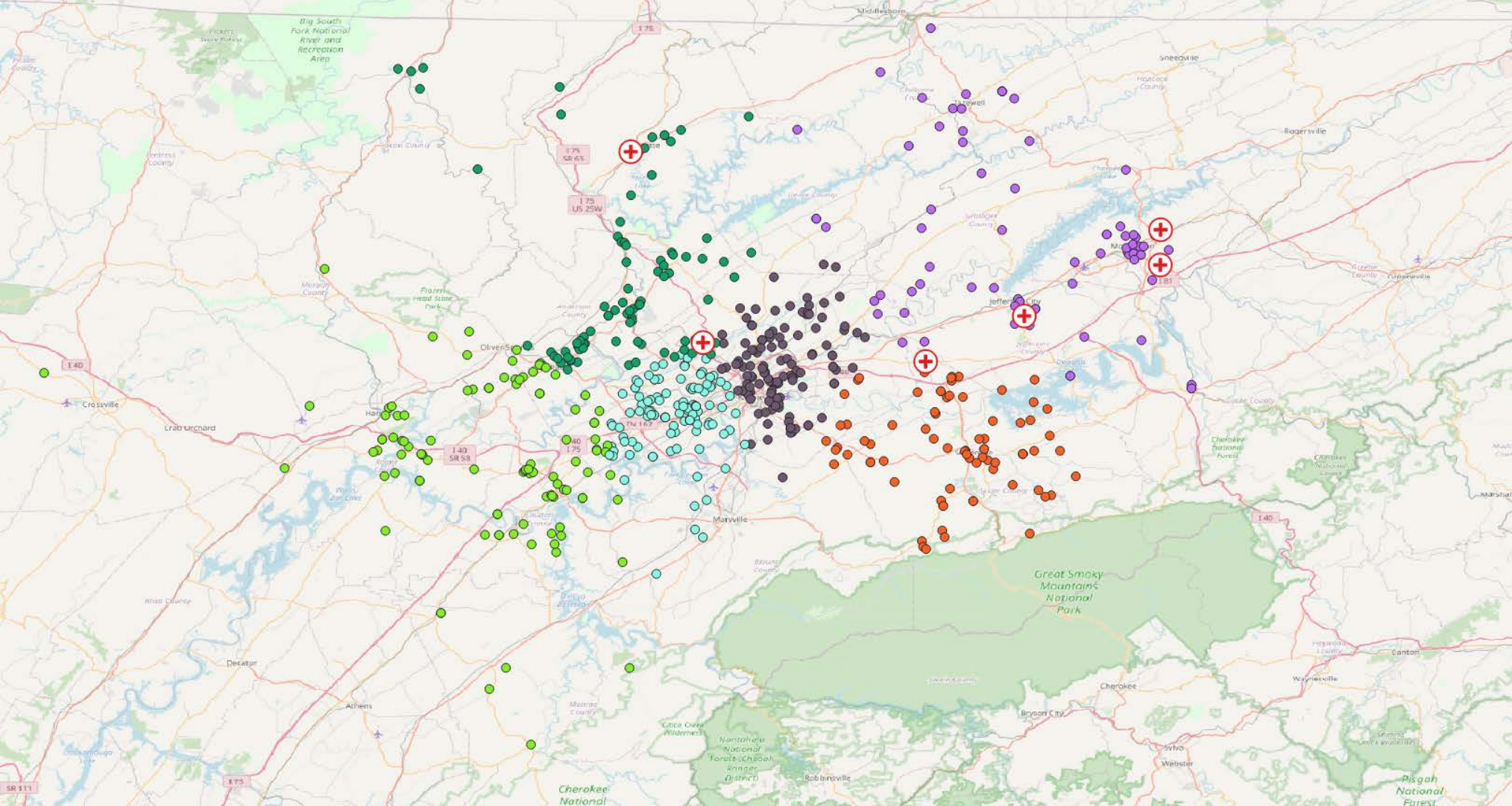}}
    \caption{Mapping of CNA caregivers allocated patients}
    \label{fig:sub1_cna}
\end{subfigure}
\begin{subfigure}[b]{.4\textwidth}
  \centering
  \includegraphics[width=\textwidth]{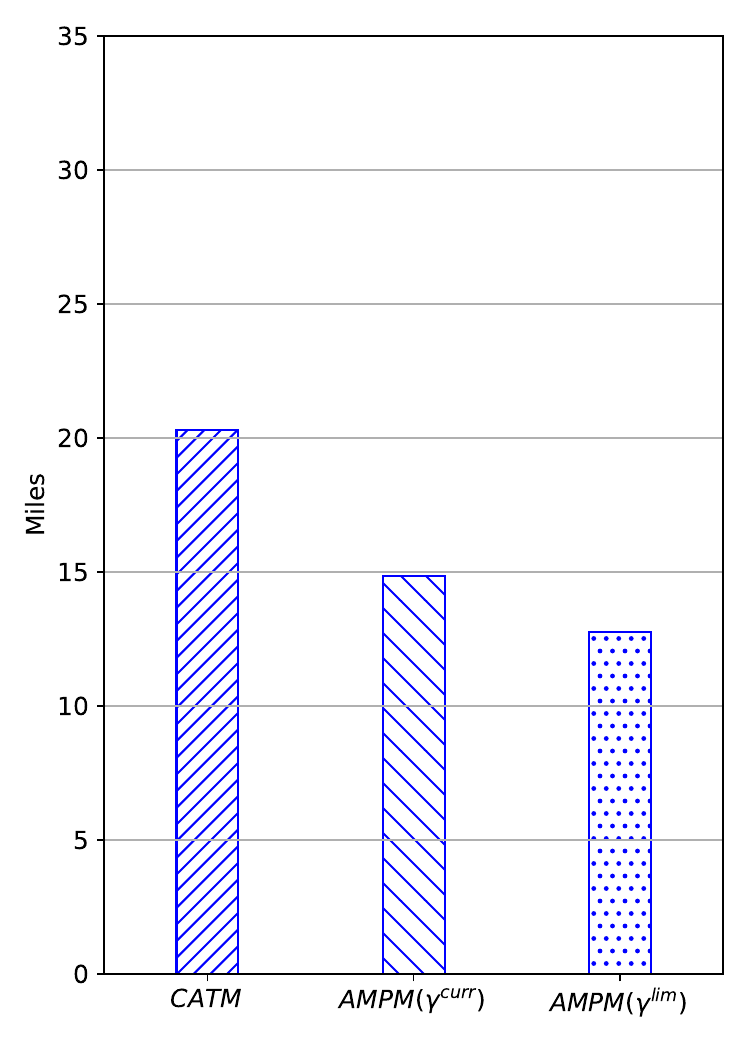}
  \caption{Current and expected average mileage of CNA caregivers}
  \label{fig:sub2_cna}
\end{subfigure}
\begin{subfigure}[b]{.4\textwidth}
  \centering
  \includegraphics[width=\textwidth]{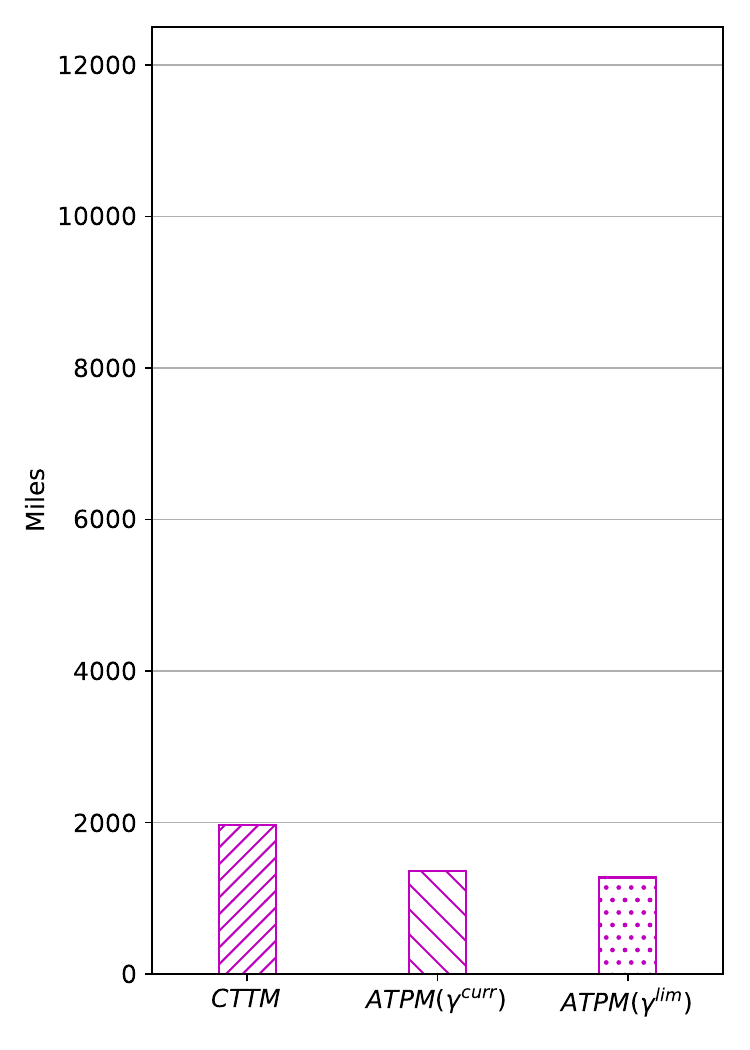}
  \caption{Current and expected total mileage of CNA caregivers}
  \label{fig:sub3_cna}
\end{subfigure}
\caption{Results for CNA discipline.}
\end{figure}

\begin{figure}[]
\centering
\begin{subfigure}{.8\textwidth}
  \raisebox{0.2cm}{%
    \centering
    \includegraphics[width=\textwidth, height=8cm]{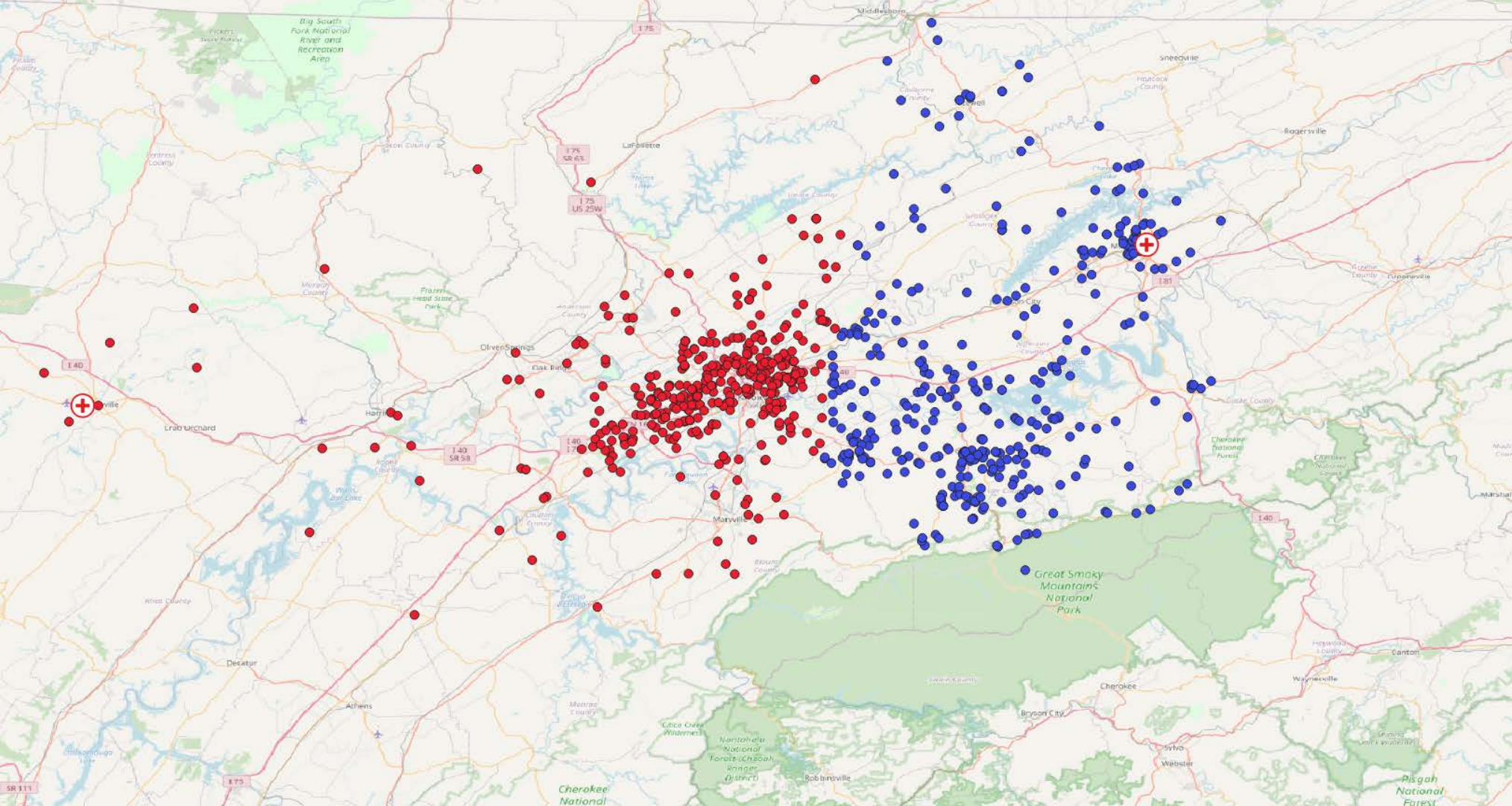}}
    \caption{Mapping of COTA caregivers allocated patients}
    \label{fig:sub1_cota}
\end{subfigure}
\begin{subfigure}[b]{.4\textwidth}
  \centering
  \includegraphics[width=\textwidth]{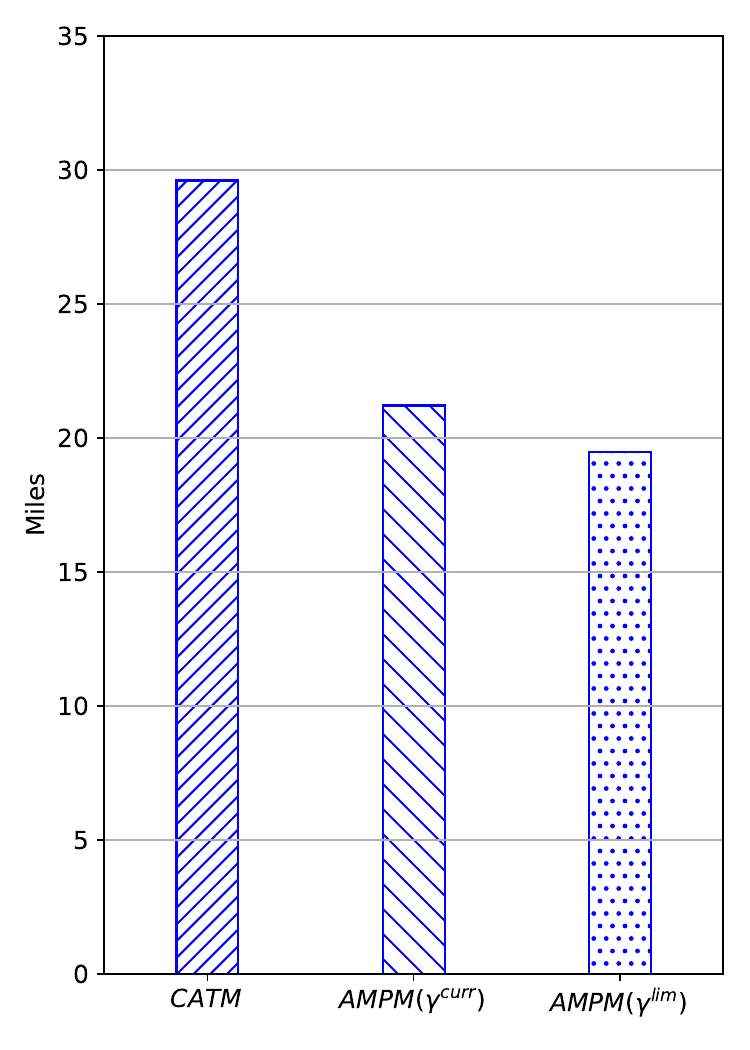}
  \caption{Current and expected average mileage of COTA caregivers}
  \label{fig:sub2_cota}
\end{subfigure}
\begin{subfigure}[b]{.4\textwidth}
  \centering
  \includegraphics[width=\textwidth]{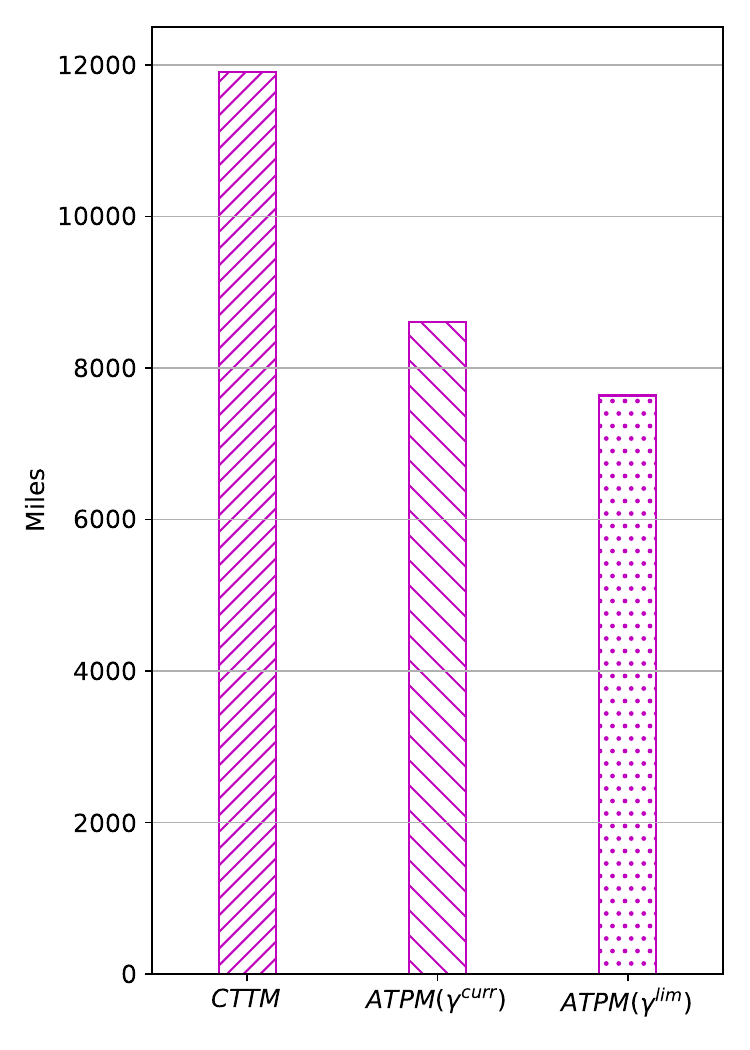}
  \caption{Current and expected total mileage of COTA caregivers}
  \label{fig:sub3_cota}
\end{subfigure}
\caption{Results for COTA discipline.}
\end{figure}

\begin{figure}[]
\centering
\begin{subfigure}{.8\textwidth}
  \raisebox{0.2cm}{%
    \centering
    \includegraphics[width=\textwidth, height=8cm]{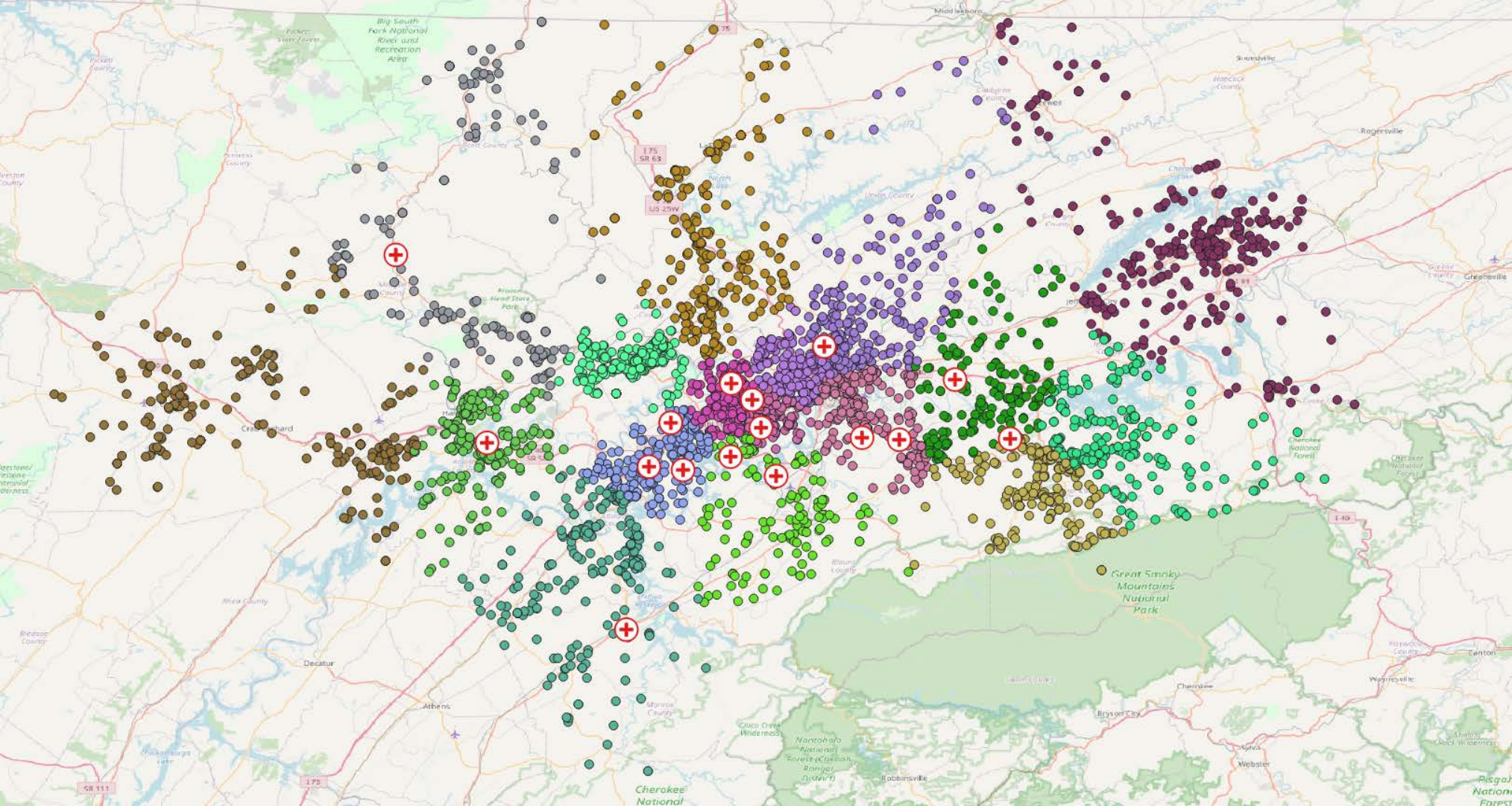}}
    \caption{Mapping of PT caregivers allocated patients}
    \label{fig:sub1_pt}
\end{subfigure}
\begin{subfigure}[b]{.4\textwidth}
  \centering
  \includegraphics[width=\textwidth]{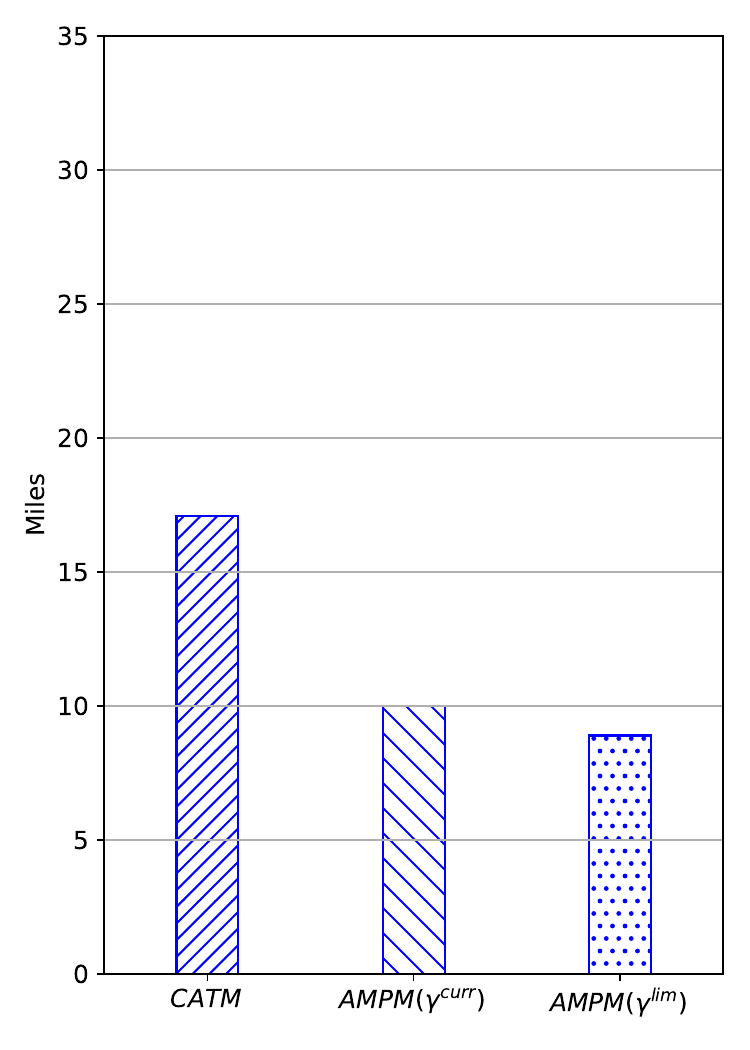}
  \caption{Current and expected average mileage of PT caregivers}
  \label{fig:sub2_pt}
\end{subfigure}
\begin{subfigure}[b]{.4\textwidth}
  \centering
  \includegraphics[width=\textwidth]{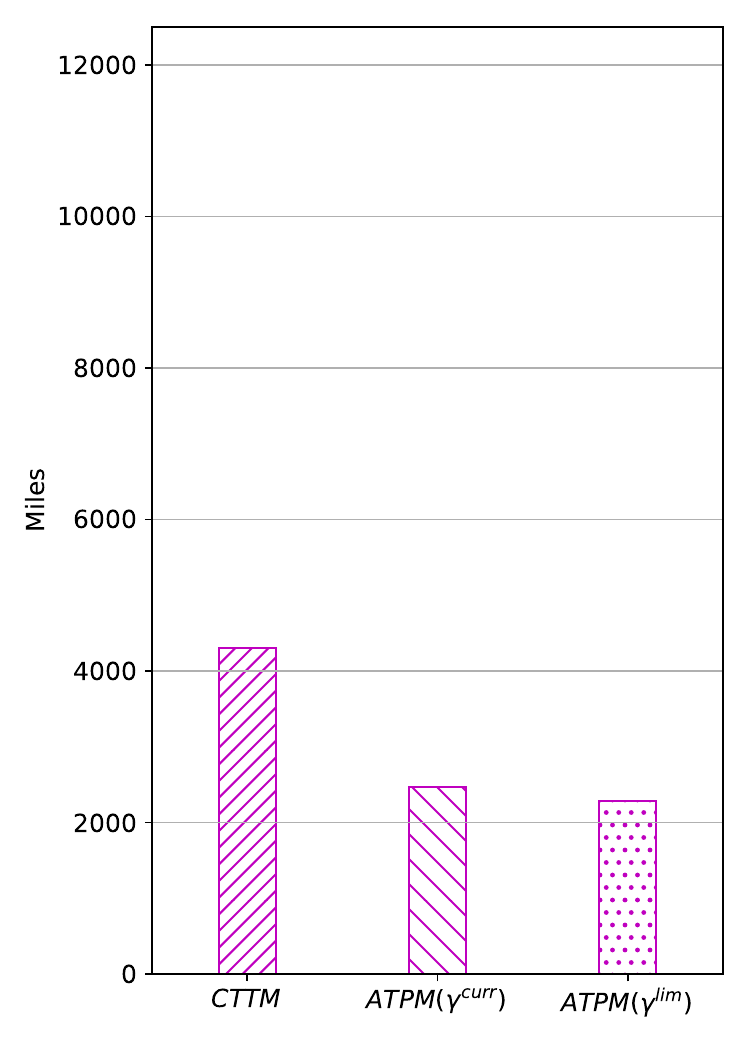}
  \caption{Current and expected total mileage of PT caregivers}
  \label{fig:sub3_pt}
\end{subfigure}
\caption{Results for PT discipline.}
\end{figure}

\begin{figure}[]
\centering
\begin{subfigure}{.8\textwidth}
  \raisebox{0.2cm}{%
    \centering
    \includegraphics[width=\textwidth, height=8cm]{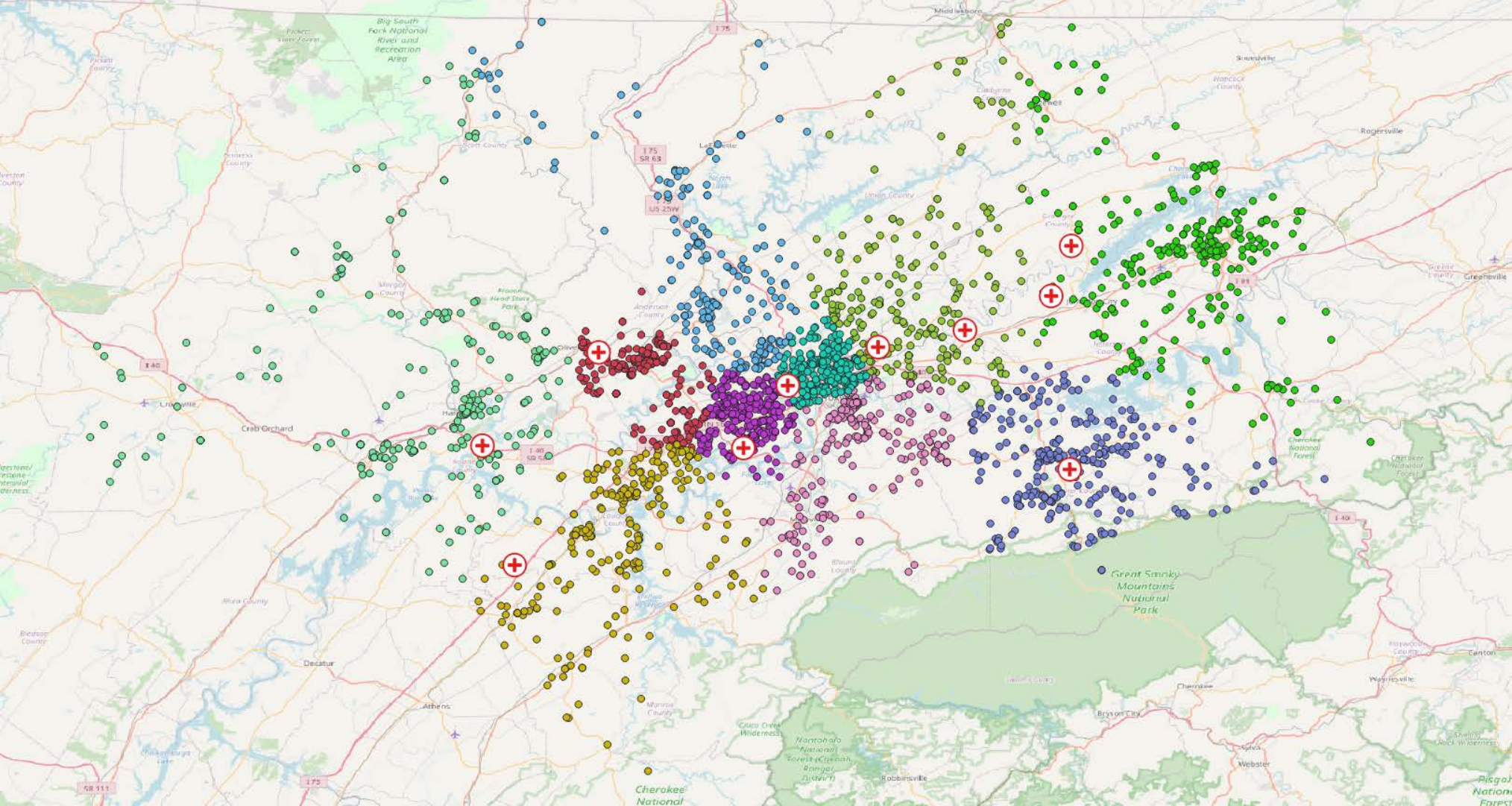}}
    \caption{Mapping of PTA caregivers allocated patients}
    \label{fig:sub1_pta}
\end{subfigure}
\begin{subfigure}[b]{.4\textwidth}
  \centering
  \includegraphics[width=\textwidth]{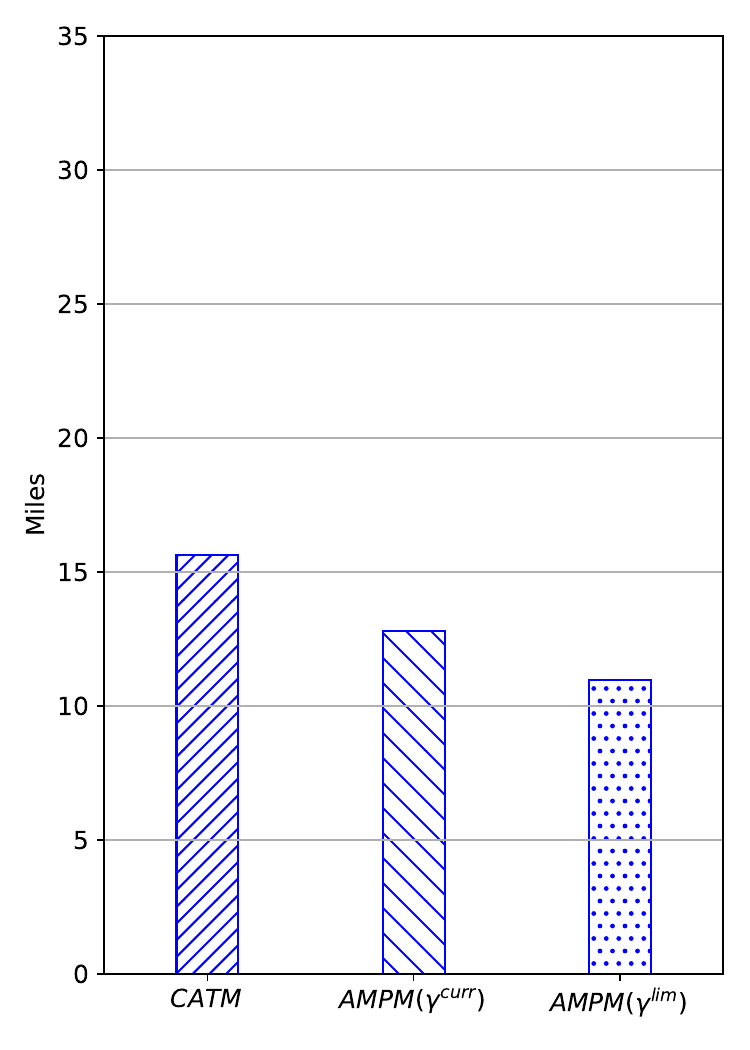}
  \caption{Current and expected average mileage of PTA caregivers}
  \label{fig:sub2_pta}
\end{subfigure}
\begin{subfigure}[b]{.4\textwidth}
  \centering
  \includegraphics[width=\textwidth]{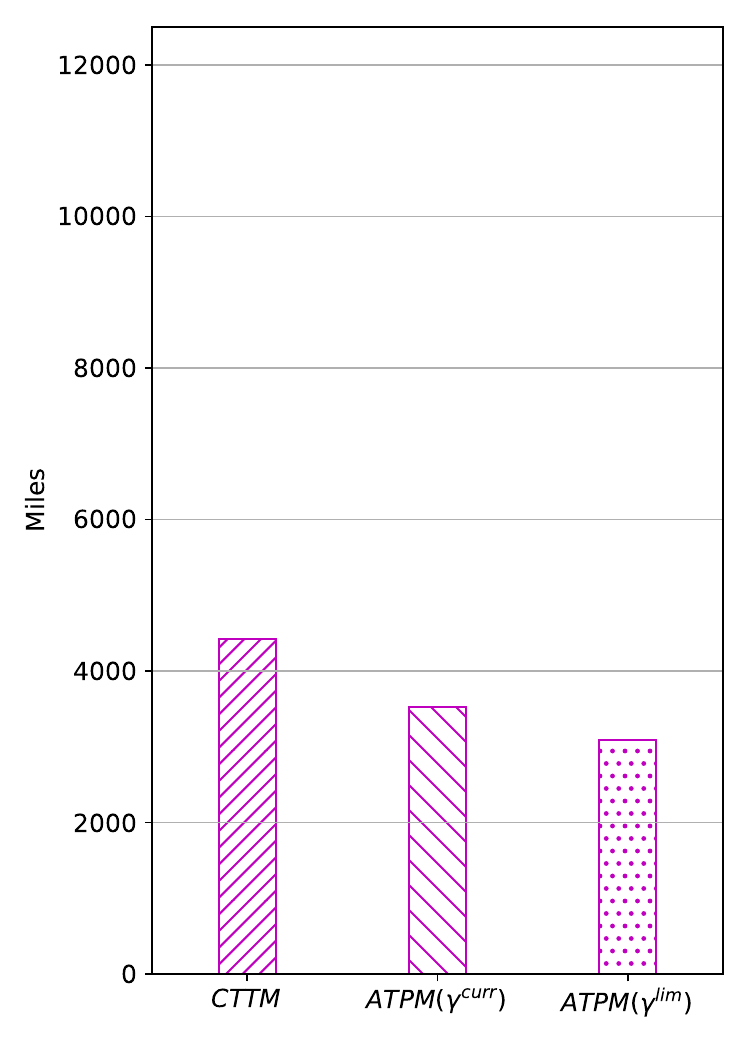}
  \caption{Current and expected total mileage of PTA caregivers}
  \label{fig:sub3_pta}
\end{subfigure}
\caption{Results for PTA discipline.}
\end{figure}

\begin{figure}[]
\centering
\begin{subfigure}{.8\textwidth}
  \raisebox{0.2cm}{%
    \centering
    \includegraphics[width=\textwidth, height=8cm]{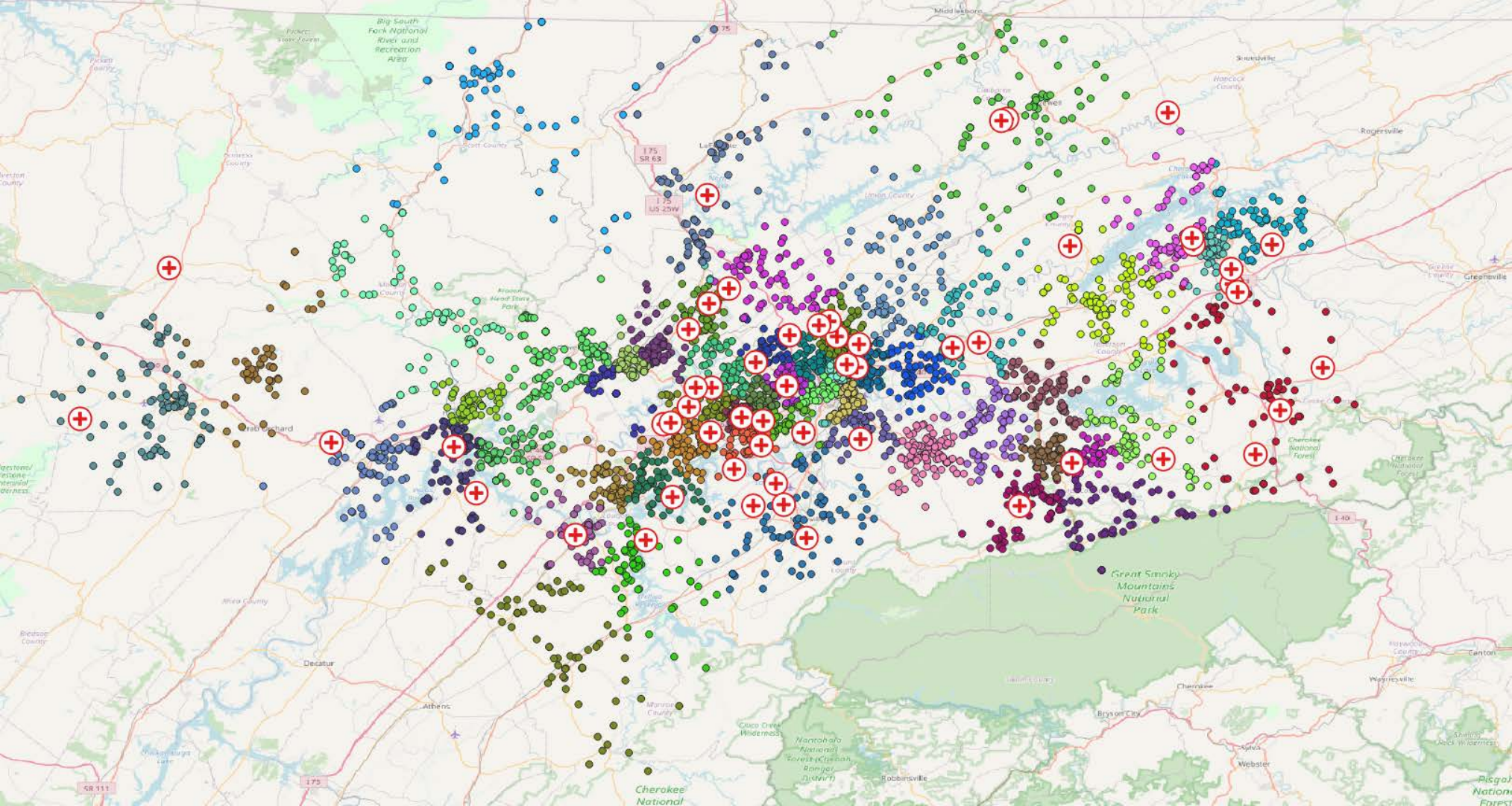}}
    \caption{Mapping of RN caregivers allocated patients}
    \label{fig:sub1_rn}
\end{subfigure}
\begin{subfigure}[b]{.4\textwidth}
  \centering
  \includegraphics[width=\textwidth]{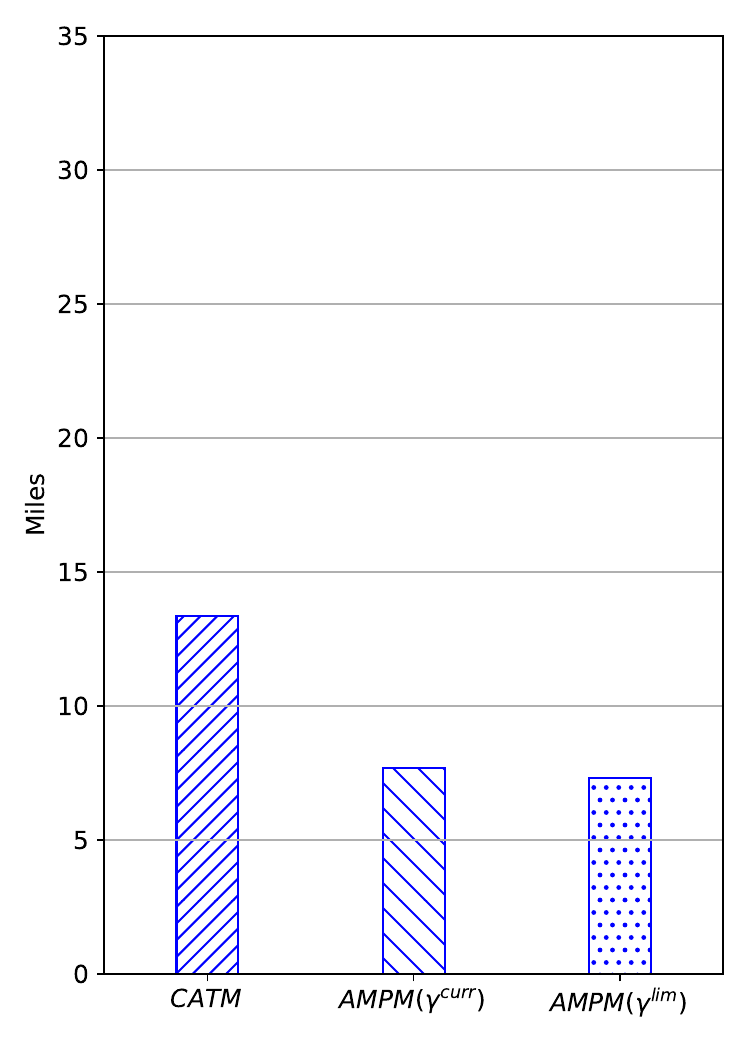}
  \caption{Current and expected average mileage of RN caregivers}
  \label{fig:sub2_rn}
\end{subfigure}
\begin{subfigure}[b]{.4\textwidth}
  \centering
  \includegraphics[width=\textwidth]{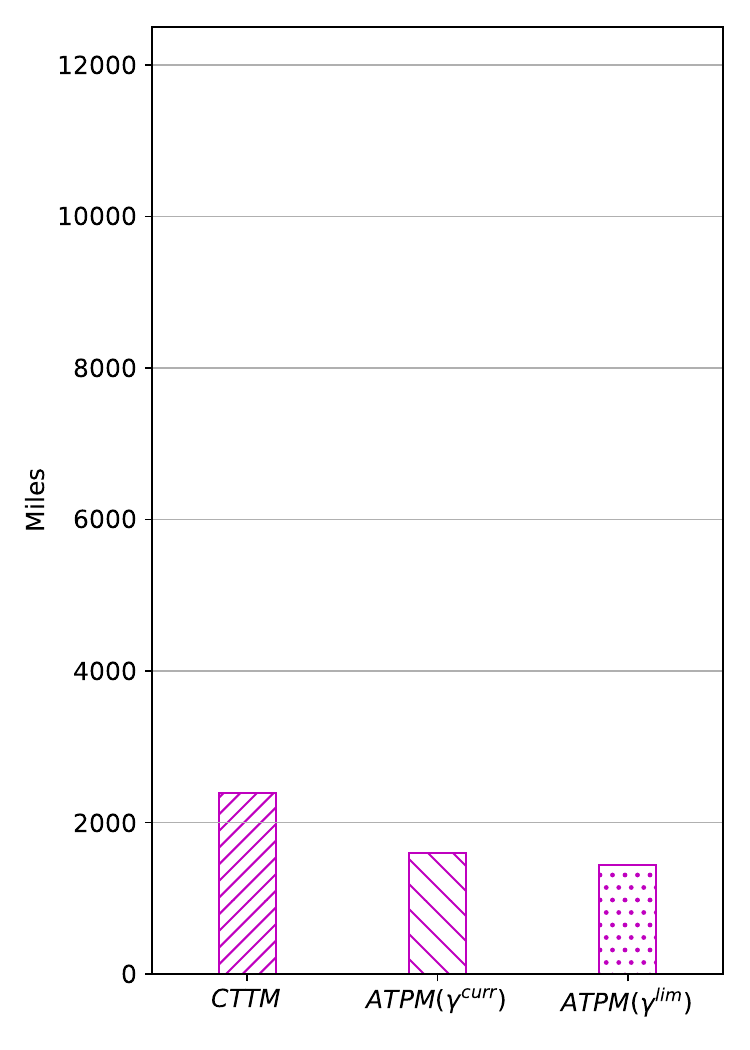}
  \caption{Current and expected total mileage of RN caregivers}
  \label{fig:sub3_rn}
\end{subfigure}
\caption{Results for RN discipline.}
\end{figure}

\begin{figure}[]
\centering
\begin{subfigure}{.8\textwidth}
  \raisebox{0.2cm}{%
    \centering
    \includegraphics[width=\textwidth, height=8cm]{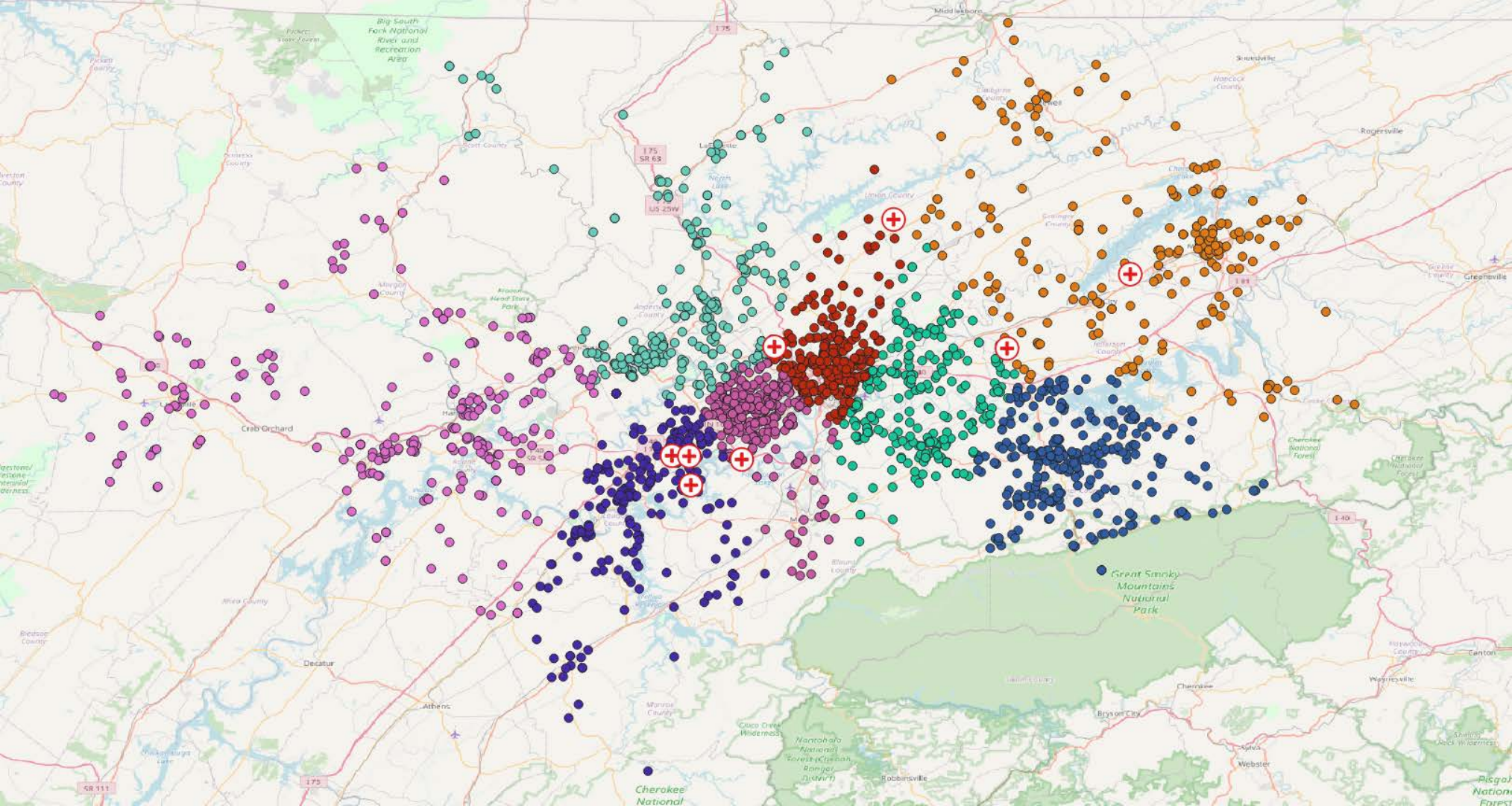}}
    \caption{Mapping of OT caregivers allocated patients}
    \label{fig:sub1_ot}
\end{subfigure}
\begin{subfigure}[b]{.4\textwidth}
  \centering
  \includegraphics[width=\textwidth]{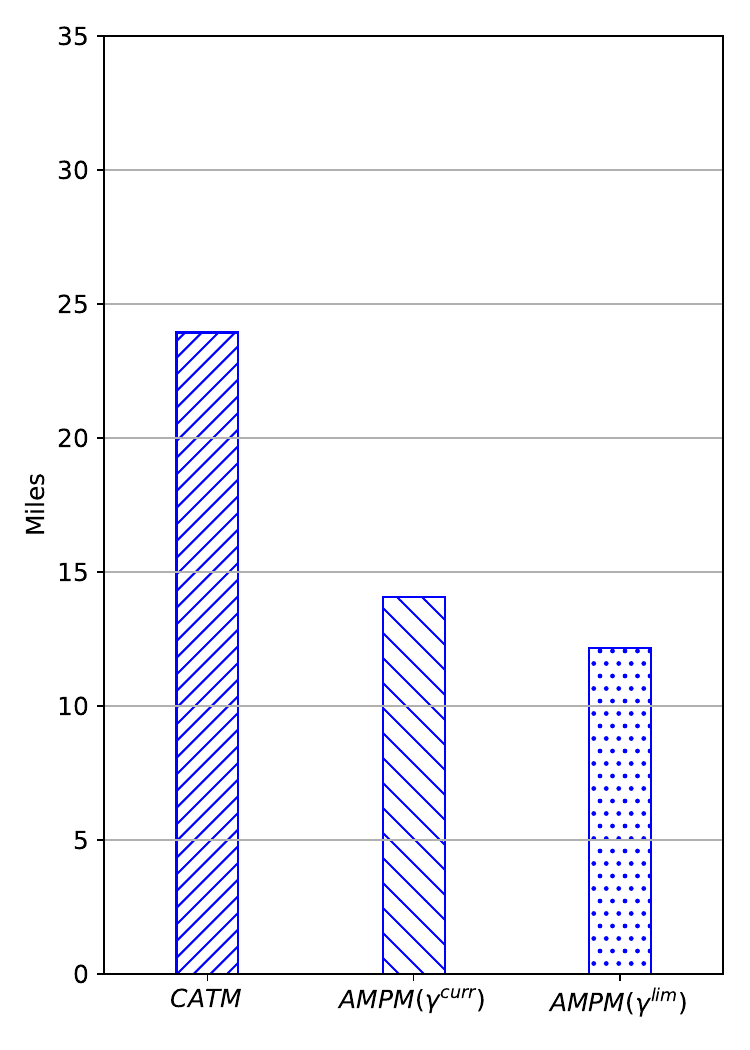}
  \caption{Current and expected average mileage of OT caregivers}
  \label{fig:sub2_ot}
\end{subfigure}
\begin{subfigure}[b]{.4\textwidth}
  \centering
  \includegraphics[width=\textwidth]{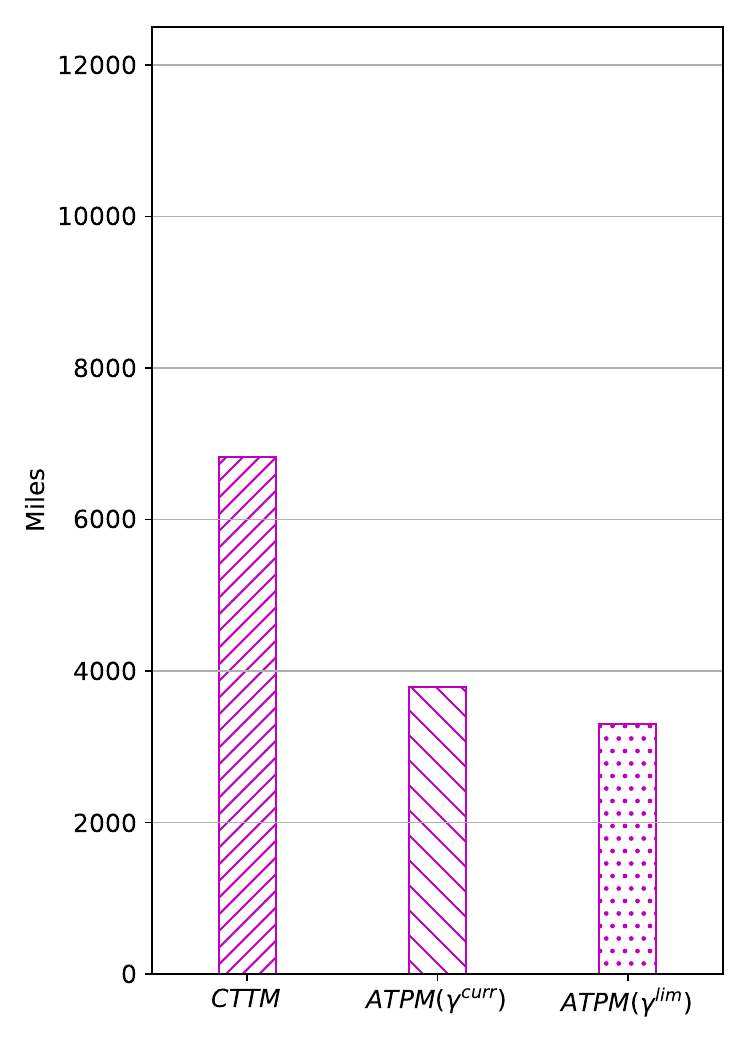}
  \caption{Current and expected total mileage of OT caregivers}
  \label{fig:sub3_ot}
\end{subfigure}
\caption{Results for OT discipline.}
\end{figure}

\begin{figure}[]
\centering
\begin{subfigure}{.8\textwidth}
  \raisebox{0.2cm}{%
    \centering
    \includegraphics[width=\textwidth, height=8cm]{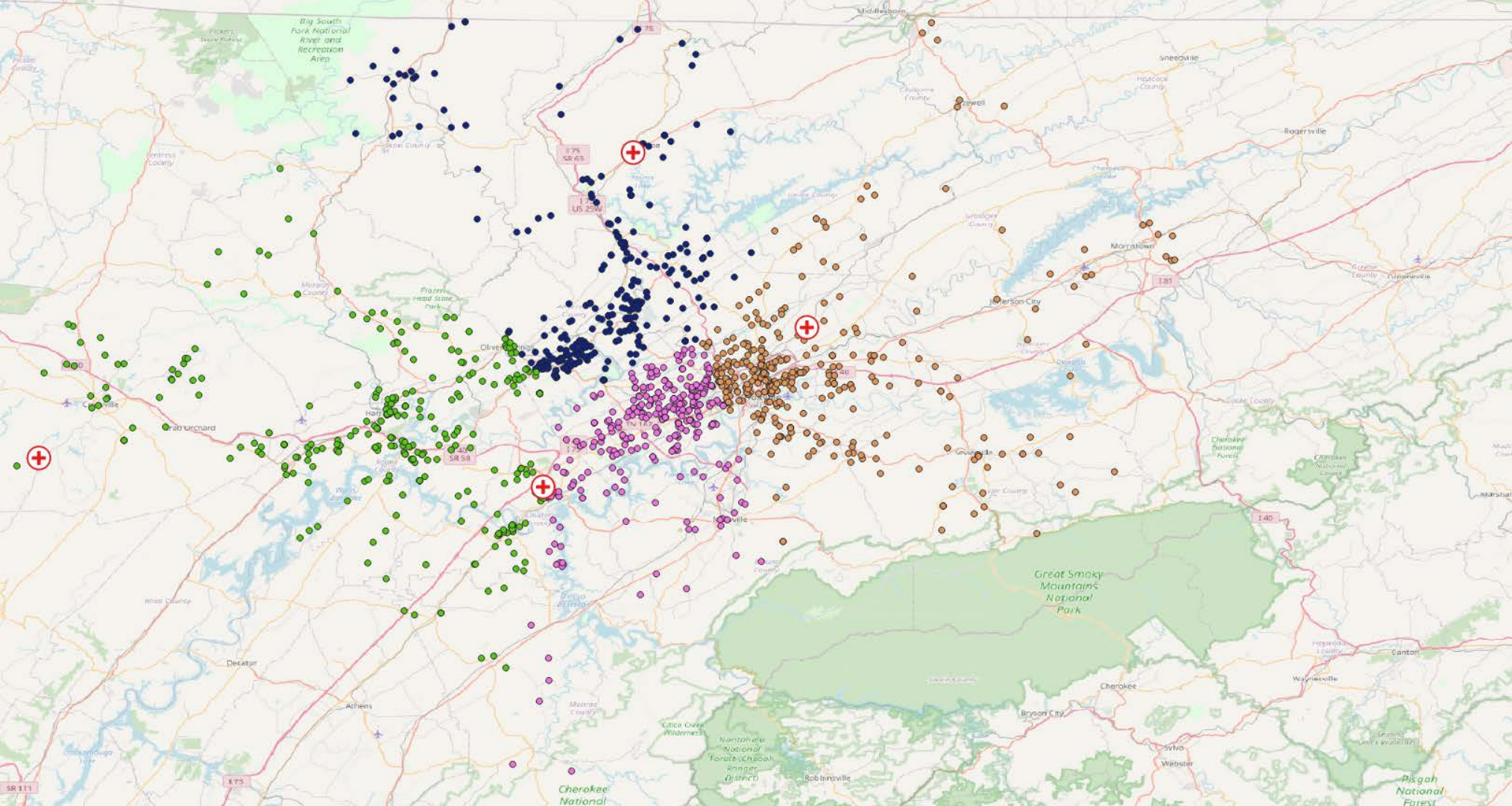}}
    \caption{Mapping of LPN caregivers allocated patients}
    \label{fig:sub1_lpn}
\end{subfigure}
\begin{subfigure}[b]{.4\textwidth}
  \centering
  \includegraphics[width=\textwidth]{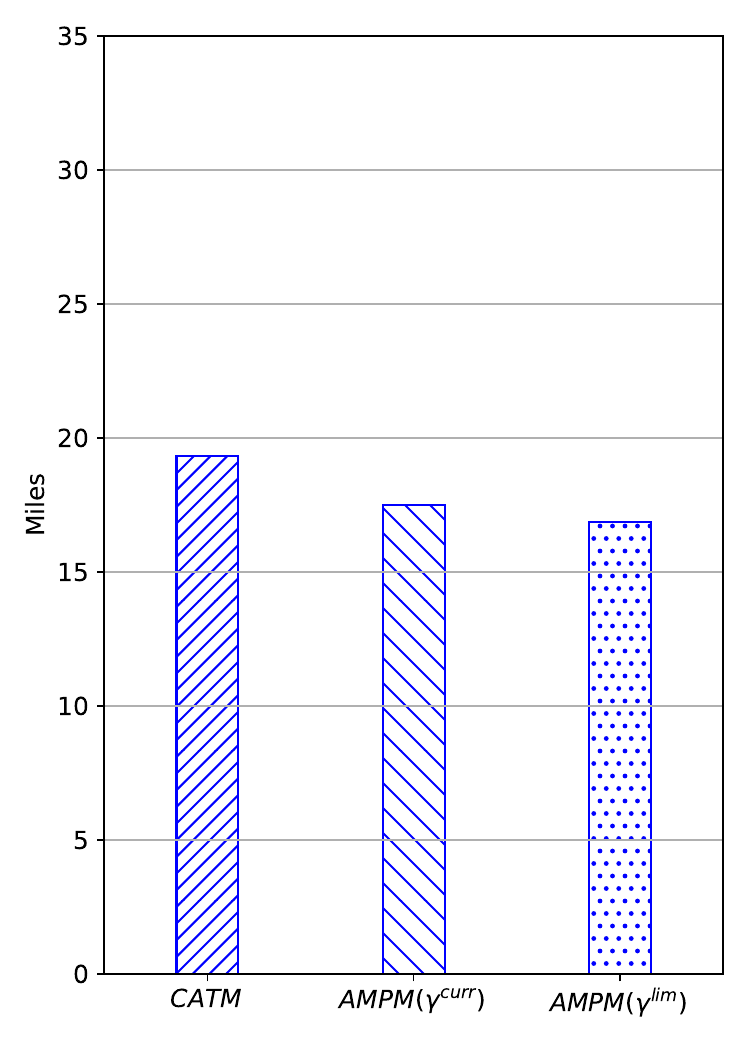}
  \caption{Current and expected average mileage of LPN caregivers}
  \label{fig:sub2_lpn}
\end{subfigure}
\begin{subfigure}[b]{.4\textwidth}
  \centering
  \includegraphics[width=\textwidth]{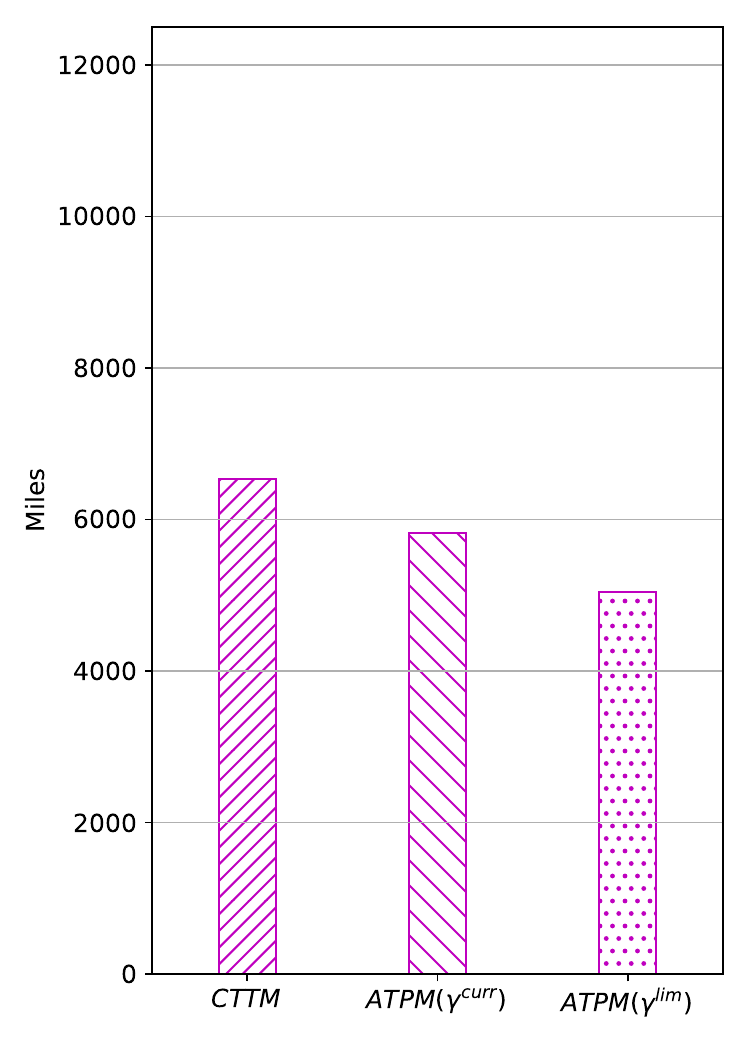}
  \caption{Current and expected total mileage of LPN caregivers}
  \label{fig:sub3_lpn}
\end{subfigure}
\caption{Results for LPN discipline.}
\end{figure}

\begin{figure}[]
\centering
\begin{subfigure}{.8\textwidth}
  \raisebox{0.2cm}{%
    \centering
    \includegraphics[width=\textwidth, height=8cm]{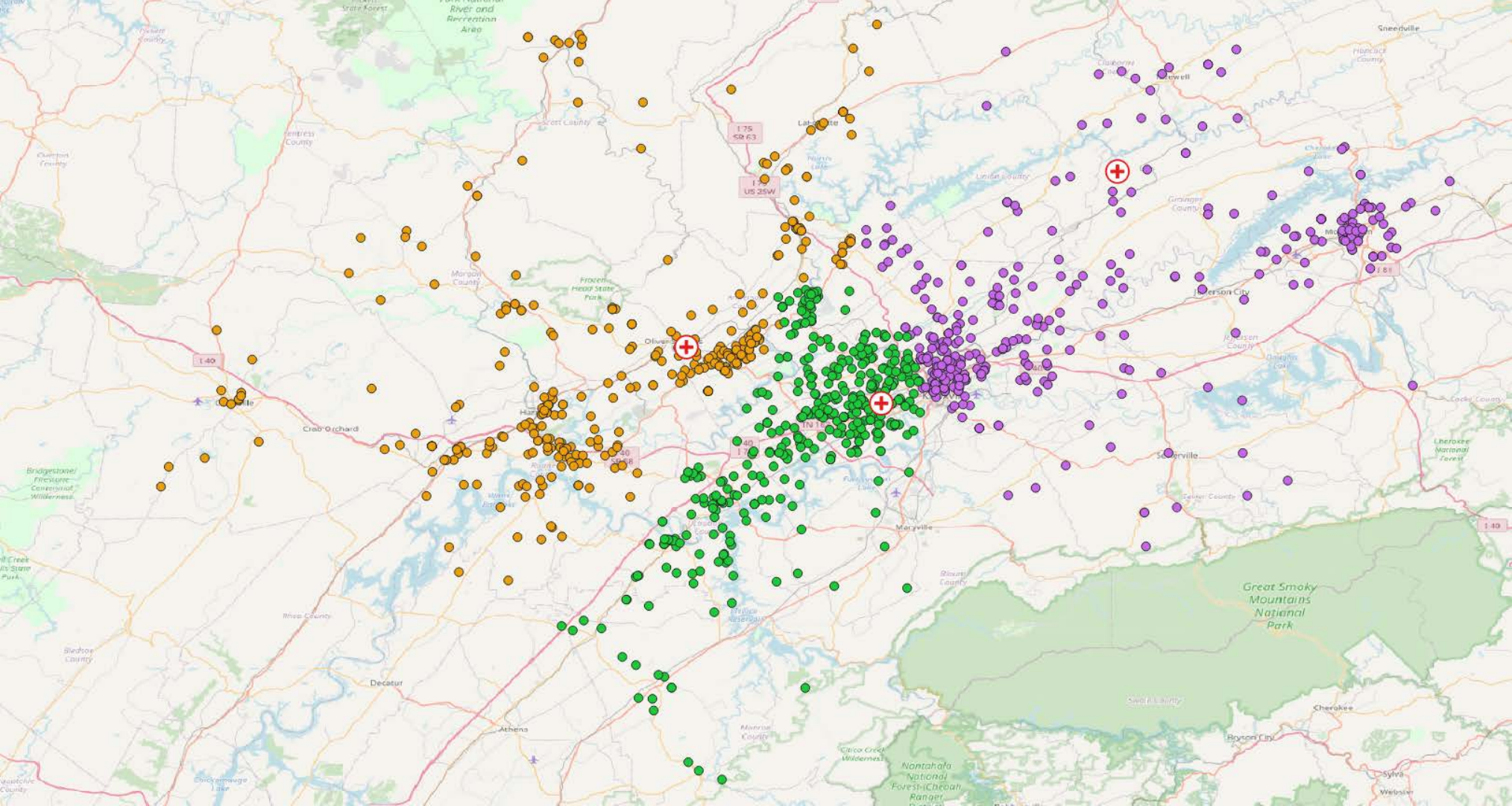}}
    \caption{Mapping of MSW caregivers allocated patients}
    \label{fig:sub1_msw}
\end{subfigure}
\begin{subfigure}[b]{.4\textwidth}
  \centering
  \includegraphics[width=\textwidth]{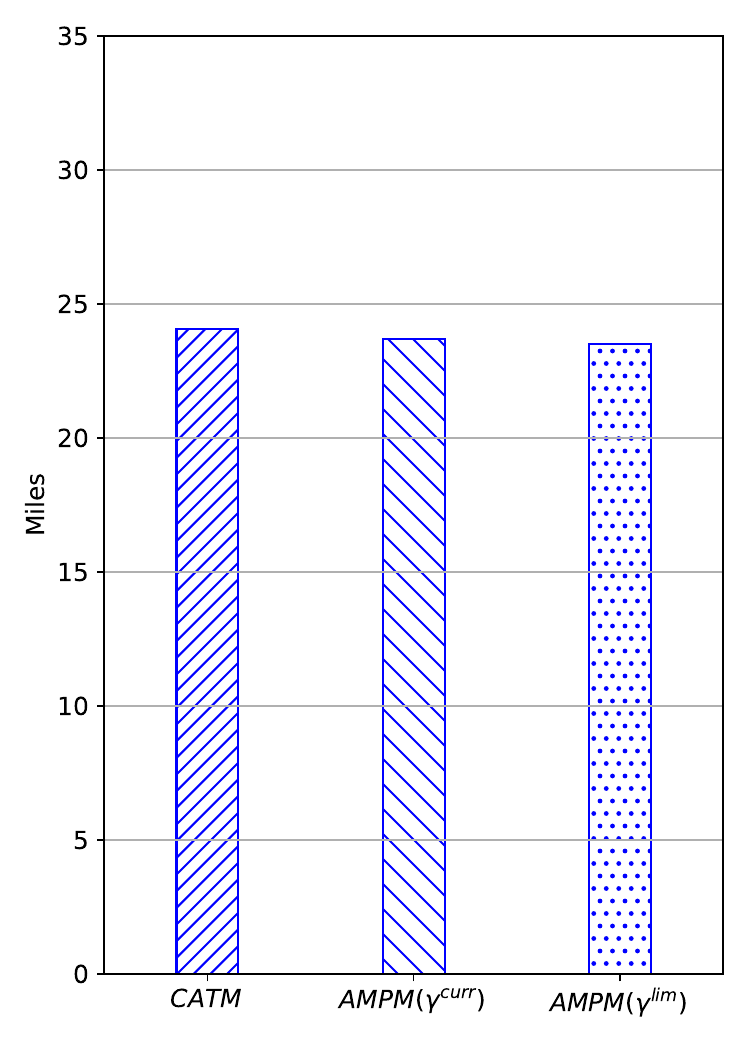}
  \caption{Current and expected average mileage of MSW caregivers}
  \label{fig:sub2_msw}
\end{subfigure}
\begin{subfigure}[b]{.4\textwidth}
  \centering
  \includegraphics[width=\textwidth]{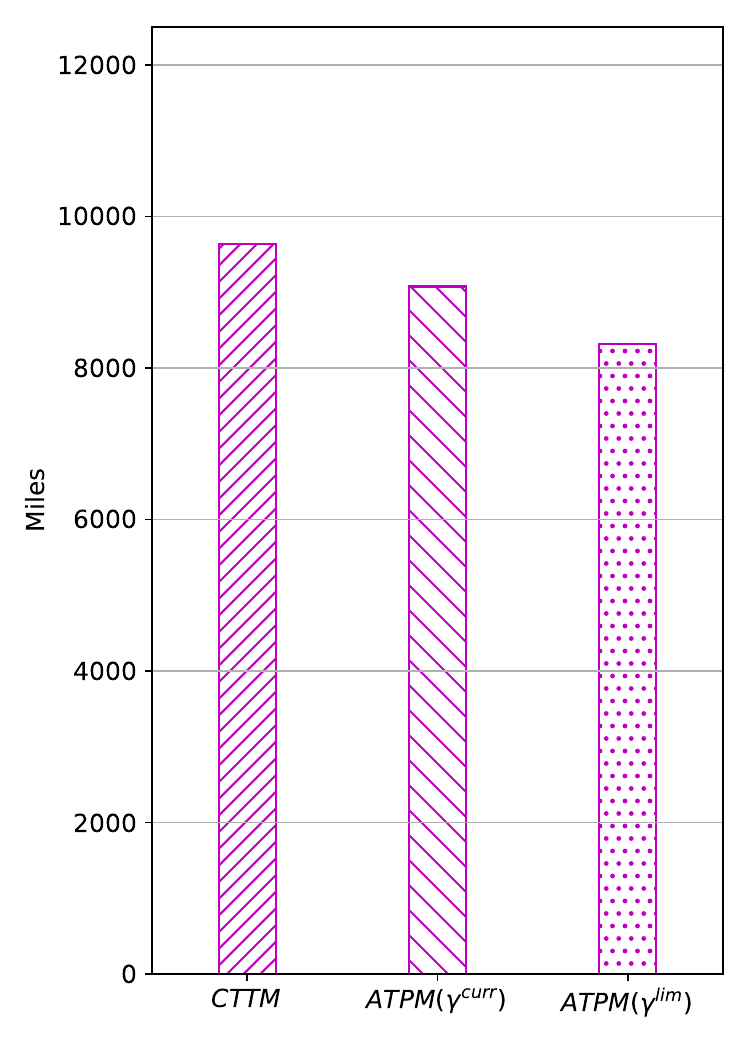}
  \caption{Current and expected total mileage of MSW caregivers}
  \label{fig:sub3_msw}
\end{subfigure}
\caption{Results for MSW discipline.}
\end{figure}

\begin{figure}[]
\centering
\begin{subfigure}{.8\textwidth}
  \raisebox{0.2cm}{%
    \centering
    \includegraphics[width=\textwidth, height=8cm]{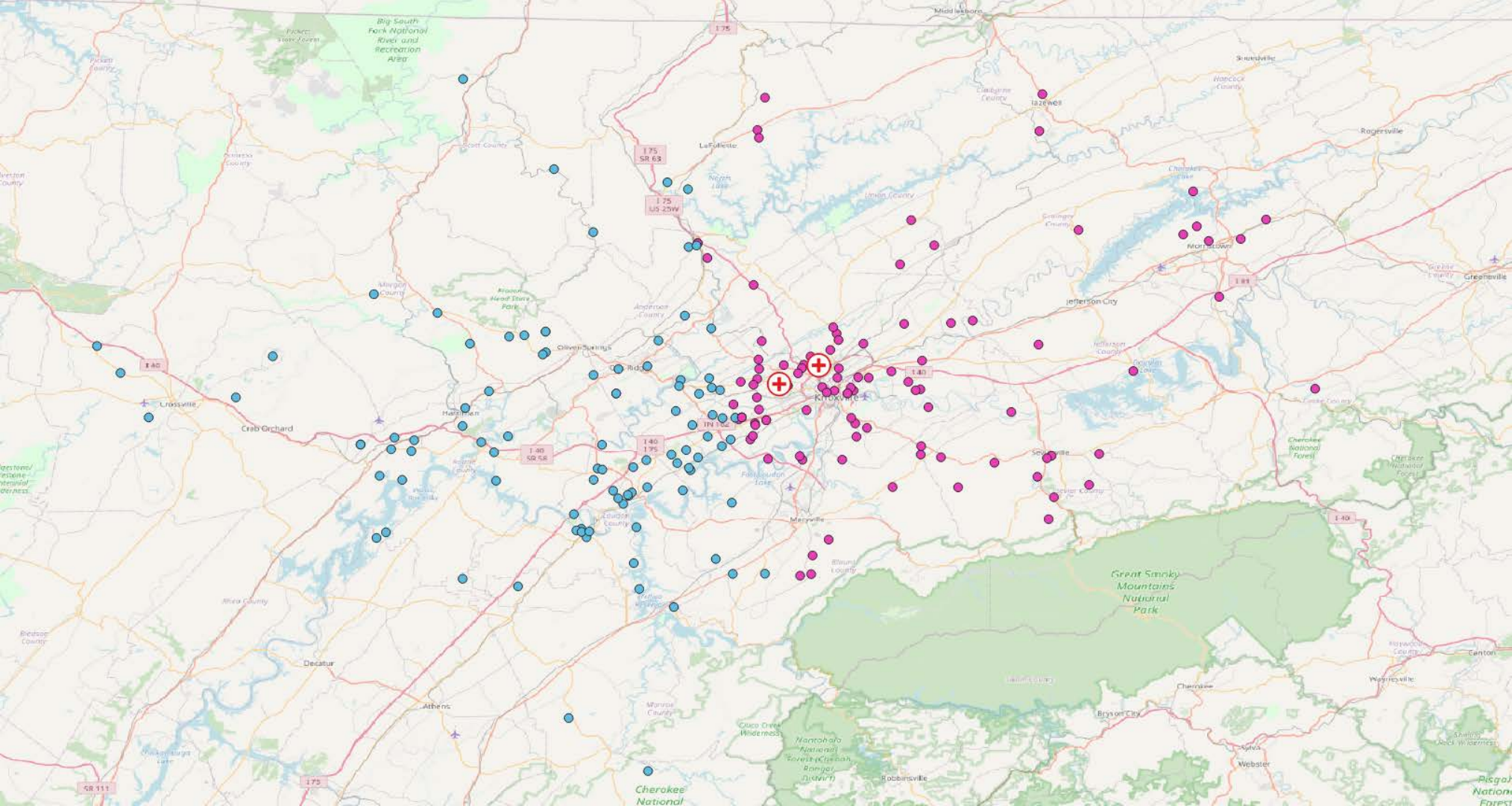}}
    \caption{Mapping of SLP caregivers allocated patients}
    \label{fig:sub1_slp}
\end{subfigure}
\begin{subfigure}[b]{.4\textwidth}
  \centering
  \includegraphics[width=\textwidth]{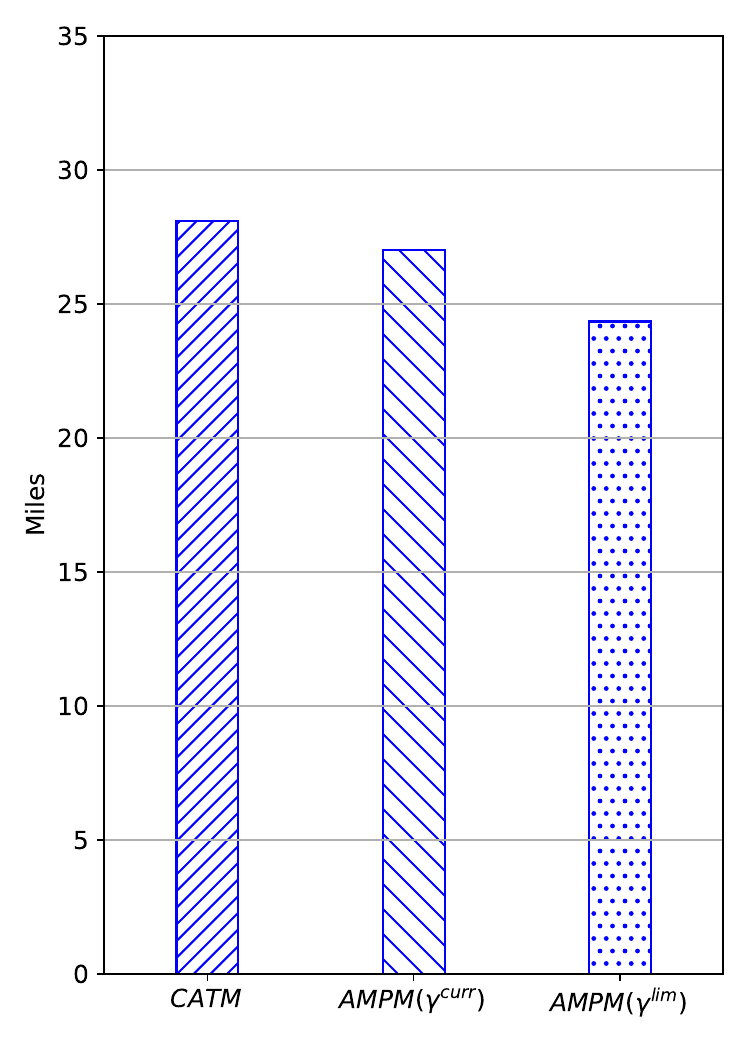}
  \caption{Current and expected average mileage of SLP caregivers}
  \label{fig:sub2_slp}
\end{subfigure}
\begin{subfigure}[b]{.4\textwidth}
  \centering
  \includegraphics[width=\textwidth]{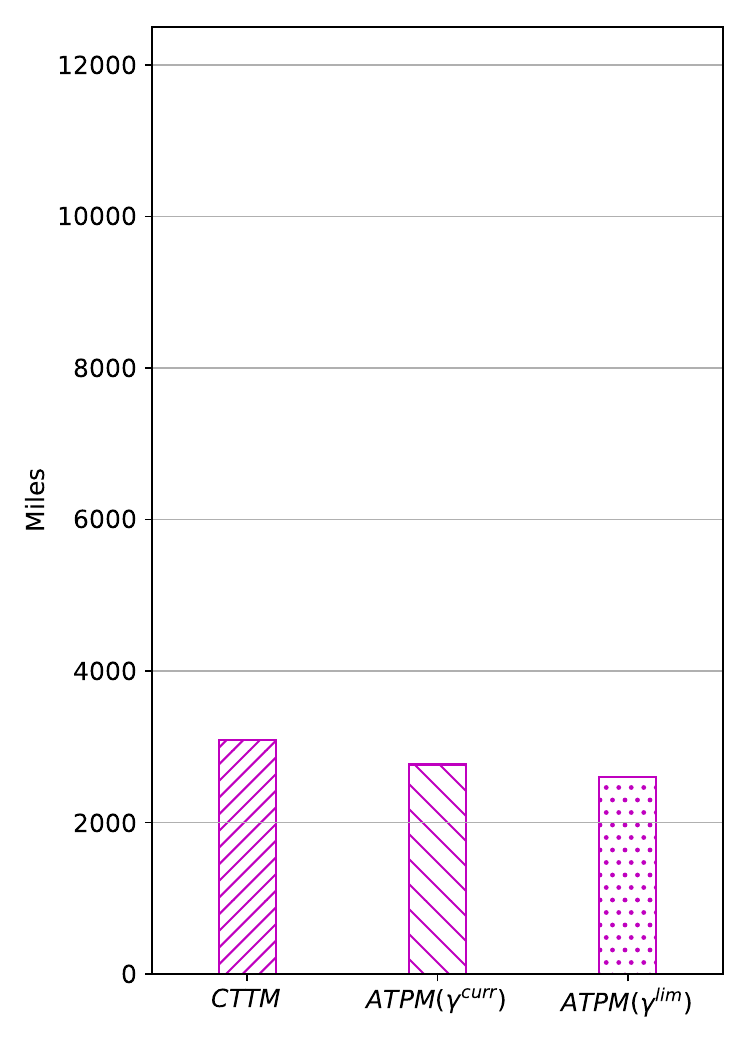}
  \caption{Current and expected total mileage of SLP caregivers}
  \label{fig:sub3_slp}
\end{subfigure}
\caption{Results for SLP discipline.}
\end{figure}

\newpage

\bibliographystyle{model5-names}
\biboptions{authoryear}
\bibliography{HHC_submitted.bib}

\end{document}